\documentclass[lettersize,journal]{IEEEtran}
\pdfoutput=1
\usepackage{amsmath,amsfonts}
\usepackage{algorithm}
\usepackage{array}
\usepackage{textcomp}
\usepackage{stfloats}
\usepackage{url}
\usepackage{verbatim}
\usepackage{graphicx}
\usepackage{cite}
\usepackage{comment}
\usepackage{amssymb}
\usepackage{algorithmicx}
\usepackage{algpseudocode}
\usepackage[switch]{lineno} 
\usepackage{subfigure}
\usepackage{verbatim}       
\usepackage{wrapfig}        
\usepackage{url}            
\usepackage{booktabs}       
\usepackage{nicefrac}       
\usepackage{microtype}      

\usepackage{hyperref}       
\usepackage{xcolor}         
\hypersetup{
	colorlinks=true,
	linkcolor=magenta,
	urlcolor=magenta,
	citecolor=[rgb]{0,0,0.4},
}

\newtheorem{theorem}{Theorem}

\begin{document}
	
	\title{Simultaneous Double Q-learning with Conservative Advantage Learning for Actor-Critic Methods}
	
	\author{Qing Li, Wengang Zhou, \emph{Senior Member, IEEE}, Zhenbo Lu, and Houqiang Li, \emph{Fellow, IEEE}
		\thanks{Q. Li, W. Zhou and H. Li are with the CAS Key Laboratory of Technology in Geo-spatial Information Processing and Application System, Department of Electronic Engineering and Information Science, University of Science and Technology of China, Hefei 230027, China (e-mail: liqingya@mail.ustc.edu.cn, zhwg@ustc.edu.cn, lihq@ustc.edu.cn).}
		\thanks{Z. Lu is with the Institute of Artificial Intelligence, Hefei Comprehensive National Science Center, Hefei 230000, China (e-mail:  luzhenbo\_2018@163.com).}
		
		\thanks{Corresponding authors: Wengang Zhou and Houqiang Li.}
	}
	
	
	
	\maketitle
	\begin{abstract}
		Actor-critic Reinforcement Learning (RL) algorithms have achieved impressive performance in continuous control tasks. However, they still suffer two nontrivial obstacles, \emph{i.e.,} low sample efficiency and overestimation bias. To this end, we propose Simultaneous Double Q-learning with Conservative Advantage Learning (SDQ-CAL). Our SDQ-CAL boosts the Double Q-learning for off-policy actor-critic RL based on a modification of the Bellman optimality operator with Advantage Learning. 
		Specifically, SDQ-CAL improves sample efficiency by modifying the reward to facilitate the distinction from experience between the optimal actions and the others. 
		Besides, it mitigates the overestimation issue by updating a pair of critics simultaneously upon double estimators.
		Extensive experiments reveal that our algorithm realizes less biased value estimation and achieves state-of-the-art performance in a range of continuous control benchmark tasks.
		We release the source code of our method at: \url{https://github.com/LQNew/SDQ-CAL}. 
	\end{abstract}
	
	
	\section{Introduction} \label{introduction}
	\IEEEPARstart{O}{ver} the past few years, RL has achieved impressive success in a wide variety of tasks, including games~\cite{mnih2015human,silver2017mastering,pachocki2018openai,vinyals2019grandmaster,zhu2016iterative} and robotic control~\cite{schulman2015trust,andrychowicz2017hindsight,mandlekar2020iris,singh2020parrot,li2019deep,yang2021hierarchical}. 
	Specifically, on continuous control, actor-critic RL algorithms~\cite{lillicrap2015continuous,fujimoto2018addressing,haarnoja2018soft,ciosek2019better} have been widely explored with remarkable performance. 
	However, there are two significant challenges hindering these approaches to scale to more complex tasks. 
	First, model-free deep RL methods are notorious for their poor sample efficiency~\cite{haarnoja2018soft,ciosek2019better}. 
	In practice, control tasks with relatively low complexity still require millions of interactions with the environment to obtain an acceptable policy. 
	Second, in continuous control domains, actor-critic algorithms suffer from overestimation bias that makes training unstable~\cite{fujimoto2018addressing}. 
	Both the above two issues severely limit the performance and the potential application of actor-critic RL algorithms.
	
	Generally, on-policy policy-gradient RL~\cite{schulman2015trust,schulman2017proximal} and off-policy actor-critic RL~\cite{lillicrap2015continuous,fujimoto2018addressing} are two representative types of model-free RL algorithms that currently dominate scalable learning for continuous control problems.
	On-policy policy-gradient algorithms are notoriously expensive in terms of their sample complexity~\cite{haarnoja2018soft,abdolmaleki2018maximum} since they require on-policy learning that is extravagantly expensive. 
	In contrast, off-policy actor-critic algorithms can obtain better sample efficiency as they can learn by reusing past experience. 
	However, they are difficult to tune due to their extreme brittleness and overestimation bias ~\cite{fujimoto2018addressing,haarnoja2018soft,abdolmaleki2018maximum}.
	To reduce the overestimation bias in the actor-critic setting, there have been research efforts to limit overestimation by taking the pessimistic estimate of multiple Q-functions for value estimation~\cite{fujimoto2018addressing,haarnoja2018soft}. 
	However, as shown recently in ~\cite{ciosek2019better,pan2020softmax}, these approaches suffer a large underestimation bias, which causes pessimistic underexploration and even worse performance.
	
	\begin{figure}[t]
		\small
		\centering
		\subfigure{
			\includegraphics[width=1.0\linewidth]{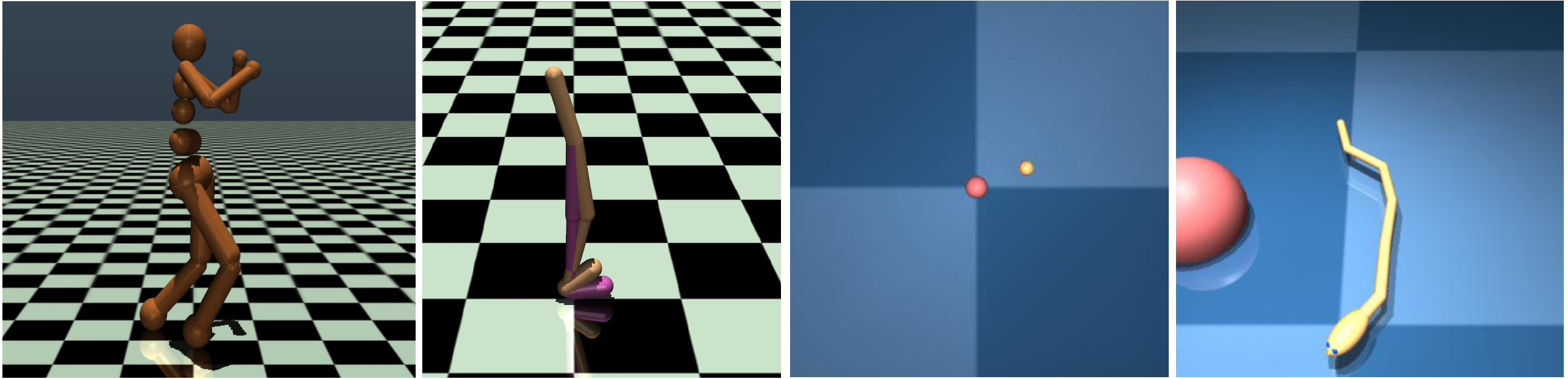}
		}
		\caption{Snapshots of MuJoCo and DeepMind Control Suite environments. The pictures from left to right are taken from Humanoid-v2, Walker2d-v2, point\_mass-easy, and swimmer-swimmer6.}
		\label{fig:figure1}
	\end{figure}
	
	To mitigate the overestimate bias, in this work, we formulate simultaneous Double Q-learning (SDQ), a novel extension of Double Q-learning~\cite{hasselt2010double}. 
	Though the mainstream view in the past was that directly applying the Double Q-learning for actor-critic methods still encountered the overestimation issue~\cite{fujimoto2018addressing,pan2020softmax}, we discover that by improving code-level design and implementation of Double Q-learning in actor-critic algorithms (\emph{i.e.,} simultaneous update and double-action selection), the actor-critic variants of Double Q-learning can bring in higher performance and lower estimation bias.
	In our work, we propose SDQ-CAL, an off-policy actor-critic algorithm that allows for evenly unbiased Q-value estimates by simultaneously updating a pair of independent critics upon double estimators. 
	
	Besides, to further improve sample efficiency, we formulate the reward function  with conservative Advantage Learning (CAL) to facilitate the distinction from experience between the optimal actions and other actions.
	As an action-gap increasing algorithm, Advantage Learning (AL)~\cite{baird1999gradient} has been extended to large discrete-action problems and has brought consistent performance improvement~\cite{bellemare2016unifying,ferret2020self,bellemare2016increasing}. 
	However, to our best knowledge, there is little work exploring Advantage Learning in continuous-action situations. 
	In our SDQ-CAL, we introduce conservative Advantage Learning into continuous control by taking the pessimistic estimate of multiple Q-functions for advantage-value estimation. 
	Compared with  Advantage Learning~\cite{baird1999gradient}, conservative Advantage Learning increases the action gap between the optimal actions and other actions by producing an approximate lower advantage-value bound, which prompts the RL agent to learn faster and further improves sample efficiency.
	
	To evaluate our algorithm, we conduct extensive experiments in continuous control benchmark tasks from OpenAI Gym~\cite{brockman2016openai} and DeepMind Control Suite~\cite{tassa2018deepmind} (Fig.~\ref{fig:figure1}) .
	The experimental results show that our SDQ-CAL attains a substantial improvement in both performance and sample efficiency over prior methods, including Deep Deterministic Policy Gradient  (DDPG)~\cite{lillicrap2015continuous}, Proximal Policy Optimization (PPO)~\cite{schulman2017proximal}, Twin Delayed Deep Deterministic policy gradient (TD3)~\cite{fujimoto2018addressing}, Soft Actor-Critic (SAC)~\cite{haarnoja2018soft}, and Softmax Deep Double Deterministic Policy Gradients (SD3)~\cite{pan2020softmax}. 
	
	In brief, the main contributions of this work are three-fold as follows:
	\begin{itemize}
		\item We show that simultaneous Double Q-learning, a simple extension based on Double Q-learning, will lead to less biased value estimation.
		
		\item We propose conservative Advantage Learning, a novel generalization of Advantage Learning for actor-critic algorithms in continuous control, to improve the sample efficiency.
		
		\item We 
		empirically demonstrate that our algorithm achieves better performance and higher sample efficiency than prior algorithms.	
	\end{itemize}
	
	The remainder of this paper is organized as follows. We describe the preliminaries and related work in Section~\ref{relatedwork}, the proposed method in Section~\ref{SDQCALtable} and Section~\ref{SDQCALActorCritic}, and experiments in Section~\ref{experiments}. Finally, we conclude this work in Section~\ref{conclusion}.

	\section{Preliminaries and Related Work} \label{relatedwork}
	\subsection{Notations in RL}
	We consider the standard Markov Decision Process (MDP) formalism $(\mathcal{S}, \mathcal{A}, P, r, \gamma)$~\cite{sutton1998introduction}, which consists of a state space $\mathcal{S}$, an action space $\mathcal{A}$, transition dynamics $P(s^\prime|s, a)$, a stochastic reward function $r: \mathcal{S} \times \mathcal{A} \rightarrow \mathbb{R}$, and a discount factor $\gamma \in [0, 1)$. 
	A stochastic policy $\pi$ maps states to a probability distribution over the actions. 
	For an agent following the policy $\pi$, the Q function, denoted as $Q^\pi(s, a)$, is defined as the expectation of cumulative discounted future rewards, \emph{i.e.}, $\begin{aligned} Q^{\pi}(s, a) \doteq \mathbb{E}_{\pi}\left[\sum_{t=0}^{\infty} \gamma^{t} r_{t+1} | s_{0}=s, a_{0}=a\right] \end{aligned}$. 
	The optimal Q function is $\begin{aligned} Q^{\star}(s, a) \doteq \max _{\pi} Q^{\pi}(s, a) \end{aligned}$. 
	The Bellman optimality equation~\cite{bellman1957dynamic} represents $Q^\star$ as: 
	\begin{equation}
		\label{OptimalBellman}
		Q^{*}(s, a) =  r(s, a) + \gamma\sum_{s^{\prime}} P\left(s^{\prime} | s, a\right) \max _{a^{\prime}} Q^{*}\left(s^{\prime}, a^{\prime}\right).
	\end{equation}
	
	\subsection{Q-learning and Its Variants}
	Q-learning~\cite{watkins1992q} is a commonly used method to iteratively improve an estimation of value function. Q-learning obtains the optimal policy $\pi^\star$ implicitly by calculating state-action value $Q(s, a)$ which measures the goodness of the given state-action with respect to the current behavioral policy.
	
	In continuous and high-dimensional action space, it is expensive to do the max operator over $\mathcal{A}$ in Eq.~\eqref{OptimalBellman}.
	To this end, Deterministic Policy Gradient (DPG)~\cite{silver2014deterministic} adopts a deterministic policy $\mu : \mathcal{S} \rightarrow \mathcal{A}$ to approximate the optimal action $\begin{aligned} a^\star = \arg \max_a Q(s, a) \end{aligned}$.
	The deterministic policy is updated by applying the chain rule to the expected return with respect to the policy parameters: 
	\begin{equation}
		\begin{aligned}
			\nabla_{\theta_{\mu}} J =\mathbb{E}_{s_{t} \sim \rho^{\beta}}\left[\left.\left.\nabla_{a} Q\left(s, a \right)\right|_{s=s_{t}, a=\mu\left(s_{t}\right)} \nabla_{\theta_{\mu}} \mu\left(s \right)\right|_{s=s_{t}}\right],
		\end{aligned}
	\end{equation}
	where $\rho^{\beta}$ is the discounted state visitation distribution for the policy $\beta$.
	DDPG~\cite{lillicrap2015continuous} further extends Q-learning~\cite{watkins1992q} to continuous control based on DPG and successfully applies the deterministic policy gradient method to solve high-dimensional problems.
	
	To mitigate the overestimation issue caused by the max operator across action values in Q-learning~\cite{watkins1992q}, Double Q-learning~\cite{hasselt2010double} is proposed to estimate the value of the next state using the double estimators, $\emph{i.e.},$ 
	\begin{equation} 
		\begin{aligned}
			Q\left(s^\prime, a^{\prime \star}\right)  \approx Q^A\left(s^\prime, \arg \max_{a^\prime} Q^B\left(s^\prime, a^\prime\right)\right),
		\end{aligned}
	\end{equation}
	where $Q^A$ and $Q^B$ are two independent function approximators for Q values. 
	
	\subsection{Q-value Estimation in RL}
	There have been many efforts to obtain good value estimation in RL. Double Q-Learning~\cite{hasselt2010double} applies double estimators to Q-Learning to make evenly unbiased value estimates. 
	Averaged Deep Q-Network (DQN)~\cite{anschel2017averaged} averages previously learned Q-value estimates to reduce the approximation error variance. Random Ensemble Mixture (REM)~\cite{agarwal2020optimistic} obtains a robust Q-learning by enforcing optimal Bellman consistency on a random convex combination of multiple Q-value estimates under the offline RL setting. In this paper, our focus is to investigate the properties and benefits of Double Q-learning coupled with Advantage Learning in continuous control, where we provide new analysis and insight. 
	
	TD3 is an efficient algorithm which mitigates the overestimation issue in continuous action space. 
	However, it suffers a large underestimation bias, which causes pessimistic underexploration~\cite{ciosek2019better}. Unlike TD3 which favors underestimation by Clipped Double Q-learning, in SDQ-CAL, we propose simultaneous Double Q-learning algorithm that updates a pair of independent critics upon double estimators simultaneously to mitigate the overestimation bias and improve the underestimation bias. 
	SD3~\cite{pan2020softmax} employs a softmax operator upon double estimators to reduce the underestimation bias encountered in TD3. However, SD3 can be computationally expensive due to the application of the softmax operator, while SDQ-CAL is efficient.
	
	\subsection{Advantage Learning}
	Advantage Learning~\cite{baird1999gradient,bellemare2016increasing} follows modified Bellman operator $\mathcal{T}_\text{AL}$ by adding the advantage $Q(s, a) - V(s)$ to the reward, with $\begin{aligned} V(s) = \max_{a} Q\left(s, a\right) \end{aligned}$:
	\begin{equation} 
		\begin{aligned}
			r_\text{AL}(s, a) &= r(s, a) + \beta \left(Q(s, a) - \max_{a} Q\left(s, a\right)\right), \\
			\mathcal{T}_\text{AL} Q(s, a) &= r_\text{AL}(s, a) + \gamma\sum_{s^{\prime}} P\left(s^{\prime} | s, a\right) \max _{a^{\prime}} Q\left(s^{\prime}, a^{\prime}\right).
		\end{aligned}
	\end{equation}
	Advantage Learning can enable faster solution of RL problems since it increases the action gap, which facilitates the distinction between the optimal actions and other actions in experience. In discrete control tasks, such as Atari games~\cite{bellemare2013arcade}, Advantage Learning is shown to achieve better performance and bring higher sample efficiency~\cite{ferret2020self} over different algorithms.
	
	\section{Simultaneous Double Q-learning with Conservative Advantage Learning} \label{SDQCALtable}
	In this section, we apply the Advantage Learning for value estimation in Q-learning algorithms with double estimators. 
	In the beginning, for a finite MDP setting, we maintain two tabular value estimates $Q^A$ and $Q^B$. At each time step, firstly, we 
	reshape the rewards for $Q^A$ and $Q^B$ independently by defining:
	\begin{equation}
		\label{CAL}
		\begin{aligned}
			r_A &= r(s,a) + \\ &  \quad \quad \beta \left(\min \left(Q^A(s, a), Q^B(s, a)\right) - \max_a Q^A\left(s, a\right) \right), \\ r_B &= r(s,a) + \\  & \quad \quad \beta \left(\min \left(Q^A(s, a), Q^B(s, a)\right) - \max_a Q^B(s, a)\right),
		\end{aligned}
	\end{equation}
	where $\beta \in [0, 1)$ denotes how much the value of conservative advantage is added to the original reward, with $\beta = 0$ corresponding to the initial reward with no other items added.
	
	We call the formulation in Eq.~\eqref{CAL} as conservative Advantage Learning since we take the minimum between the two value estimates as the Q values of the sampled state-action pairs, which produces an approximate lower advantage-value bound. 
	Although the mathematical form of conservative Advantage Learning is similar with the mathematical form of CDQ in TD3~\cite{fujimoto2018addressing}, they have completely different purposes and usages. 
	Specifically, in our method, conservative Advantage Learning is designed to increase the action gap between the optimal actions and other actions. Generally, increasing the action gap can make the RL agent learn faster and bring in improved sample efficiency, as it contributes to the distinction in experience between the optimal actions and the others.
	
	Next, we select actions $\begin{aligned} a^{\prime \star}= \arg\max_{a^\prime} Q^A(s^{\prime},a^\prime)\end{aligned}$ and $\begin{aligned} b^{\prime \star}= \arg\max_{a^\prime} Q^B(s^{\prime},a^\prime) \end{aligned}$. 
	Then we make unbiased value estimations of the selected optimal actions by using the opposite value estimate~\cite{hasselt2010double} and perform an update by setting targets $y_A$ and $y_B$ as follows:
	\begin{equation}
		\begin{aligned}
			y_A &= r_A + \gamma Q^B(s^{\prime}, a^{\prime\star}), \\ y_B &= r_B + \gamma Q^A(s^{\prime}, b^{\prime\star}).
		\end{aligned}
	\end{equation}
	
	Finally, we update the value estimates $Q^A$ and $Q^B$ simultaneously with respect to the targets and learning rate $\alpha(s,a)$:
	\begin{equation}
		\begin{aligned}
			&Q^{A}(s, a) \leftarrow Q^{A}(s, a) + \alpha(s, a)\left(y_A - Q^{A}(s, a)\right), \\
			&Q^{B}(s, a) \leftarrow Q^{B}(s, a)+\alpha(s, a)\left(y_B-Q^{B}(s, a)\right).
		\end{aligned}
	\end{equation}
	It is worth noting that when updating the value estimates, we no longer update either $Q^A$ or $Q^B$ randomly as suggested in~\cite{hasselt2010double}. 
	We update both $Q^A$ and $Q^B$ simultaneously at each iteration since our interest is more towards the approximation error under the function approximation setting, which will be discussed in the next section. 
	Taking all the methods mentioned above together, we propose the new approach called SDQ-CAL and summarize the approach in Algorithm~\ref{alg:algorithm1}.
	
	\begin{algorithm} 
		\caption{SDQ-CAL Algorithm}  
		\begin{algorithmic}[1] 
			\State \textbf{Initialize} $Q^A$, $Q^B$, $s$.
			\State \textbf{Loop for each episode:}
			\State \quad Choose $a$ based on $Q^A$ and $Q^B$, observe $r, s^{\prime}$.
			\State \quad // Reshape the reward.
			\State \quad $Q(s, a) = \min \left(Q^A(s, a), Q^B(s, a)\right)$.
			\State \quad Set $\begin{aligned} r_A = r(s,a) + \beta \left(Q(s, a) - \max_a Q^A(s, a)\right) \end{aligned}$. 
			\State \quad Set $\begin{aligned} r_B = r(s,a) + \beta \left(Q(s, a) - \max_a Q^B(s, a)\right) \end{aligned}$.
			\State \quad // Update $Q^A$ and $Q^B$.
			\State \quad Define $\begin{aligned} a^{\prime \star}= \arg\max_{a^\prime} Q^A(s^{\prime},a^\prime) \end{aligned}$.
			\State \quad Define $\begin{aligned} b^{\prime \star}= \arg\max_{a^\prime} Q^B(s^{\prime},a^\prime) \end{aligned}$.
			\State \quad Set $y_A = r_A + \gamma Q^B(s^{\prime}, a^{\prime\star})$.
			\State \quad Set $y_B = r_B + \gamma Q^A(s^{\prime}, b^{\prime\star})$.
			\State \quad $Q^{A}(s, a) \leftarrow Q^{A}(s, a) + \alpha(s, a)\left(y_A - Q^{A}(s, a)\right)$,
			\State \quad $Q^{B}(s, a) \leftarrow Q^{B}(s, a) + \alpha(s, a)\left(y_B - Q^{B}(s, a)\right)$.
			\State \quad $s \leftarrow s^\prime$.
			\State \textbf{until} $s$ is terminal.
		\end{algorithmic}
		\label{alg:algorithm1}
	\end{algorithm}

	\textbf{The effect of \texorpdfstring{$\beta$}{Lg}.} 
	We conclude this section by discussing the parameter $\beta$ in  SDQ-CAL. 
	We can find that while $\beta$ gets larger, the proportion of the conservative advantage in the reshaped reward becomes larger. 
	In the meantime, since the proportion of the conservative advantage becomes larger, the relative proportion of Q values for the expected future return decreases, which means that the return objective takes future rewards into account more weakly. 
	In a nutshell, introducing $\beta$ encourages the RL agent to concentrate more on learning the distinction between the optimal actions and sampled actions in experience, which prompts the RL agent to learn faster and further brings in improved sample efficiency. 
	However, a large $\beta$ can also make the RL agent become more focused on the current reshaped rewards while reducing access to future rewards so that the return may be reduced. 
	Therefore, with conservative Advantage Learning, we look for a balance between sample efficiency and future rewards with a relatively small adjustable coefficient $\beta$.
	
	\section{SDQ-CAL for Actor-Critic Methods} \label{SDQCALActorCritic}
	In this section, we propose to build SDQ-CAL upon DDPG and employ our algorithm in standard actor-critic algorithms. 
	
	\subsection{Policy Iteration in Actor-Critic}
	We maintain a pair of actors ($\pi_{\phi_1}$, $\pi_{\phi_2}$) and critics  ($Q_{\theta_1}$, $Q_{\theta_2}$) where $\pi_{\phi_1}$ is optimized with respect to $Q_{\theta_1}$ and $\pi_{\phi_2}$ with respect to $Q_{\theta_2}$. 
	Analogous to Algorithm~\ref{alg:algorithm1}, we perform conservative Advantage Learning by defining the reshaped rewards as:
	\begin{equation}
		\label{deepCAL}
		\begin{aligned}
			r_1 &= r(s,a) + \\ & \quad \quad \beta \left(\min \left(Q_{\theta^\prime_1}(s, a), Q_{\theta^\prime_2}(s, a)\right) - Q_{\theta^\prime_1}\left(s, \pi_{\phi_1}\left(s\right)\right)\right), \\ r_2 &= r(s,a) + \\ & \quad \quad \beta \left(\min \left(Q_{\theta^\prime_1}(s, a), Q_{\theta^\prime_2}(s, a)\right) - Q_{\theta^\prime_2}\left(s, \pi_{\phi_2}\left(s\right)\right)\right),
		\end{aligned}
	\end{equation}
	where $\theta^\prime_1$ and $\theta^\prime_2$ are the delayed parameters of $\theta_1$ and $\theta_2$, respectively.
	
	In the policy evaluation process, firstly, we select continuous actions $\begin{aligned} a_1^{\prime \star}= \pi_{\phi_{1}}\left(s^{\prime}\right)\end{aligned}$ and $\begin{aligned} a_2^{\prime \star}= \pi_{\phi_{2}}\left(s^{\prime}\right)\end{aligned}$. Next, we set the target updates of the critics as follows:
	\begin{equation}
		\label{targetupdate}
		\begin{aligned}
			y_{1} &= r_1 + \gamma Q_{\theta_{2}^{\prime}}\left(s^{\prime}, a_1^{\prime \star}\right),\\ y_{2} &= r_2 + \gamma Q_{\theta_{1}^{\prime}}\left(s^{\prime}, a_2^{\prime \star}\right). 
		\end{aligned}
	\end{equation}
	Then the update of the critics can be performed by minimizing the loss of each critic:
	\begin{equation}
		\begin{aligned}
			J_Q(\theta_i) = \frac{1}{2}\left(y_{i}-Q_{\theta_i}\left(s, a\right)\right)^2|_{i=1,2}.
		\end{aligned}
	\end{equation}
	
	It is notable that when applying SDQ-CAL for actor-critic methods, the approximate Q function is represented not as a table but as the parameterized functional form with neural networks. Since our interest is more towards the approximate error, we argue that performing simultaneous update can make neural networks fit Q functions better. Therefore, in SDQ-CAL, we update both $Q_{\theta_1}$ and $Q_{\theta_2}$ simultaneously at each time step.
	
	As for the policy improvement process, we apply the deterministic policy gradient~\cite{silver2014deterministic} to update the actors:
	\begin{equation}
		\begin{aligned}
			\nabla_{\phi_i} J_\pi(\phi_i) = \nabla_{a} Q_{\theta_i}(s, a) |_{a=\pi_{\phi_i}(s)} \nabla_{\phi_i} \pi_{\phi_i}(s)|_{i=1,2}.
		\end{aligned}
	\end{equation}
	
	\subsection{Interaction in Actor-Critic} 
	Under the finite MDP setting, we can calculate the average or the sum of the two Q values for each action and then perform $\epsilon$-greedy exploration with the resulting Q values to interact with the environment. 
	However, in continuous domains, it is nontrivial to select a reasonable action since the action space is continuous and high-dimensional. 
	In our work, we innovatively propose double-action selection (DAS), which selects actions based on both actors and critics. 
	Specifically, at each time step $t$, firstly, we calculate a pair of candidate actions based on $\pi_{\phi_1}$ and $\pi_{\phi_2}$: $a_t^1 = \pi_{\phi_1}(s_t)$ and $a_t^2 = \pi_{\phi_2}(s_t)$. 
	Then we utilize the critics $Q_{\theta_1}$ and $Q_{\theta_2}$ to select the action which maximizes the sum of the Q values:
	\begin{equation}
		\label{actionselection1}
		a_t^\prime= \arg \max_{a_t^i} \left(Q_{\theta_1}(s_t, a_t^i) + Q_{\theta_2}(s_t, a_t^i)\right) |_{i=1, 2}.
	\end{equation}
	When interacting with the environment, we add a small amount of random noise sampled from normal distribution to the chosen action $a_t^\prime$ to encourage exploration: 
	\begin{equation}
		\label{actionselection2}
		a_t = a_t^\prime + \epsilon,\,\,\, \epsilon \sim \mathcal{N}(0, \sigma).
	\end{equation}
	
	The pseudo-code of the body algorithm of SDQ-CAL for actor-critic is presented in Algorithm~\ref{alg:algorithm3}.
	
	\begin{figure*}[htbp]
		\centering
		\subfigure[Ant-v2]{
			\includegraphics[width=0.240\linewidth]{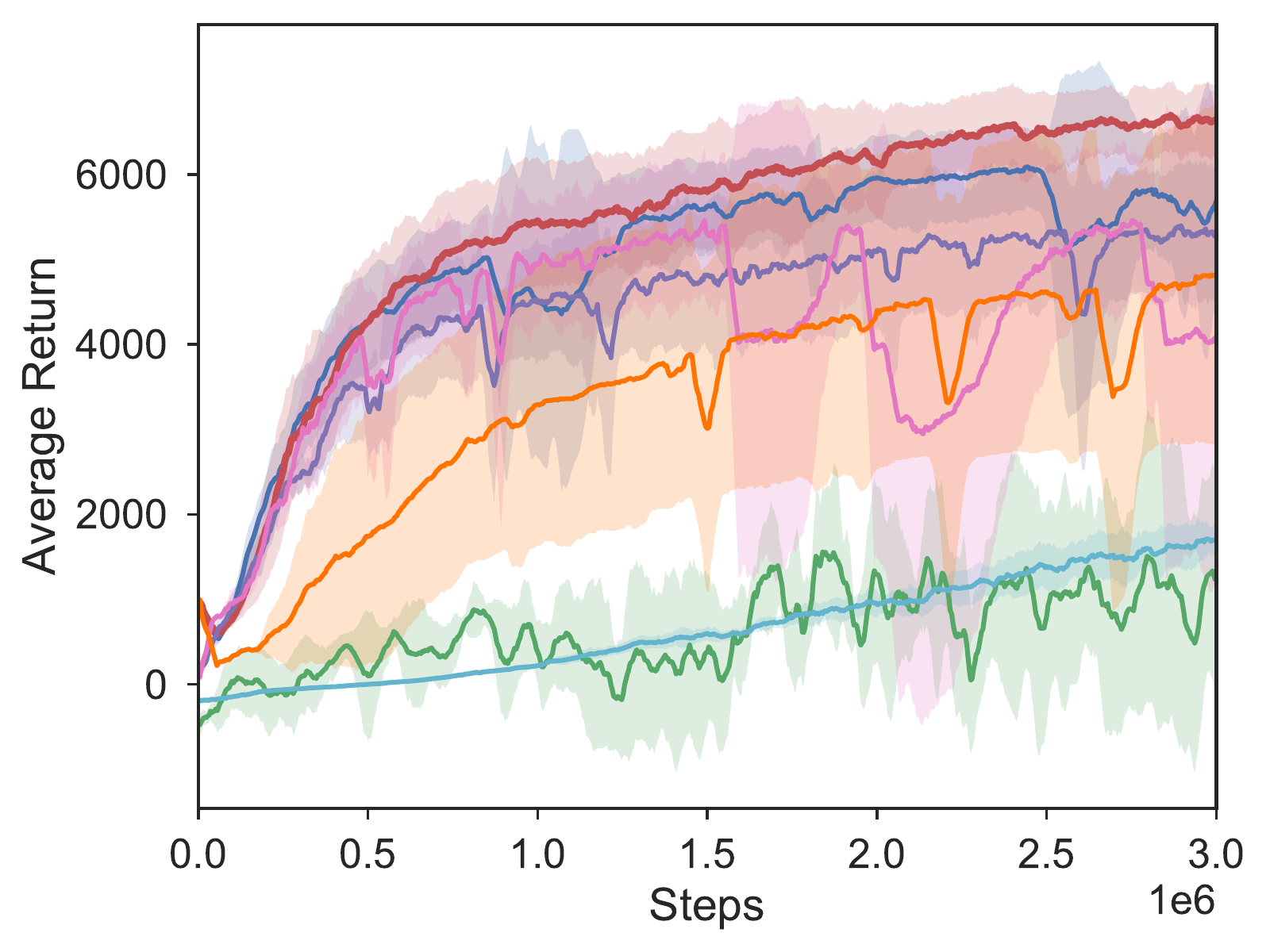}
		}
		\subfigure[HalfCheetah-v2]{
			\includegraphics[width=0.240\linewidth]{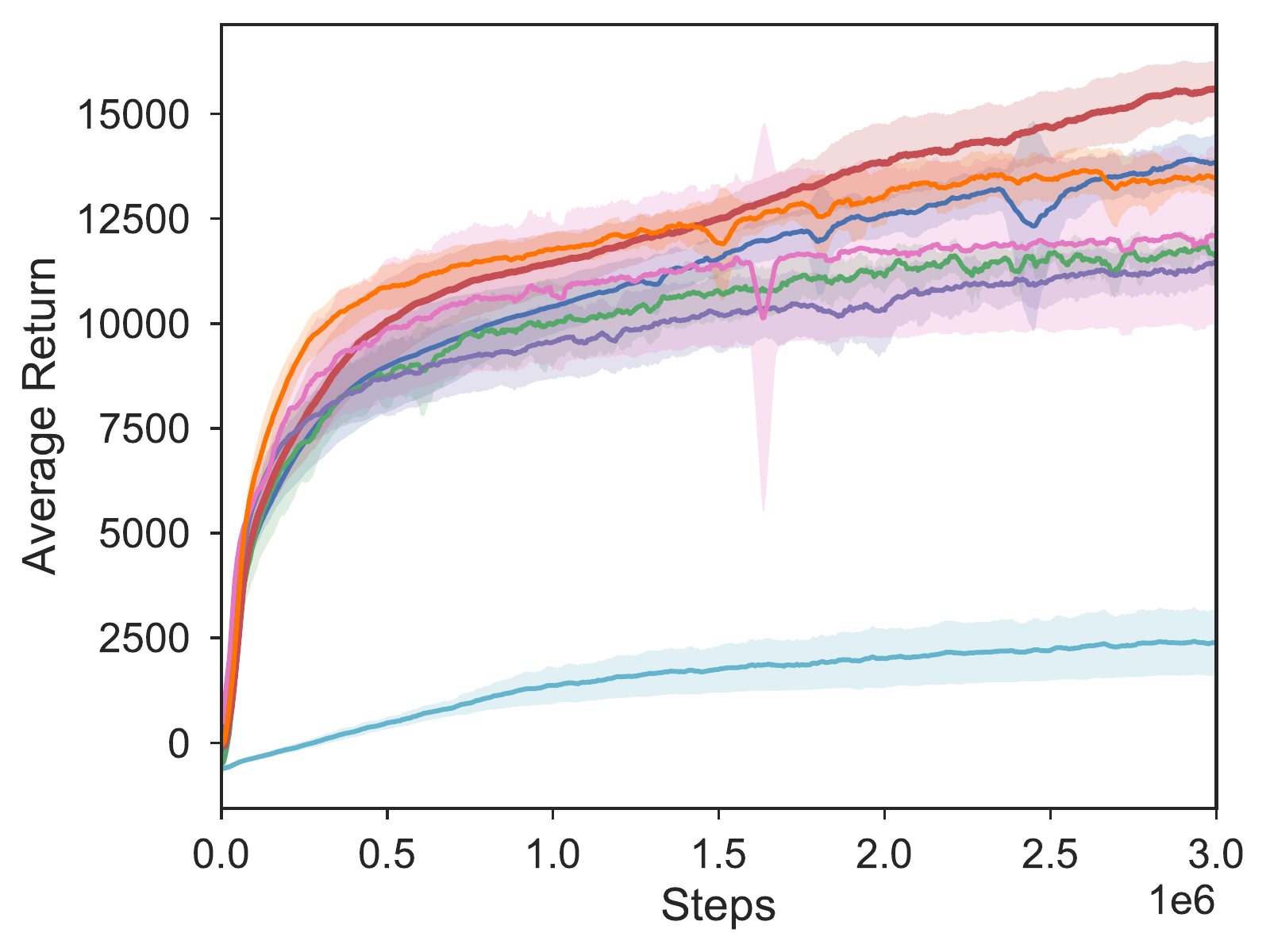}
		}
		\subfigure[Humanoid-v2]{
			\includegraphics[width=0.240\linewidth]{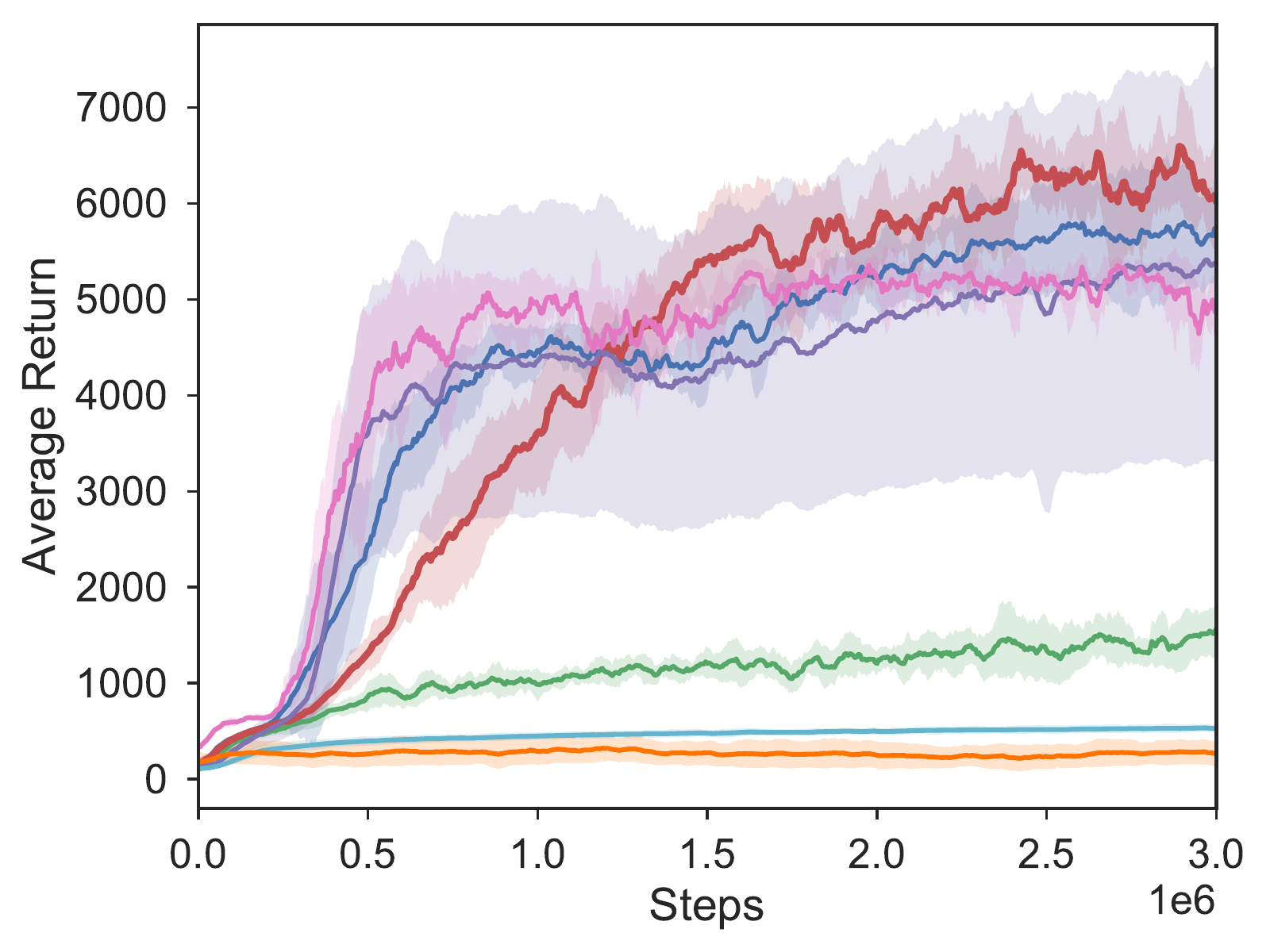}
		}
		\subfigure[Walker2d-v2]{
			\includegraphics[width=0.240\linewidth]{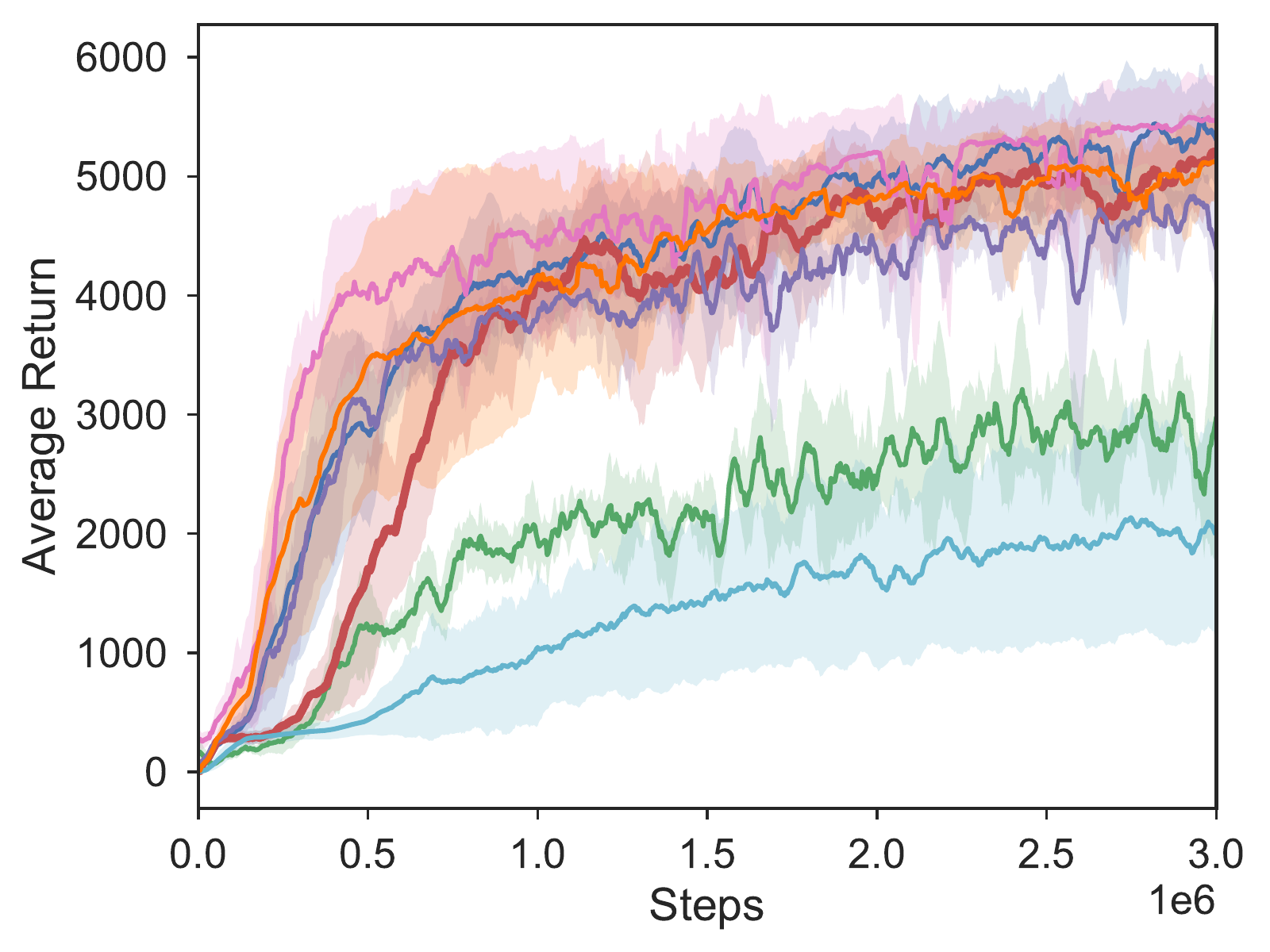}
		}
		\subfigure[finger-spin]{
			\includegraphics[width=0.240\linewidth]{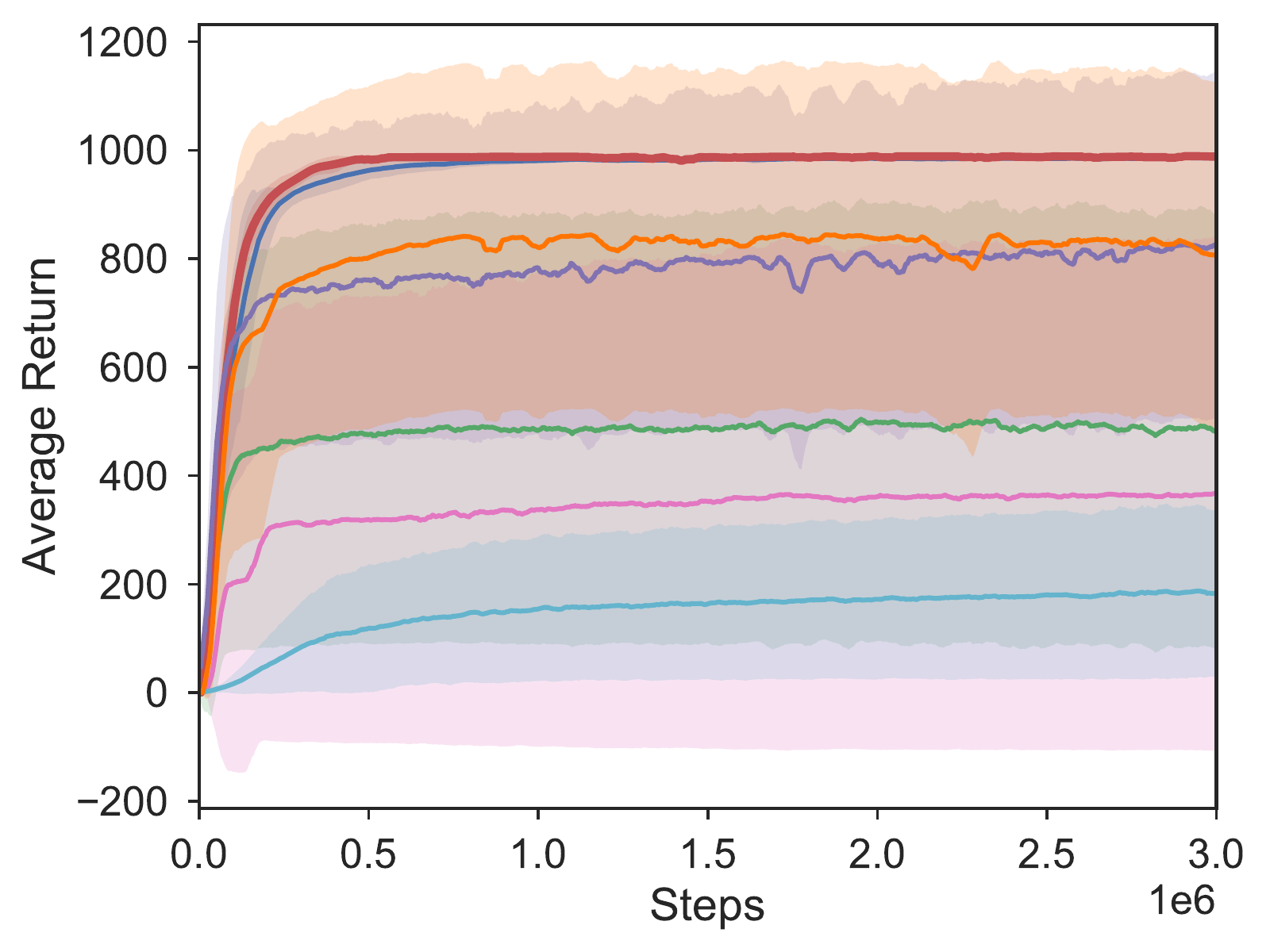}
		}
		\subfigure[point\_mass-easy]{
			\includegraphics[width=0.240\linewidth]{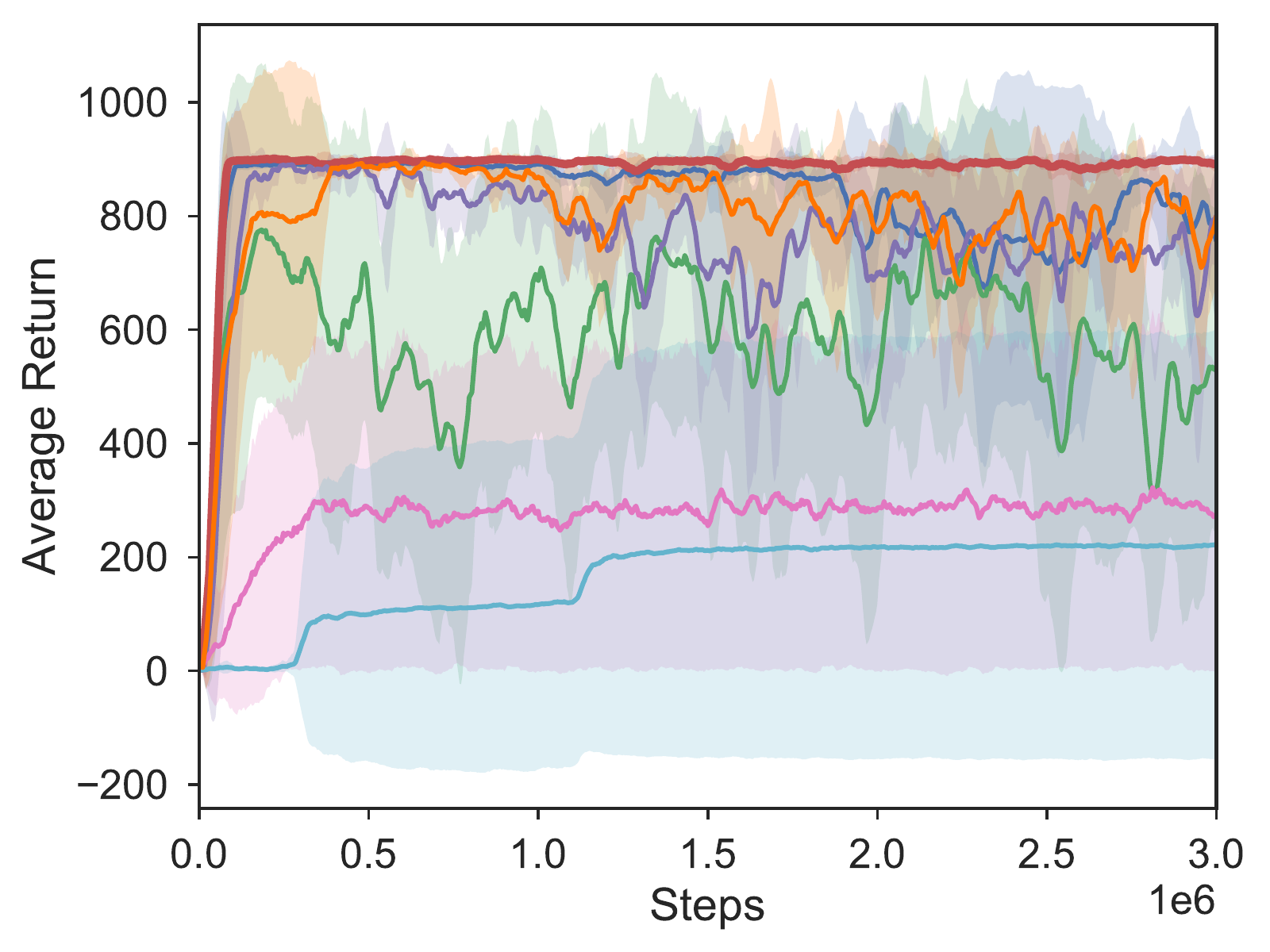}
		}
		\subfigure[quadruped-run]{
			\includegraphics[width=0.240\linewidth]{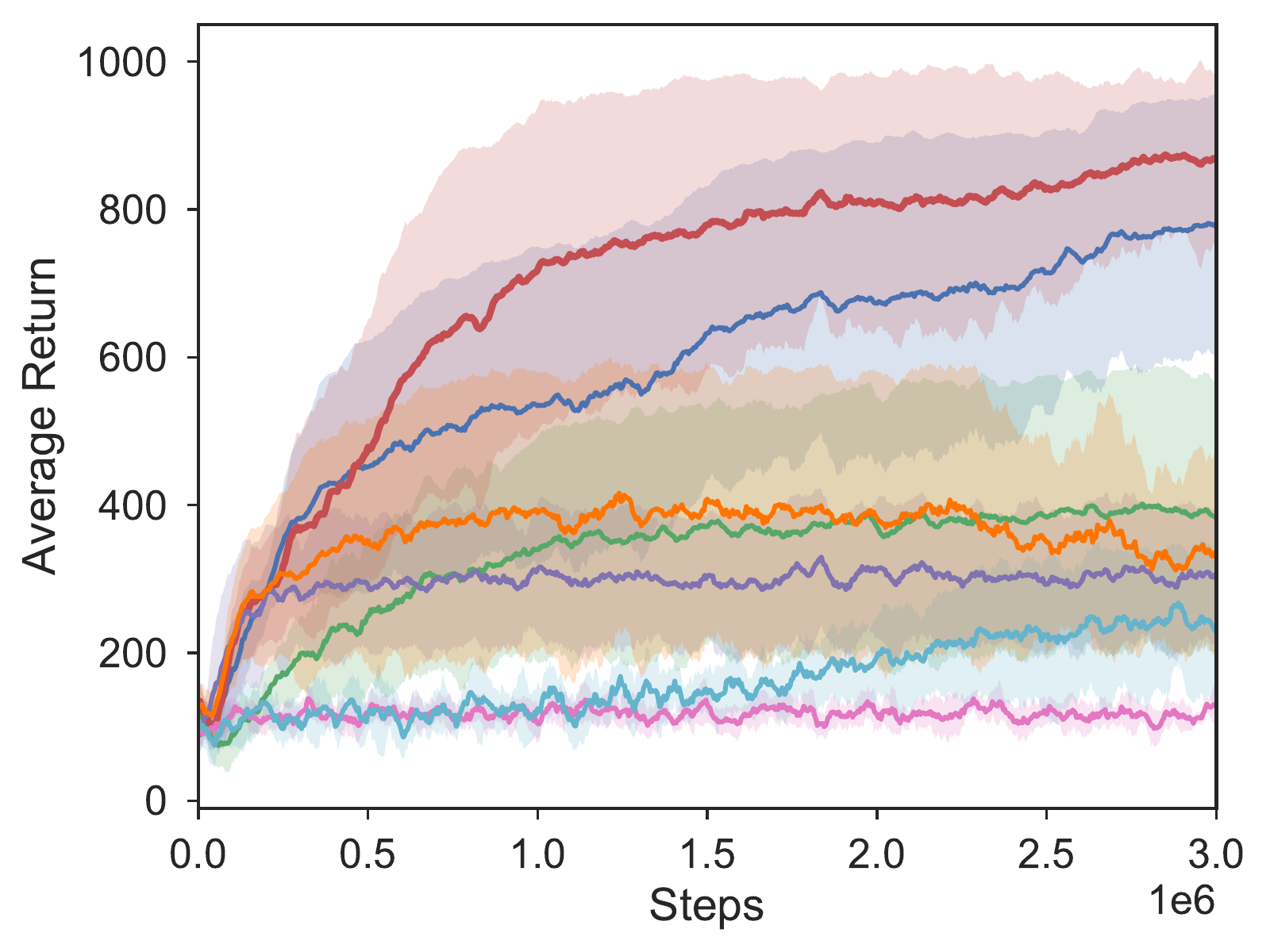}
		}
		\subfigure[swimmer-swimmer6]{
			\includegraphics[width=0.240\linewidth]{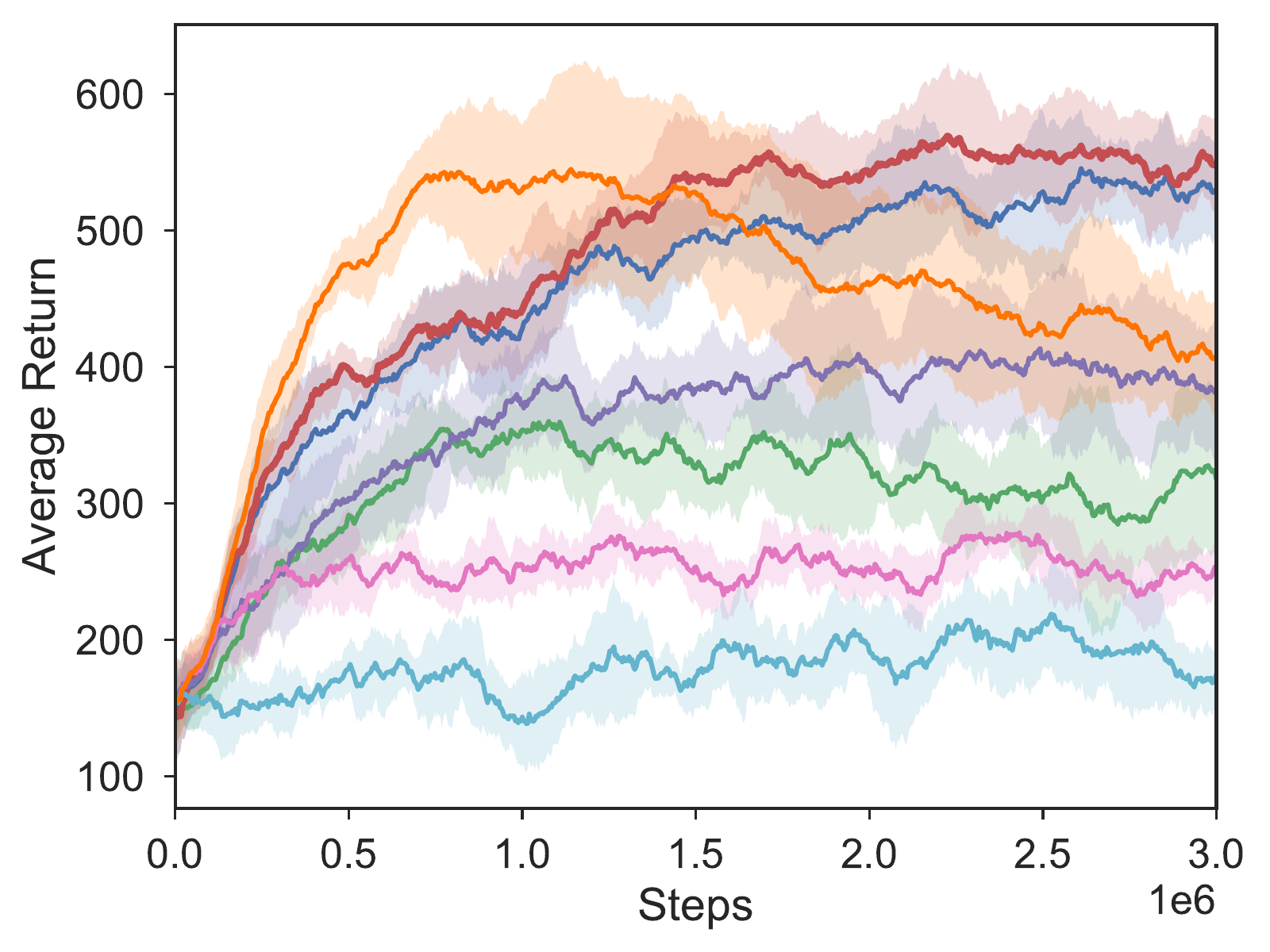}
		}
		\subfigure[walker-run]{
			\includegraphics[width=0.240\linewidth]{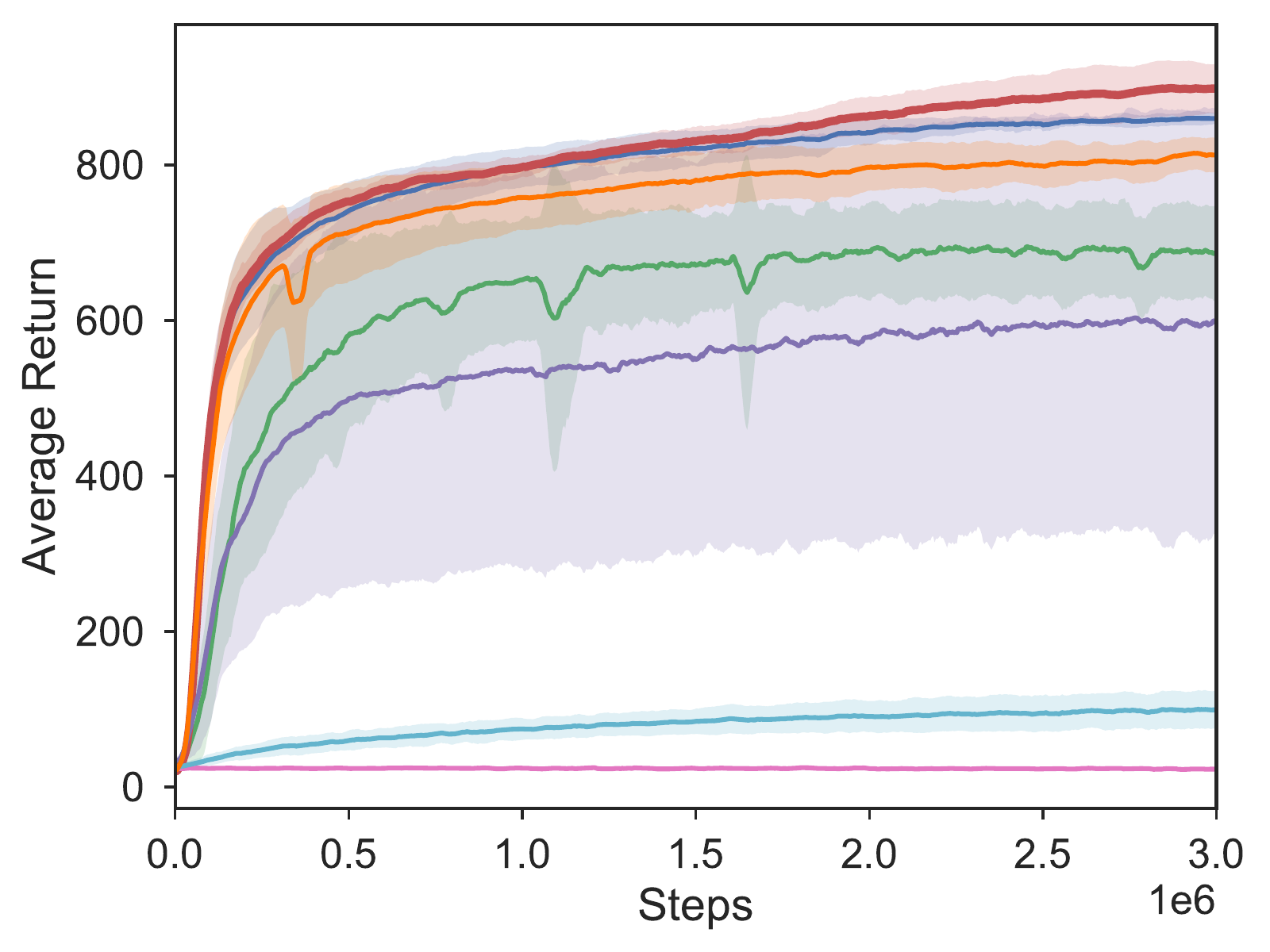}
		}
		\hfill
		\subfigure{
			\includegraphics[width=0.55\linewidth]{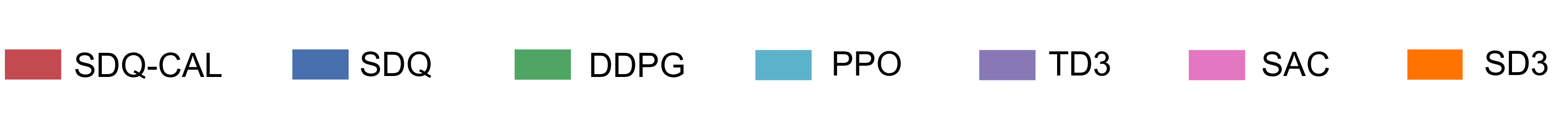}
		}
		\caption{Learning curves for SDQ-CAL, SDQ, DDPG, PPO, TD3, SAC, and SD3 on continuous control benchmarks. SDQ-CAL (red) performs consistently across all environments and outperforms both on-policy and off-policy algorithms in the most challenging tasks.}
		\label{fig:figure3}
	\end{figure*}

	\begin{algorithm}[htbp]
		\caption{SDQ-CAL for Actor-Critic}  
		\begin{algorithmic}[1] 
			\State Initialize critic networks  $Q_{\theta_1}$, $Q_{\theta_2}$ with random parameters $\theta_1$, $\theta_2$.
			\State Initialize actor networks $\pi_{\phi_1}$, $\pi_{\phi_2}$ with random parameters $\phi_1$, $\phi_2$.
			\State Initialize target networks $\theta^\prime_1 \leftarrow \theta_1$, $\theta^\prime_2 \leftarrow \theta_2$.
			\State Initialize replay buffer $\mathcal{B}$.
			\For{t = $1$ \textbf{to} $T$}
			\State Select action $a_t$ according to Eq.~\eqref{actionselection1} and Eq.~\eqref{actionselection2}.
			\State Observe next state $s_{t+1}$ and reward $r_t$.
			\State Store transition tuple $(s_t, a_t, r_t, s_{t+1})$ in $\mathcal{B} $.
			\State Sample minibatch $\begin{aligned} \{(s_j, a_j, r_j, s_{j+1})\}_{j = 1}^N \sim \mathcal{B} \end{aligned}$.
			\State Set $r_{j, 1}$, $r_{j, 2}$ according to Eq.~\eqref{deepCAL}.
			\State Set $\begin{aligned}	y_{j,1} = r_{j,1} + \gamma Q_{\theta_{2}^{\prime}}\left(s_{j+1}, \pi_{\phi_{1}}\left(s_{j+1}\right)\right) \end{aligned}$.
			\State Set $\begin{aligned}	y_{j,2} = r_{j,2} + \gamma Q_{\theta_{1}^{\prime}}\left(s_{j+1}, \pi_{\phi_{2}}\left(s_{j+1}\right)\right) \end{aligned}$.
			\State // Gradient Update.
			\State Update the critics by minimizing the loss: 
			\State \quad $\theta_i \leftarrow \theta_i - \lambda_{Q_i} \nabla_{\theta_i} J_{Q}(\theta_i)$.
			\State Update the actors by the deterministic policy gradient:
			\State \quad $\phi_i \leftarrow \phi_i + \lambda_{\pi_i} \nabla_{\phi_i} J_{\pi}(\phi_i)$.
			\State Update target networks: $\theta^\prime_i \leftarrow \tau \theta_i+(1-\tau) \theta^\prime_i$.
			\EndFor
		\end{algorithmic}
		\label{alg:algorithm3}
	\end{algorithm}
	
	\section{Experiments} \label{experiments}
	In this section, we seek to answer the following questions:
	\begin{itemize}
		\item Does SDQ-CAL achieve better performance and less biased value estimation than existing state-of-the-art model-free RL algorithms, such as DDPG, PPO, TD3, SAC, and SD3?
		\item Can SDQ-CAL attain higher sample efficiency?
		\item How does $\beta$ affect the performance of SDQ-CAL?
		\item What is the contribution of each technique in SDQ-CAL?
	\end{itemize}

	\begin{table}
		\centering
		\caption{Detailed setup for environments.}
		\begin{tabular}{l|ccccc}
			\toprule[1.0pt]
			\textbf{Environment}  & \textbf{State dim.} & \textbf{Action dim.}  \\
			\midrule[1.0pt]
			Ant-v2 & 111 & 8 \\
			HalfCheetah-v2 & 17 & 6 \\
			Humanoid-v2 & 376 & 17  \\
			Walker2d-v2 & 17 & 6 \\
			finger-spin & 12 & 2 \\
			point\_mass-easy & 4 & 2 \\
			quadruped-run & 78 & 12 \\
			swimmer-swimmer6 & 25 & 5 \\
			walker-run & 24 & 6 \\
			\bottomrule[1.0pt]
		\end{tabular}
		\label{table:table7}
	\end{table}

	\begin{table*}[htbp]
		\centering
		\caption{Average return over the last $10$ evaluations of $3$ million time steps. The maximum value for each task is bolded.}
		\begin{tabular}{lcccccccc}
			\toprule[1.0pt]
			\textbf{Environment} & \textbf{SDQ} & \textbf{DDPG} & \textbf{PPO} & \textbf{TD3} & \textbf{SAC}  &\textbf{SD3}  & \textbf{SDQ-CAL} \\
			\midrule[1.0pt]
			Ant-v2  & 5797.67 & 1270.54 & 1557.12 & 5718.16 & 4186.11 & 4759.79 & \textbf{6642.11} \\
			HalfCheetah-v2 & 13720.51 & 11754.14 & 2401.22 & 11302.26 & 12386.36 & 13394.96 & \textbf{15511.98} \\
			Humanoid-v2 & 5642.68 & 1672.60 & 528.70 & 5163.43 & 4972.98 & 287.67 & \textbf{6682.16} \\
			Walker2d-v2 & 5241.98 & 2881.93 & 1887.68 & 4967.00 & \textbf{5530.00} & 5105.69 & 5381.55 \\
			finger-spin & 986.53 & 481.33 & 184.94 & 822.93 & 368.65 & 796.11 & \textbf{987.65} \\
			point\_mass-easy & 696.21 & 429.48 & 221.27 & 823.81 & 284.03 & 743.12 & \textbf{899.48}\\
			quadruped-run & 783.46 & 385.97 & 204.04 & 289.19 & 97.01 & 323.15 & \textbf{862.09} \\
			swimmer-swimmer6 & 542.04 & 334.27 & 226.42 & 350.86 & 223.11 & 452.18 & \textbf{564.96} \\
			walker-run & 859.38 & 688.28 & 96.60 &  603.20 & 23.81 & 819.24 & \textbf{900.17} \\
			\bottomrule[1.0pt]
		\end{tabular}
		\label{table:table1}
	\end{table*}
	
	We compare our method to DDPG, PPO, TD3, SAC, and SD3 on a range of challenging continuous control tasks from OpenAI Gym~\cite{brockman2016openai} and DeepMind Control Suite~\cite{tassa2018deepmind}. For a fair comparison, we make no special modifications to the original environments or reward functions. The performance of algorithms under each environment is demonstrated by plotting the mean cumulative rewards. For the plots, the shaded regions represent the standard deviation of the average evaluation over $6$ different random seeds, and the solid lines represent the mean cumulative rewards.
	For convenience of reproducing our results, we release our source code in  GitHub\footnote{\url{https://github.com/LQNew/SDQ-CAL}}.
	
	\begin{table*}[htbp]
		\centering
		\caption{Sample efficiency comparison. The number of time steps for each algorithm to reach the final performance of SD3 are normalized based on the total environment steps, \emph{i.e.,} $3$ million time steps. The minimum value for each task is bolded.}
		\begin{tabular}{ccccccccc}
			\toprule[1.0pt]
			& Ant-v2 & HalfCheetah-v2 & Walker2d-v2 & finger-spin & point\_mass-easy & quadruped-run  & walker-run \\
			\midrule[1.0pt]
			\textbf{SD3} & 1.00x & 1.00x & \textbf{1.00x} & 1.00x & 1.00x & 1.00x & 1.00x \\
			\textbf{SDQ-CAL} & \textbf{0.23x} & \textbf{0.60x} & \textbf{1.00x} & \textbf{0.07x} & \textbf{0.17x} & \textbf{0.12x} & \textbf{0.33x} \\
			\bottomrule[1.0pt]
		\end{tabular}
		\label{table:table4}
	\end{table*}
	
	\subsection{Implementation Details}
	We select nine well-known benchmark continuous tasks (\emph{i.e.,} Ant-v2, HalfCheetah-v2, Humanoid-v2, Walker2d-v2, finger-spin, point\_mass-easy, quadruped-run, swimmer-swimmer6, and walker-run) available from OpenAI Gym and DeepMind Control Suite to compare our method with existing state-of-the-art algorithms. 
	The detailed descriptions and complexity of the selected tasks are shown in Table~\ref{table:table7}.

	\textbf{Hyper-parameters.}
	For existing model-free RL algorithms (DDPG, PPO, TD3, SAC, and SD3), both policy-network and value-network are represented using multilayer perceptron (MLP) with two hidden layers ($256, 256$). 
	During the training process, the network parameters are optimized using Adam~\cite{kingma2014adam} with a learning rate of $3 \times 10^{-4}$ and the batch size is $256$. 
	The agents are run for $3$ million time steps with evaluations every $5000$ time steps, where each evaluation records the average reward over $10$ episodes. 
	For a fair comparison, we follow the identical setup employed in previous works~\cite{lillicrap2015continuous,fujimoto2018addressing,haarnoja2018soft,schulman2017proximal,pan2020softmax}.
	
	\textbf{SDQ-CAL.}
	The only specific hyper-parameter of SDQ-CAL is $\beta$. We do a coarse grid-search for $\beta$ over the set $\{0.009, 0.019, 0.05, 0.1, 0.5, 0.9\}$. Finally, for all the experiments, we set $\beta = 0.019$ and $\gamma=0.98$. 
	
	\textbf{SDQ.}
	We implement SDQ as the baseline and apply it to the continuous control tasks. 
	Specifically, in SDQ, we do not modify the reward function for performing conservative Advantage Learning anymore. Other than that, the rest of the algorithm flow is the same as SDQ-CAL.
	With the simple modification provided by SDQ-CAL, we get the performance of SDQ across all environments.
	
	\begin{figure}[htbp]
		\centering
		\small
		\subfigure{
			\includegraphics[width=1.0\linewidth]{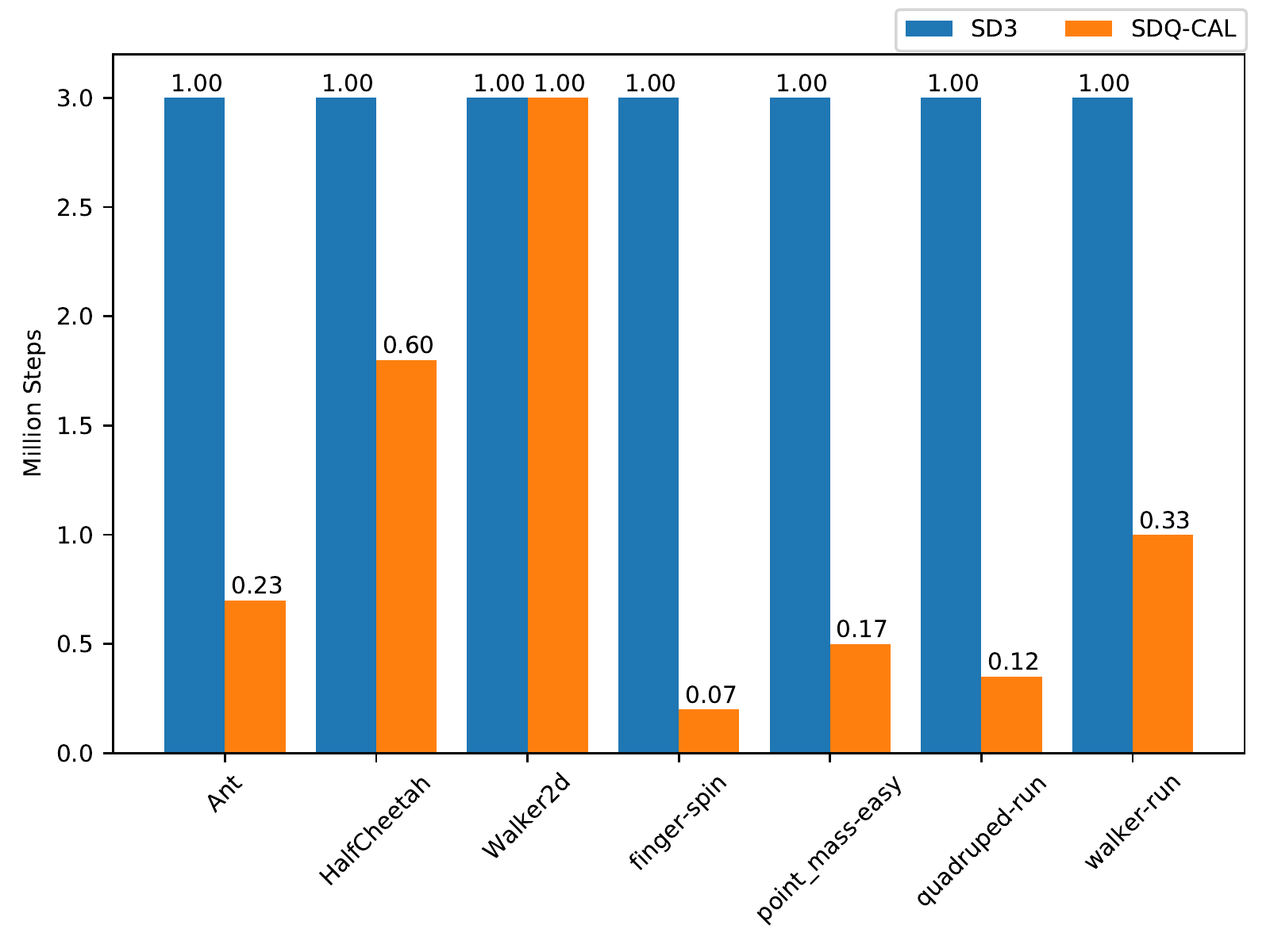}
		}
		\vspace{-0.4cm}
		\caption{Sample efficiency comparison. The normalized steps are plotted above the bars. Since SD3 performs extremely poorly on Humanoid-v2 and shows a rapid decline in the later stages of training process on swimmer-simmer6, we omit these two environments for the compactness of the plot.}
		\label{fig:figure4}
	\end{figure}
	
	\subsection{Comparison with State-of-the-art Algorithms}
	We present the final performance on $3$ million time steps in Table~\ref{table:table1} and the training curves of each algorithm in Fig.~\ref{fig:figure3}. 
	As the results show, overall, SDQ-CAL performs comparably to the existing state-of-the-art methods in the easier tasks and outperforms them in the more challenging tasks with a large margin, both in terms of sample efficiency and the final performance. 
	To give a more intuitive illustration of sample efficiency, Fig.~\ref{fig:figure4} and Table~\ref{table:table4} show the number of time steps for SD3~\cite{pan2020softmax} and SDQ-CAL to reach the final performance of SD3. The number of time steps  are normalized based on the total environment steps, \emph{i.e.,} $3$ million time steps.
	Here, we choose SD3 as the baseline algorithm for contrast since we notice that SD3 matches or outperforms other baseline algorithms across most tasks. 
	We observe that SDQ-CAL learns efficiently and takes much fewer time steps to achieve the same performance as SD3.
	
	It is worth noting that SDQ can also learn on all tasks and even achieves better performance than most existing RL algorithms (\emph{i.e.}, DDPG, PPO, TD3, SAC, and SD3). 
	But as demonstrated in Fig.~\ref{fig:figure3} and Table~\ref{table:table1}, SDQ learns slower than SDQ-CAL and has the worse asymptotic performance. 
	This phenomenon verifies that conservative Advantage Learning can improve both performance and sample efficiency over SDQ.
	
	\begin{figure*}[htbp]
		\centering
		\small
		\subfigure[Learning curves of \textbf{SDQ-CAL} with $\gamma=0.98$ and different $\beta$.]{
			\begin{minipage}[b]{0.90\linewidth}
				\centering
				\includegraphics[width=0.300\linewidth]{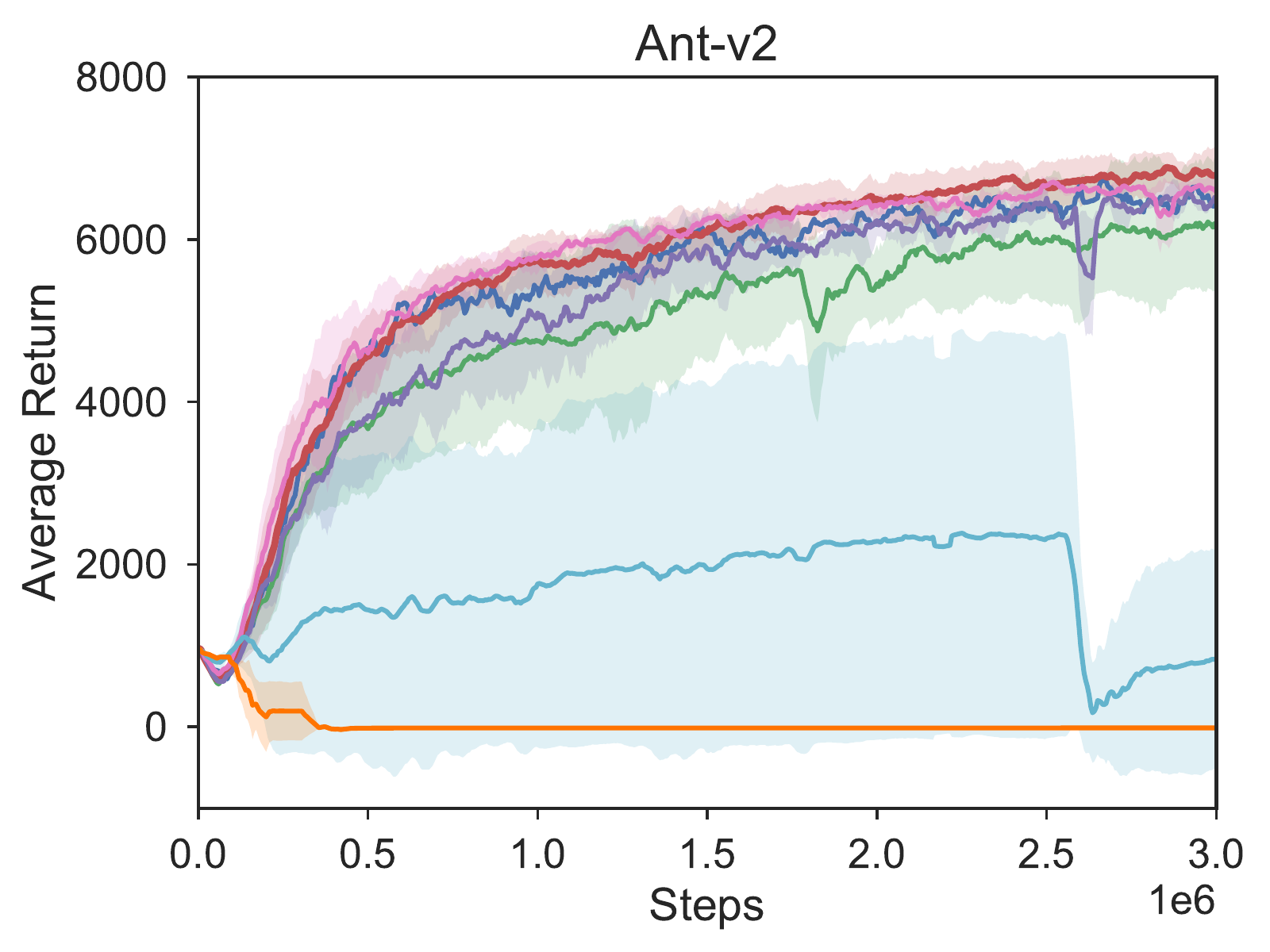}
				\includegraphics[width=0.300\linewidth]{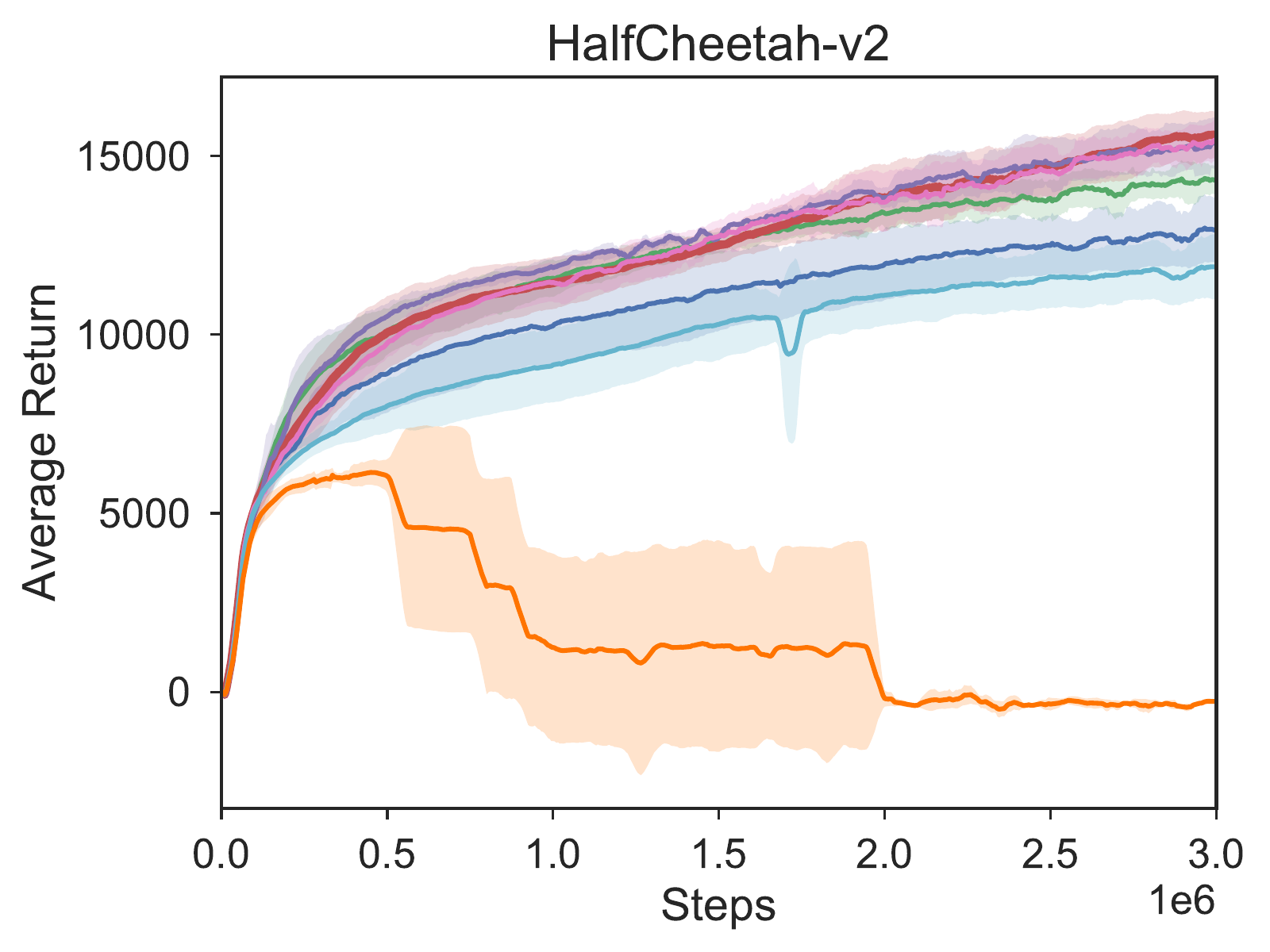}
				\includegraphics[width=0.300\linewidth]{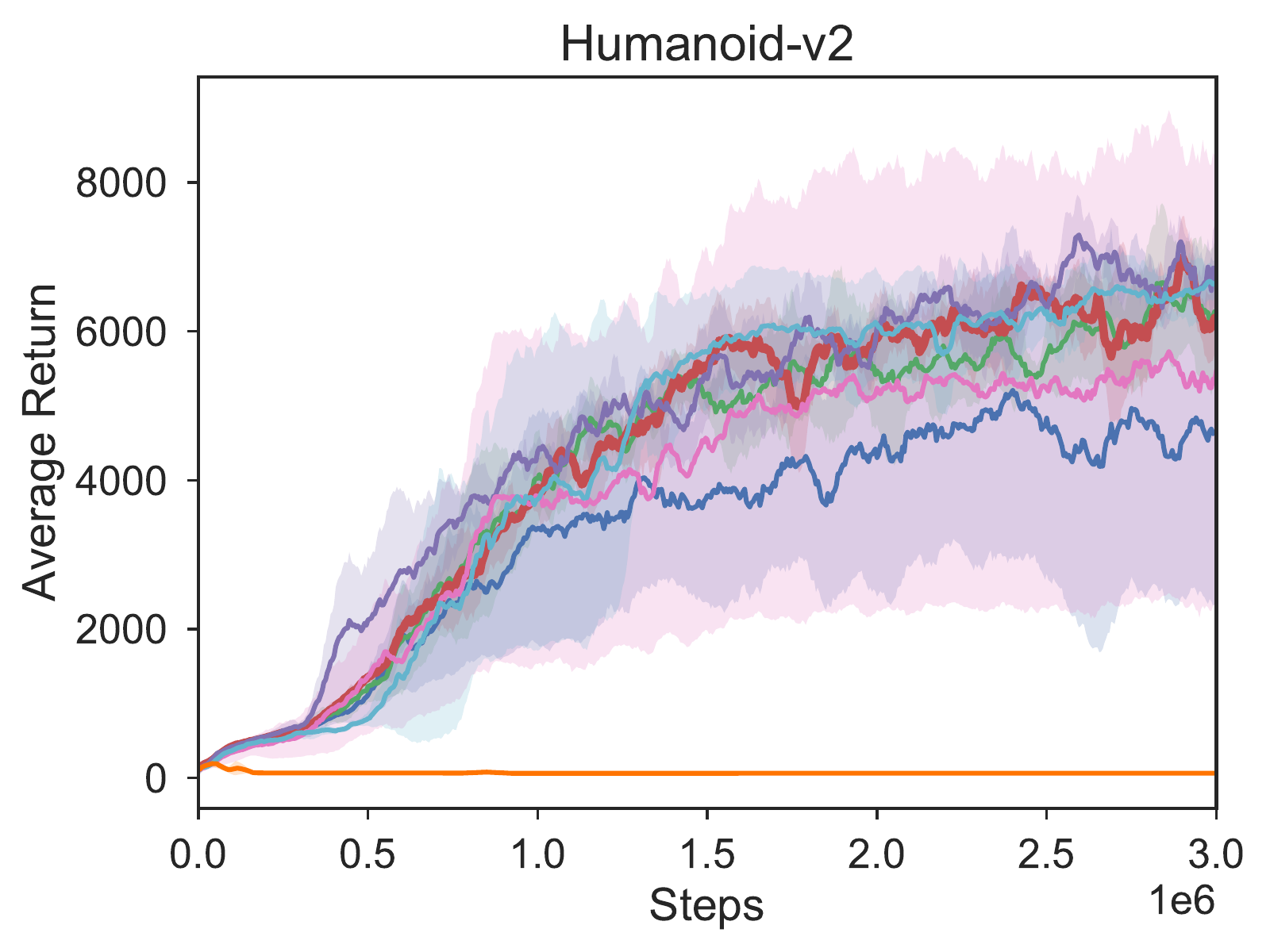}
			\end{minipage}
		}
		\subfigure[Learning curves of \textbf{SDQ-AL} with $\gamma=0.98$ and different $\beta$.]{
			\begin{minipage}[b]{0.90\linewidth}
				\centering
				\includegraphics[width=0.300\linewidth]{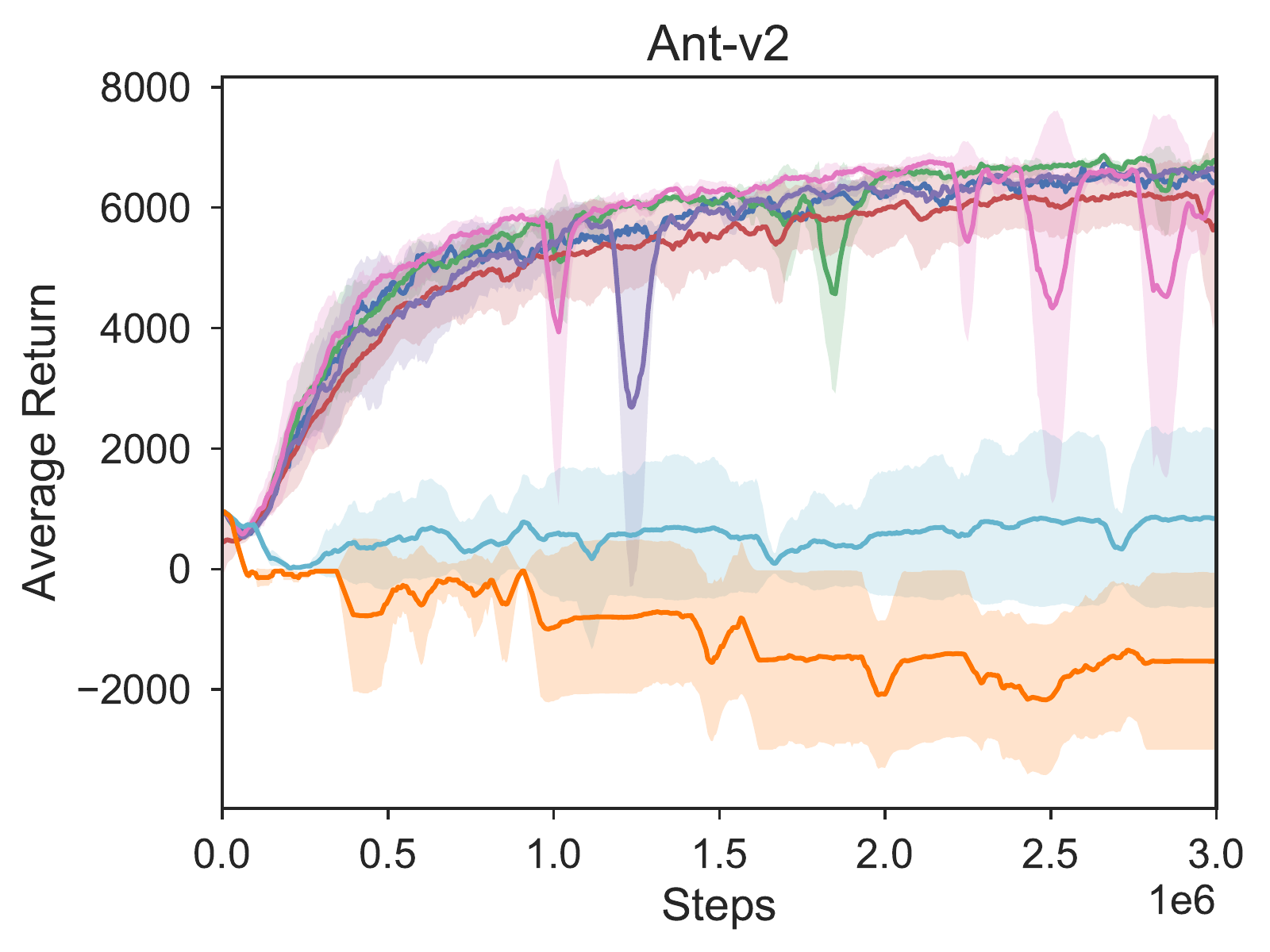}
				\includegraphics[width=0.300\linewidth]{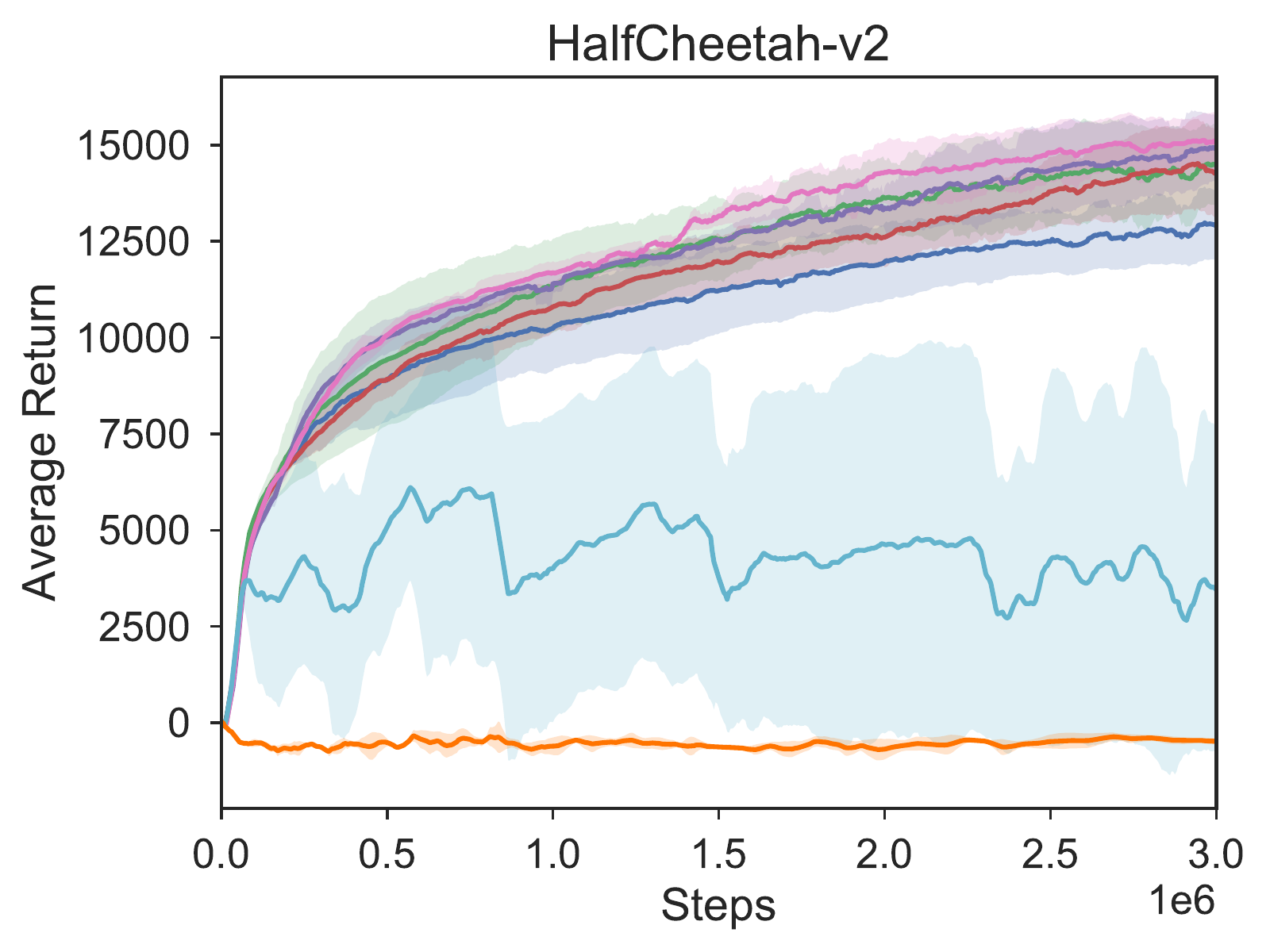}
				\includegraphics[width=0.300\linewidth]{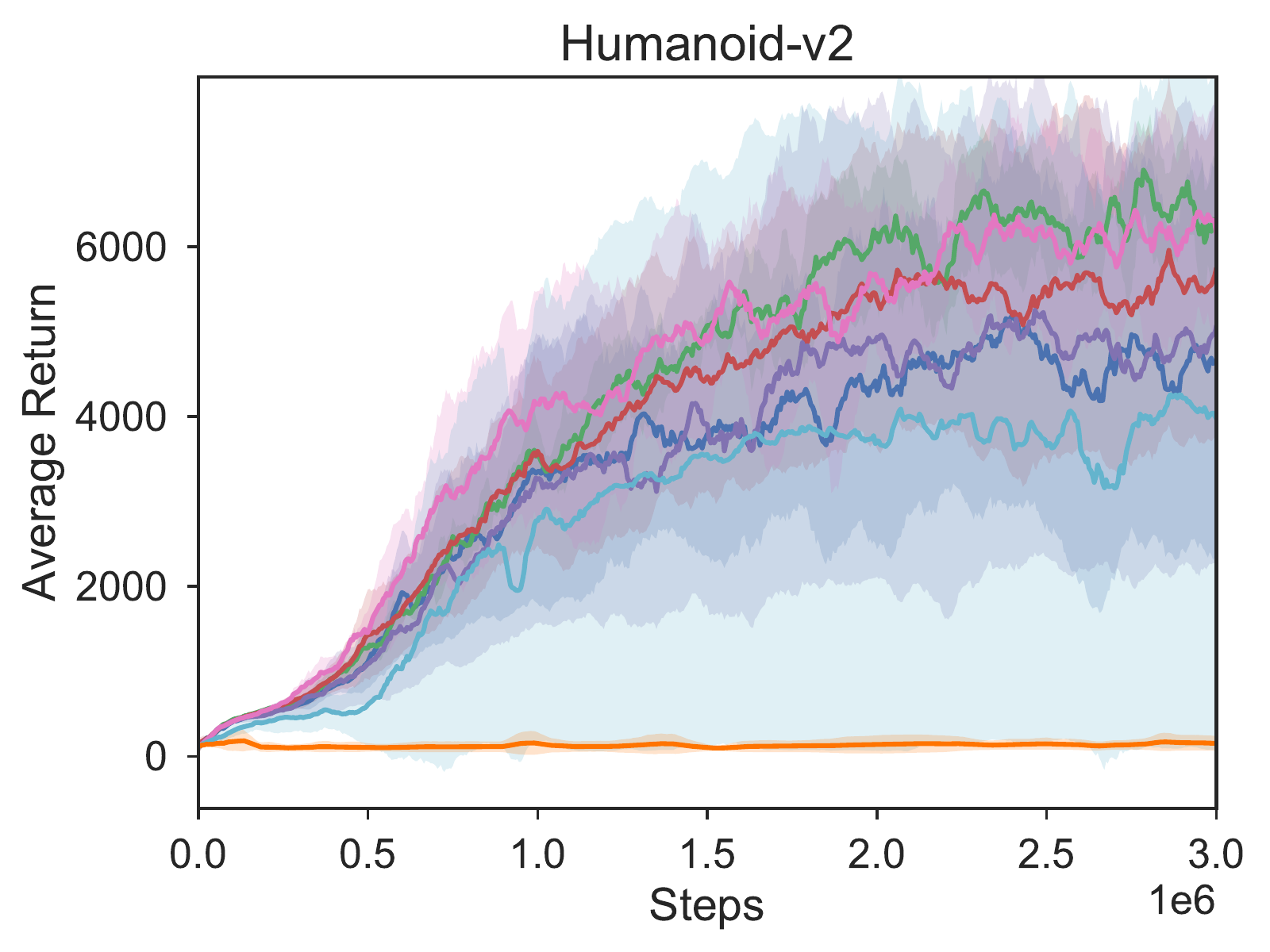}
			\end{minipage}
		}
		\hfill
		\subfigure{
			\includegraphics[width=0.55\linewidth]{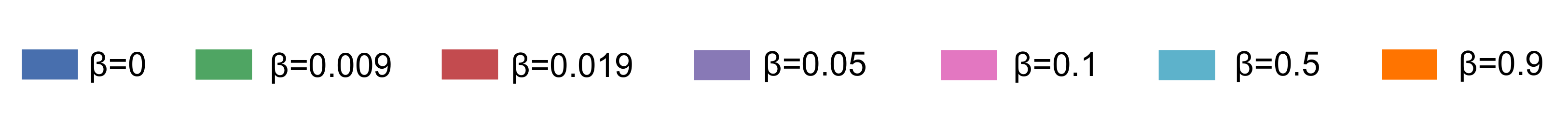}
		}
		\caption{Ablation study of parameter $\beta$ on Ant-v2, HalfCheetah-v2, and Humanoid-v2.}
		\label{fig:figure7}
	\end{figure*}
	
	\begin{table}[htbp]
		\centering
		\caption{Final scores of SDQ-CAL with different $\beta$. We set $\gamma = 0.98$. }
		\begin{tabular}{lcccccccc}
			\toprule[1.0pt]
			& Ant-v2 & HalfCheetah-v2 & Humanoid-v2 \\
			\midrule[1.0pt]
			\textbf{$\beta=0$} & 5853.46 & 12972.31 & 5332.66 \\
			\textbf{$\beta=0.009$} & 5992.82 & 14990.92 & 5955.49 \\
			\textbf{$\beta=0.019$} & 6642.11 & \textbf{15511.98} & \textbf{6682.16} \\
			\textbf{$\beta=0.05$} & 6421.17 & 15292.44 & 5421.52 \\
			\textbf{$\beta=0.1$} & \textbf{6752.75} & 15468.46 & 5433.45 \\
			\textbf{$\beta=0.5$} & 822.18 & 11865.60 & 6591.10  \\
			\textbf{$\beta=0.9$} & -15.04 & -339.86 & 65.75  \\
			\bottomrule[1.0pt]
		\end{tabular}
		\label{table:table5}
	\end{table}
	
	\begin{table}[htbp]
		\centering
		\caption{Final scores of SDQ-AL with different $\beta$. We set $\gamma = 0.98$.}
		\begin{tabular}{lcccccccc}
			\toprule[1.0pt]
			& Ant-v2 & HalfCheetah-v2 & Humanoid-v2 \\
			\midrule[1.0pt]
			\textbf{$\beta=0$} & 5853.46 & 12972.31 & 5332.66 \\
			\textbf{$\beta=0.009$} & \textbf{6381.66} & 14902.86 & \textbf{5926.61} \\
			\textbf{$\beta=0.019$} & 5969.92 & 14397.36 & 5913.09 \\
			\textbf{$\beta=0.05$} & 6356.26 & 14970.94 & 5316.02 \\
			\textbf{$\beta=0.1$} & 6130.91 & \textbf{15250.61} & 5757.32 \\
			\textbf{$\beta=0.5$} & 879.04 & 3309.51 & 4274.79 \\
			\textbf{$\beta=0.9$} & -1563.84 & -463.63 & 147.99 \\
			\bottomrule[1.0pt]
		\end{tabular}
		\label{table:table6}
	\end{table}
	
	\subsection{Impact of \texorpdfstring{$\beta$}{Lg}}
	Table~\ref{table:table5},~\ref{table:table6} and Fig.~\ref{fig:figure7} show the ablation study of parameter $\beta$ over the set $\{0.009, 0.019, 0.05, 0.1, 0.5, 0.9\}$ in MuJoCo environments~\cite{todorov2012mujoco}.
	Specifically, we implement a variant of SDQ-CAL that only performs Advantage Learning~\cite{bellemare2016increasing} to further explore the impact of $\beta$.
	In SDQ-CAL or simultaneous Double Q-learning with Advantage Learning (SDQ-AL), $\beta$ determines how much the value of conservative advantage or advantage added to the reward, with $\beta$ approaching 0, corresponding to the original reward without any modification, which is also the form of SDQ. 
	From the Fig.~\ref{fig:figure7}, we can find that SDQ-CAL and SDQ-AL both bring performance improvement over SDQ. 
	But compared with SDQ-AL, SDQ-CAL can make the training process more stable and obtain higher performance improvements. 
	It is rather remarkable that SDQ-CAL performs consistently and outperforms SDQ with $\beta \in \{0.009, 0.019, 0.05, 0.1\}$, which means that conservative Advantage Learning is indeed an effective method to improve sample efficiency and performance.
	
	Overall, as shown in Fig.~\ref{fig:figure7}, the performance gradually increases before reaching a certain value (\emph{i.e.,} $\beta = 0.019$ in SDQ-CAL, and $\beta = 0.009$ in SDQ-AL), but after that, the performance even decreases with $\beta$ increasing. 
	This phenomenon also verifies that a larger $\beta$ can make the RL agent more shortsighted so that the return may be reduced and we need to find a balance between sample efficiency and future rewards with a reasonable value of $\beta$. 
	Finally, we set $\beta = 0.019$ for SDA-CAL, and $\beta=0.009$ for SDQ-AL.
	
	\begin{figure*}[htbp]
		\centering
		\small
		\subfigure[The bias of value estimations for DDPG.]{
			\begin{minipage}[b]{0.485\linewidth}
				\centering
				\includegraphics[width=0.494\linewidth]{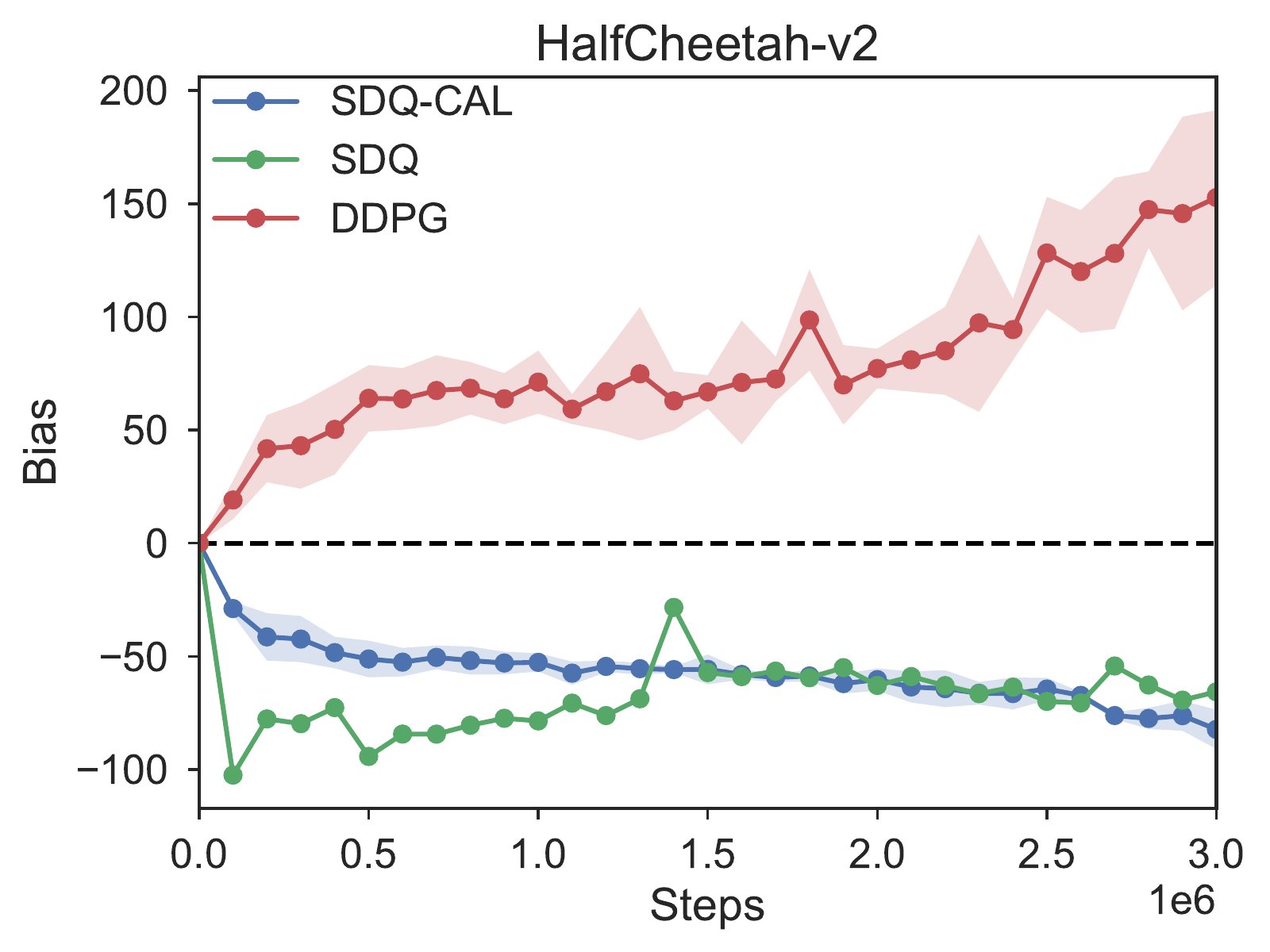}
				\includegraphics[width=0.494\linewidth]{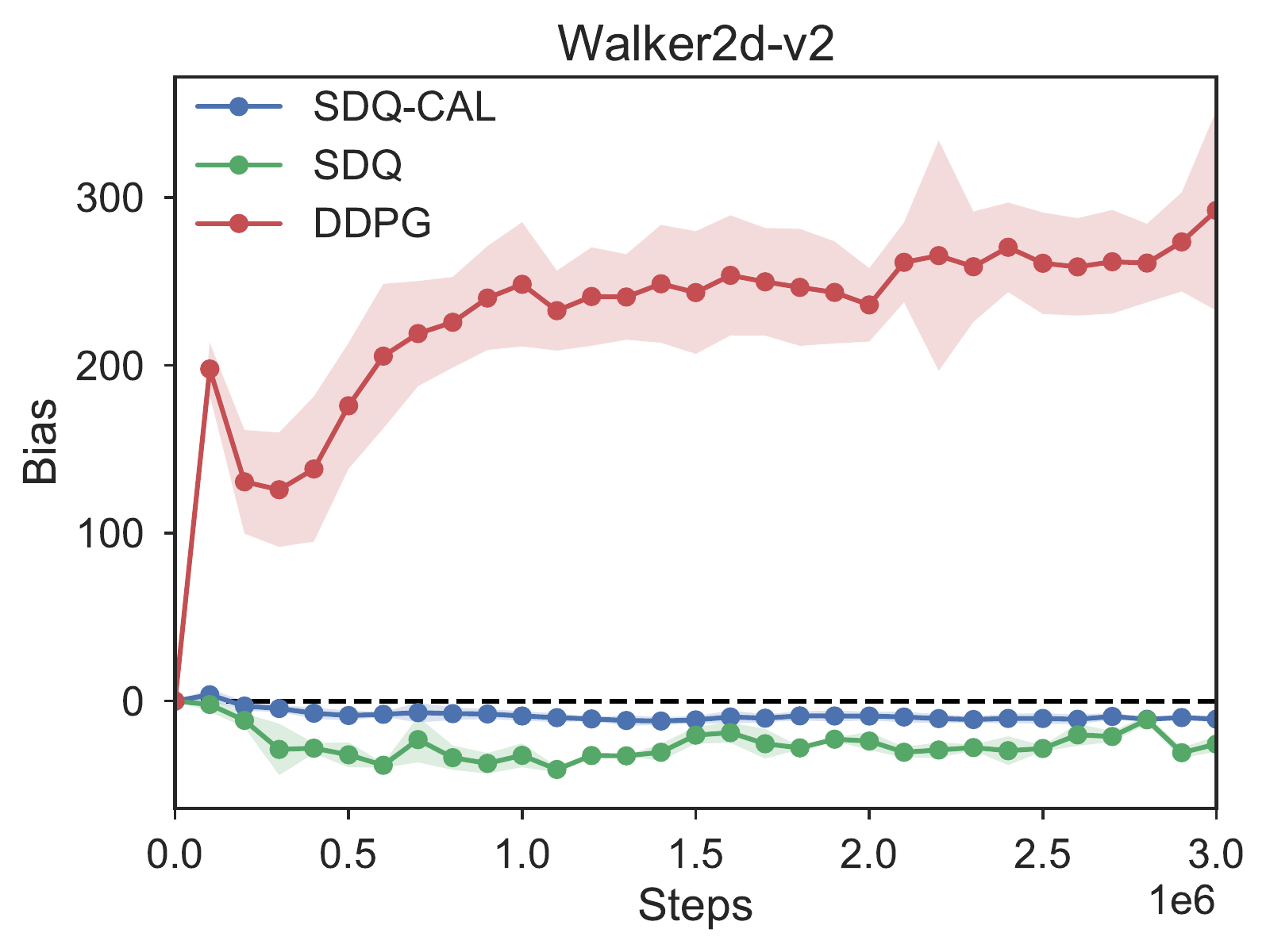}
			\end{minipage}
		}
		\subfigure[The bias of value estimations for TD3 and SD3.]{
			\begin{minipage}[b]{0.485\linewidth}
				\centering
				\includegraphics[width=0.494\linewidth]{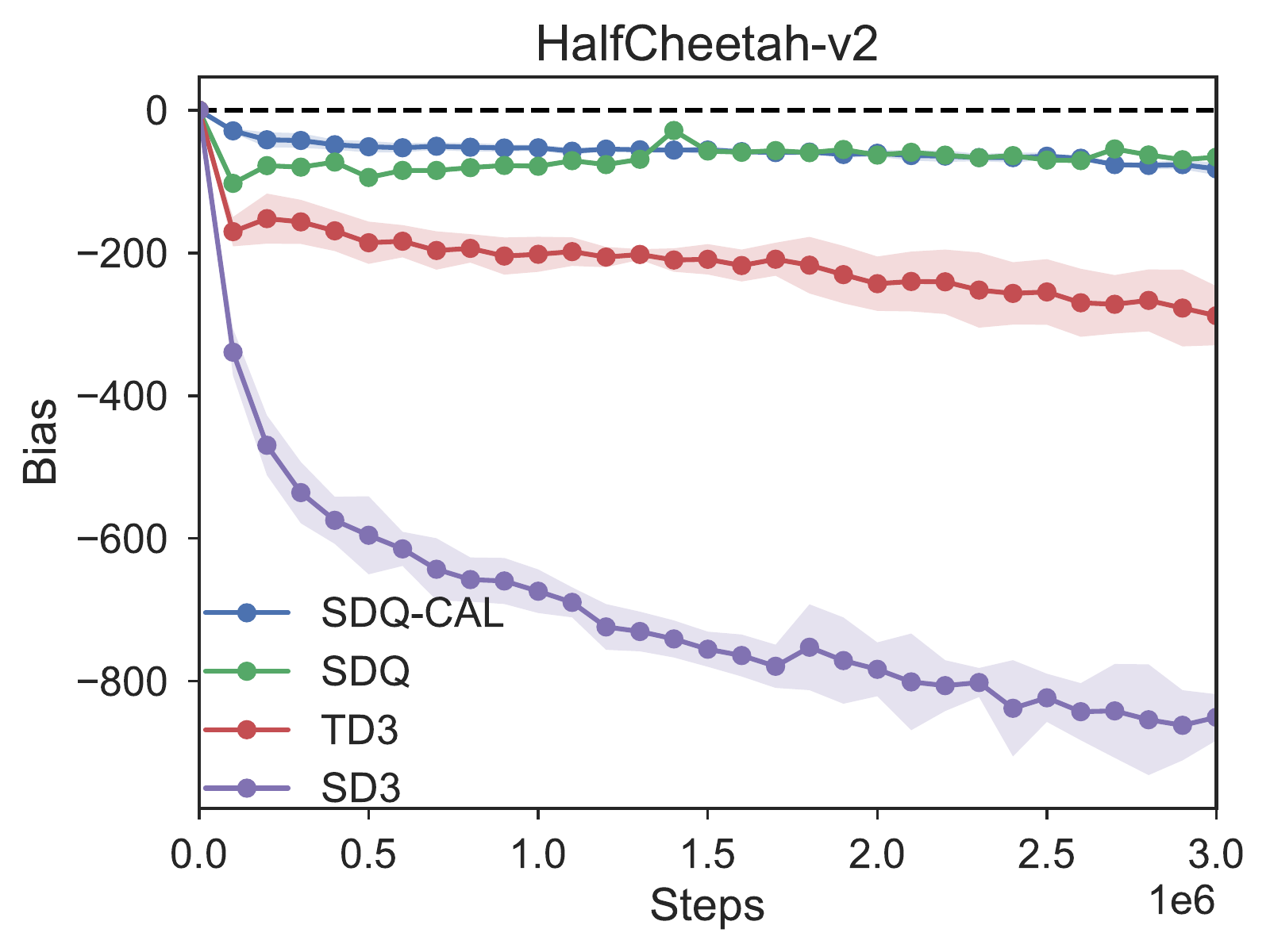}
				\includegraphics[width=0.494\linewidth]{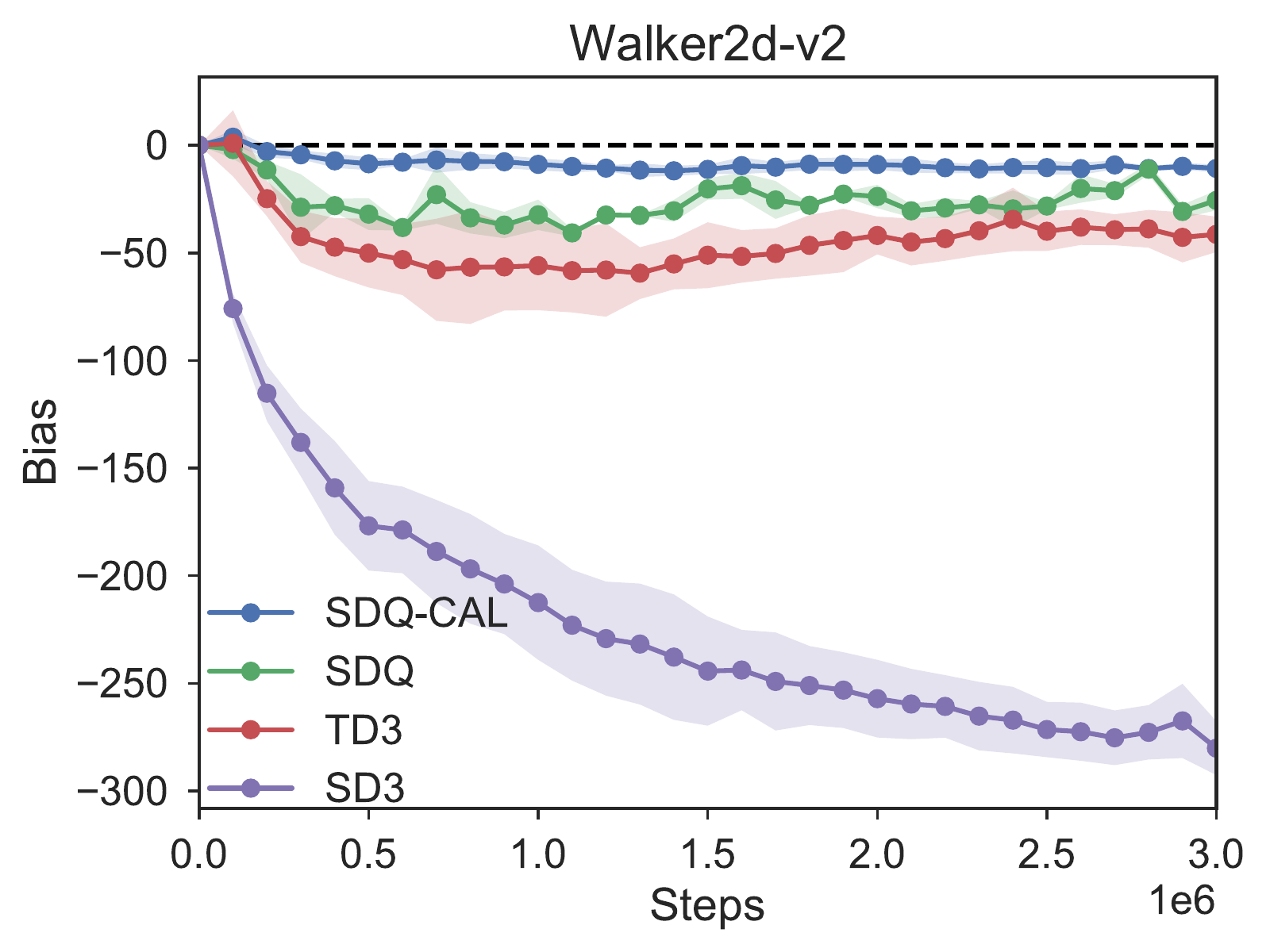}
			\end{minipage}
		}
		\caption{Measuring estimation bias in the value functions of DDPG, TD3, SD3, SDQ, and SDQ-CAL on MuJoCo environments over $3$ million steps.}
		\label{fig:figure2}
	\end{figure*}
	
	\subsection{Q-value Estimation in Actor-Critic} \label{SDQQValueEstimation}
	As discussed in TD3~\cite{fujimoto2018addressing}, overestimation issue occurs in actor-critic RL, such as DDPG~\cite{lillicrap2015continuous}. 
	To avoid this problem, in TD3~\cite{fujimoto2018addressing}, Clipped Double Q-learning uses a lower bound approximation to the critics. However, it suffers a large underestimation bias~\cite{ciosek2019better, pan2020softmax}. 
	In our work, we perform simultaneous Double Q-learning that allows for evenly unbiased Q-value estimates.
	We show that our method can lead to more accurate Q-value estimation by plotting the value estimates of DDPG, TD3, SD3, SDQ, and SDQ-CAL in two MuJoCo~\cite{todorov2012mujoco} environments, \emph{i.e.,} HalfCheetah-v2 and Walker2d-v2. 
	The bias of the corresponding value estimates and the true values is shown in Fig.~\ref{fig:figure2}.
	The value estimates are averaged over $1000$ states sampled from the replay buffer every $10^5$ iterations, which denoted as $\widehat{V(s)}$. The true values are estimated using the average discounted long-term rewards following the current policy, starting from the sampled states. The true values are denoted as $\widehat{G(s)}$. Then, we can obatin the bias of Q-value by calculating $\widehat{V(s)} - \widehat{G(s)}$.
	
	From Fig.~\ref{fig:figure2}, a few trends are readily apparent: i) Compared with DDPG, we can clearly observe that SDQ-CAL significantly reduces overestimation bias and leads to a less biased value estimation. ii) SDQ-CAL achieves a much smaller absolute bias than TD3 and SD3, which also validates that SDQ-CAL can not only reduce overestimation bias but also improve underestimation bias to avoid a large underestimation bias. iii) SDQ-CAL and SDQ both obtain relatively small estimation bias, which means that simultaneous Double Q-learning is the key to obtaining less biased Q-value estimates. Furthermore, the introduction of conservative Advantage Learning for improving sample efficiency does not lead to an increase in underestimation bias. 
	
	\subsection{Evaluation of Components in SDQ-CAL}
	In this section, we conduct a series of experiments to further examine which particular components of SDQ-CAL are essential for the performance. We present our results in Table~\ref{table:table2}, ~\ref{table:table3} and Fig.~\ref{fig:figure13}, ~\ref{fig:figure12}.
	
	\begin{table*}[htbp]
		\centering
		\caption{Ablation study over SDQ, SDQ-AL, DQ-CAL, and SDQ-Pi1. The top half of the table shows the average return over the last $10$ evaluations of $3$ million time steps and the maximum value for each task is bolded. The bottom half of the table shows the sample efficiency compared with SDQ-Pi1. We choose SDQ-Pi1 as the baseline approach for contrast since SDQ-Pi1 always learns slowest. The minimum value for each task is bolded.}
		
		\begin{tabular}{c|l|ccccc}
			\toprule[1.0pt]
			& \textbf{Environment} & \textbf{SDQ} & \textbf{SDQ-AL} & \textbf{DQ-CAL} & \textbf{SDQ-Pi1}  & \textbf{SDQ-CAL} \\
			\midrule[1.0pt]
			& Ant-v2  & 5797.67 & 6381.66 & 6418.00 & 4327 .98 & \textbf{6642.11}\\
			\textbf{Average} & HalfCheetah-v2 & 13720.51 & 14902.86 & 14161.01 & 13160.01 & \textbf{15511.98} \\
			\textbf{Return} & Humanoid-v2 & 5642.68 & 5926.61 & 5964.95 & 3843.89 & \textbf{6682.16} \\
			& Walker2d-v2 & 5241.98 & 5273.24 & 4745.89& 4667.32 & \textbf{5381.55}  \\
			\midrule[1.0pt]
			& Ant-v2 & 0.58x & 0.67x& 0.50x & 1.00x & \textbf{0.33x}  \\
			\textbf{Sample} & HalfCheetah-v2 & 0.92x & 0.92x & 0.95x & 1.00x & \textbf{0.57x} \\
			\textbf{Efficiency} & Humanoid-v2 & 0.58x & 0.50x & 0.50x & 1.00x & \textbf{0.40x} \\
			& Walker2d-v2& 0.58x & 0.53x & 1.00x & 1.00x & \textbf{0.42x}  \\
			\bottomrule[1.0pt]
		\end{tabular}
		\label{table:table2}
	\end{table*}
	
	\subsubsection{Effectiveness of double-action selection} We study the performance of a variant of SDQ-CAL that selects actions with a fixed policy to interact with the environment (SDQ-Pi1). Specifically, SDQ-Pi1 chooses actions only with the first policy $\pi_{\phi_{1}}$:
	\begin{equation}
		\label{actionselection3}
		a_t = \pi_{\phi_{1}}(s_t) + \epsilon,\, \, \, \epsilon \sim \mathcal{N}(0, \sigma).
	\end{equation}
	As shown in Table~\ref{table:table2}, we remark that double-action selection is the key factor for the performance improvement in SDQ-CAL. We find that other proposed techniques coupled with double-action selection (\emph{i.e.,} SDQ, SDQ-AL, DQ-CAL, and SDQ-CAL) outperform SDQ-Pi1 with a large margin on all environments. 
	
	\begin{table}[htbp]
		\centering
		\caption{Ablation study over TD3 and TD3-DAS. The table shows the average return over the last $10$ evaluations of $3$ million time steps and the maximum value for each task is bolded.}
		\begin{tabular}{c|l|cc}
			\toprule[1.0pt]
			& \textbf{Environment}  & \textbf{TD3} & \textbf{TD3-DAS} \\
			\midrule[1.0pt]
			& Ant-v2 & 5718.16 & \textbf{6450.99} \\
			\textbf{Average} & HalfCheetah-v2 & 11302.26 & \textbf{13688.00} \\
			\textbf{Return} & Humanoid-v2 & 5163.43 & \textbf{5507.83} \\
			& Walker2d-v2 & \textbf{4967.00} & 4736.89  \\
			\bottomrule[1.0pt]
		\end{tabular}
		\label{table:table3}
	\end{table}
	
	It is also worth studying the performance of a variant of TD3
	using double-action selection (TD3-DAS). Since TD3 only utilizes a single actor to improve the policy, we propose a simple method that uses the target actor $\pi_{\phi^\prime}$ as another actor. Specifically, in TD3-DAS, double-action selection is computed as: $a_{t}^{\prime}=\left.\arg \max_{a_{t}^{i}}\left(Q_{\theta_{1}}\left(s_{t}, a_{t}^{i}\right)+Q_{\theta_{2}}\left(s_{t}, a_{t}^{i}\right)\right)\right|\tiny{i=1,2}$ , where $a_t^1 = \pi_\phi(s_t)$ and $a_t^2 = \pi_{\phi^\prime}(s_t)$. From Table~\ref{table:table3}, we can observe that DAS brings consistent performance gains to TD3 in most of the environments.
	The results confirm that double-action selection is efficient for obtaining better performance.

	\subsubsection{Conservative Advantage Learning vs. Advantage Learning} 
	Compared with original Advantage Learning~\cite{baird1999gradient}, conservative Advantage Learning takes the pessimistic estimate of value functions for the sampled state-action pairs to expand the action gap between the optimal actions and other actions further. To compare how conservative Advantage Learning affects the performance, we compare to a variant of SDQ-CAL that only performs Advantage Learning (SDQ-AL). Following the settings in the previous sub-section for $\beta$, we set $\beta=0.009$ and $\gamma=0.98$ for SDQ-AL. 
	Table~\ref{table:table2} shows that conservative Advantage Learning achieves better performance while obtaining improved sample efficiency.
	
	\subsubsection{Effectiveness of simultaneous update} We remark that when applying SDQ-CAL to continuous control tasks, performing simultaneous update can make neural networks fit Q functions better. We also compare SDQ-CAL with its variant DQ-CAL (without simultaneous update) that updates either $Q_{\theta_1}$ or $Q_{\theta_2}$ randomly as suggested in~\cite{hasselt2010double}, which underperforms SDQ-CAL by a large margin as shown in Table~\ref{table:table2}. 
	Besides, DQ-CAL obtains poorer sample efficiency due to the inadequate update of Q functions.
	
	\begin{figure*}[htbp]
		\centering
		\includegraphics[width=0.60\linewidth]{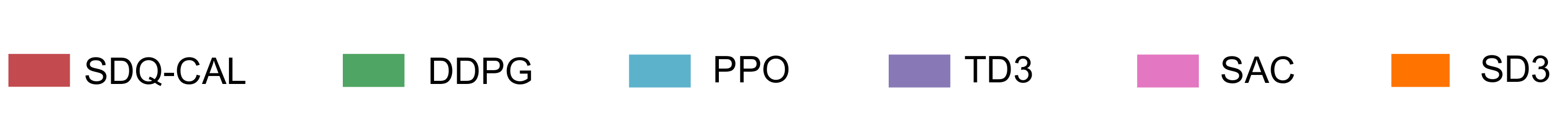}
		\\
		\vspace{0.5cm}
		\includegraphics[scale=0.248]{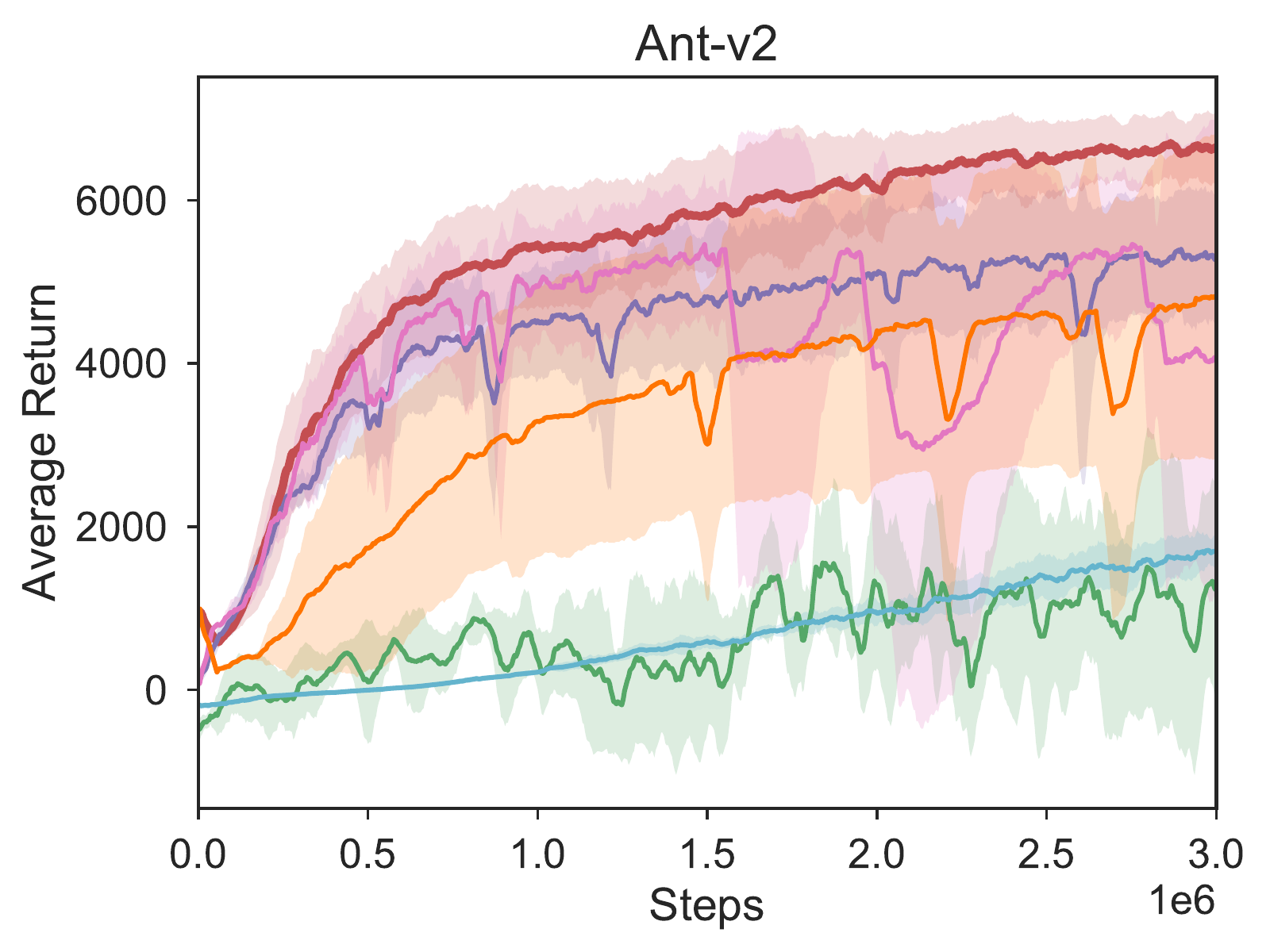}
		\hspace{0.1cm}
		\includegraphics[scale=0.248]{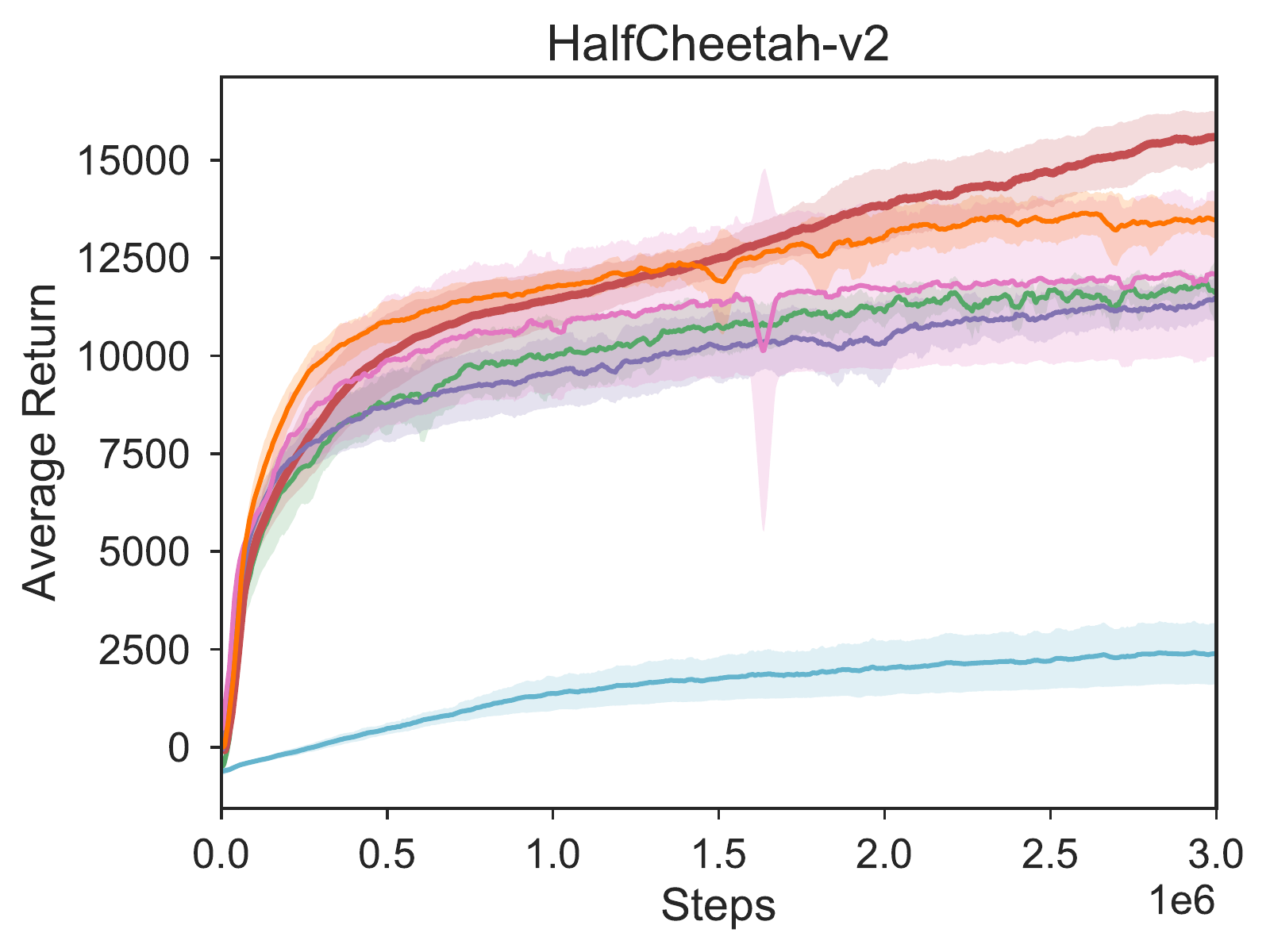}
		\hspace{0.1cm}
		\includegraphics[scale=0.248]{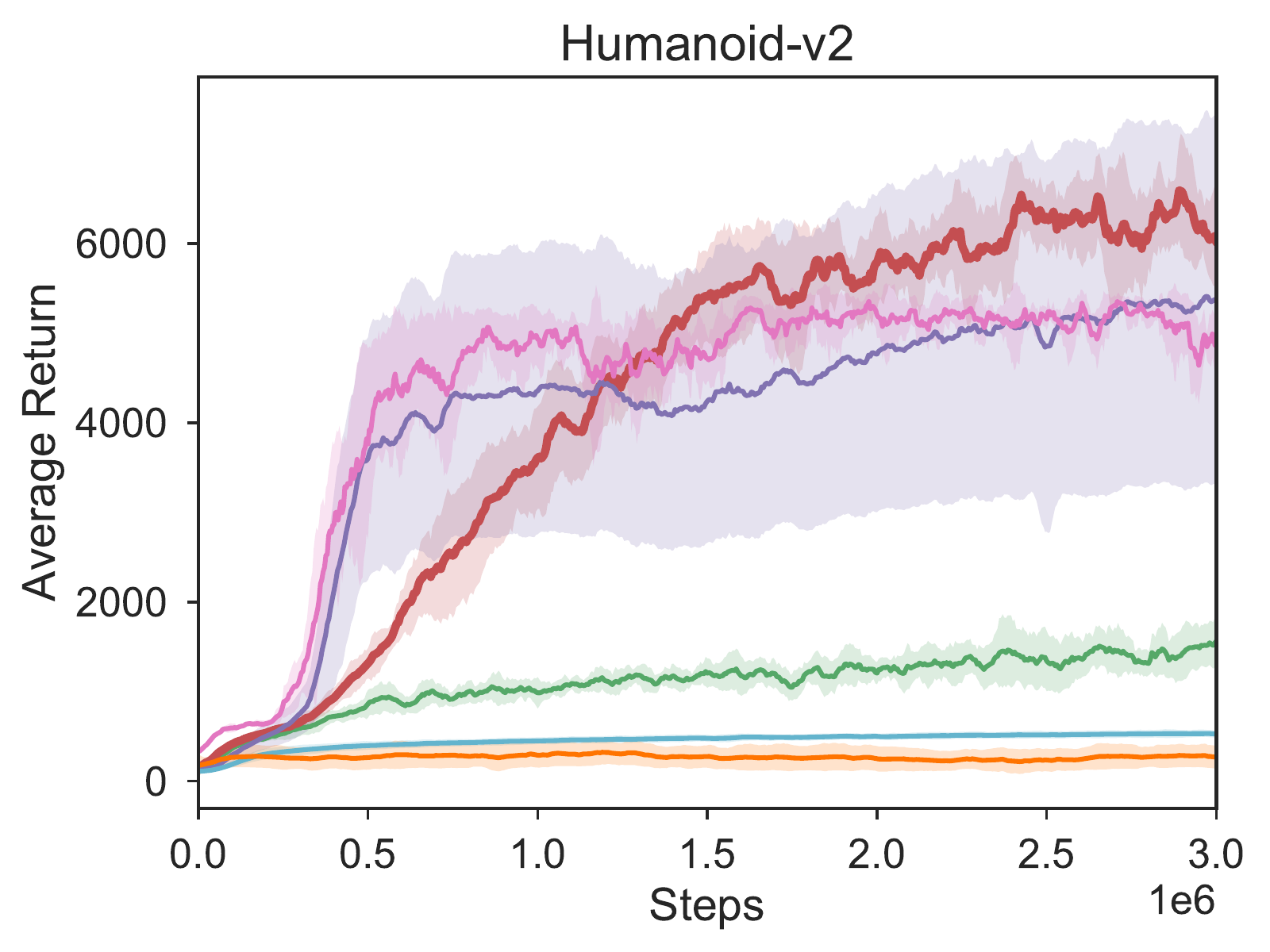}
		\hspace{0.1cm}
		\includegraphics[scale=0.248]{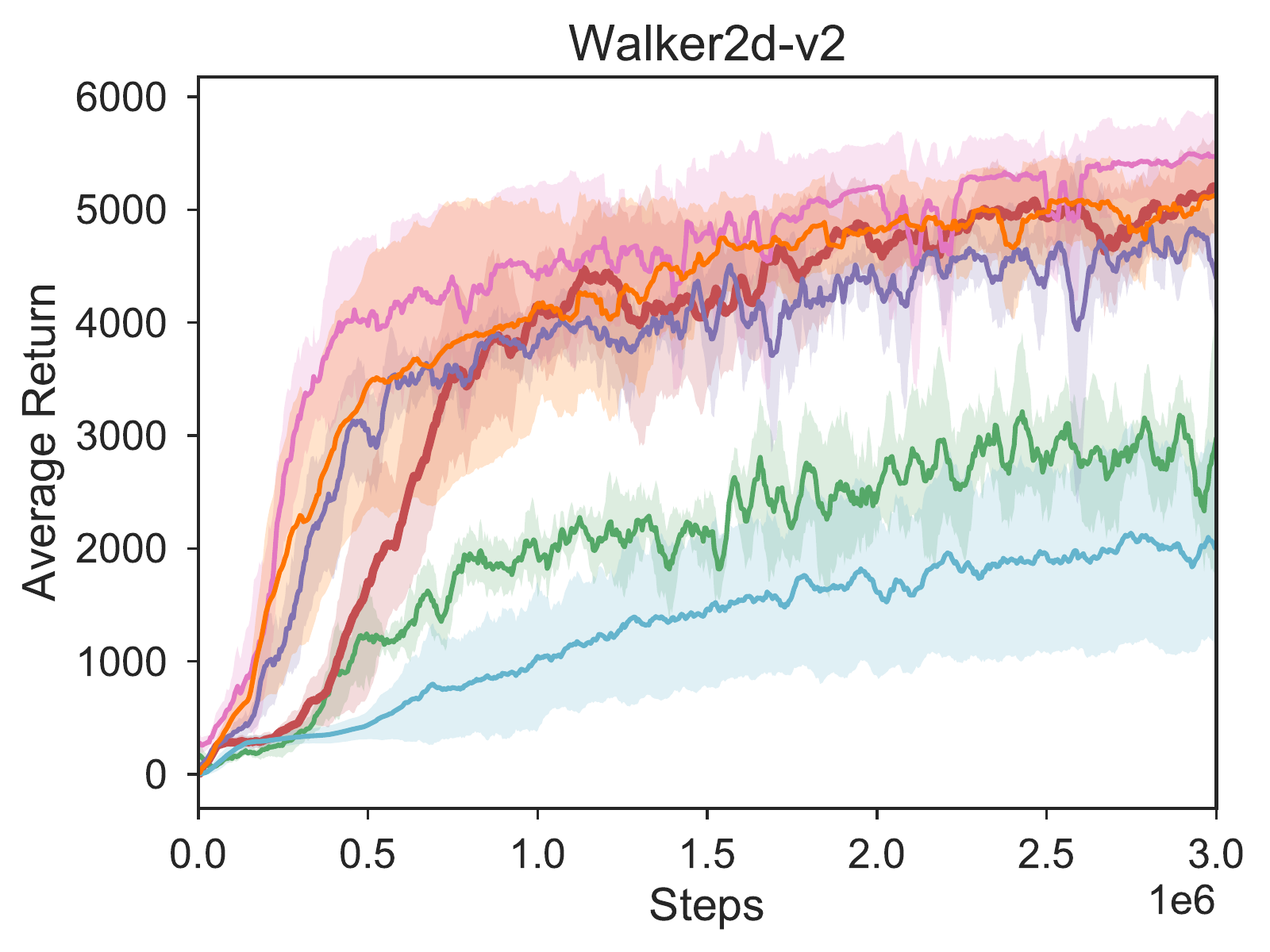}
		\hspace{0.1cm}
		\\
		\vspace{0.5cm}
		\includegraphics[scale=0.248]{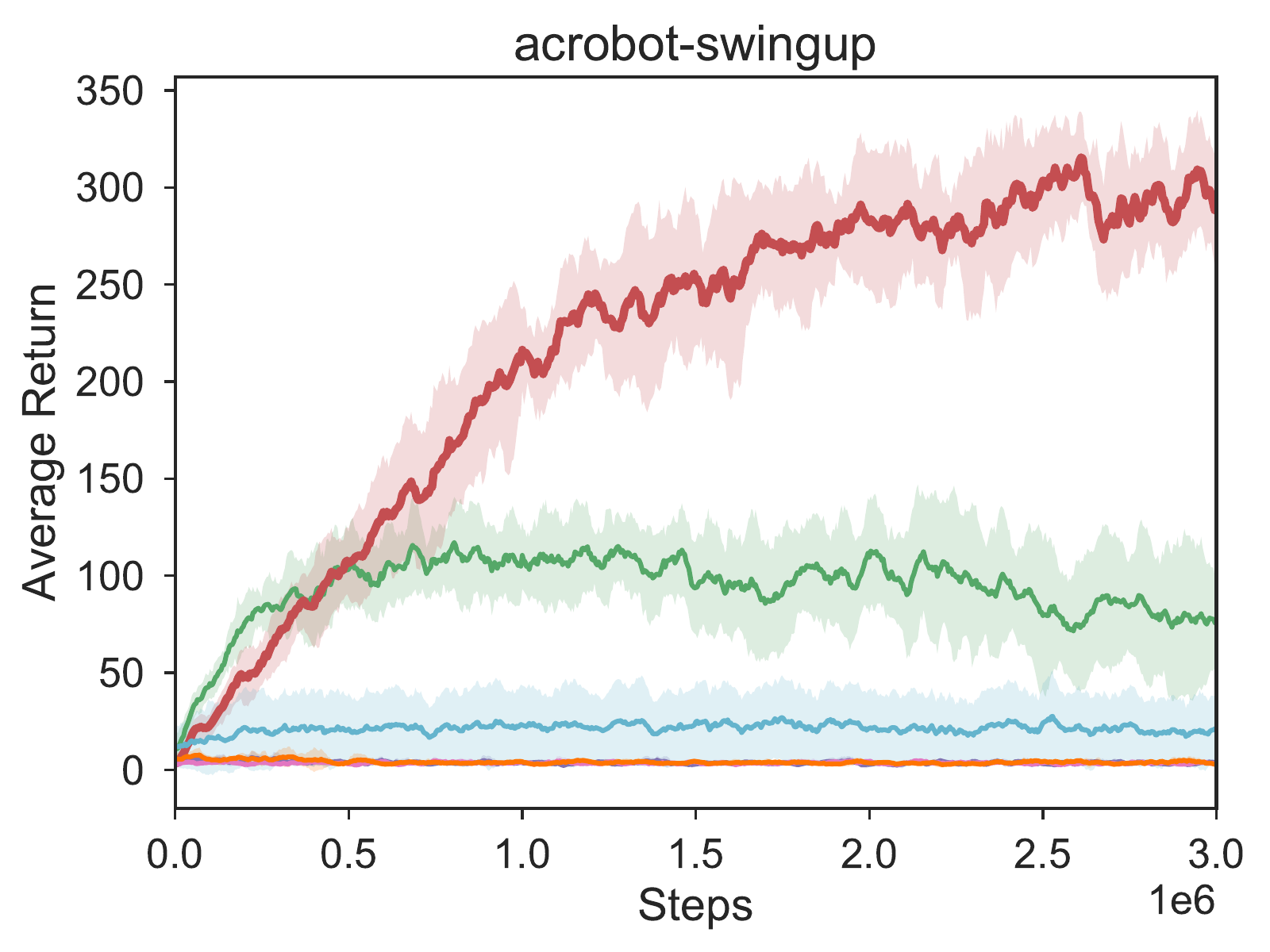}
		\hspace{0.1cm}
		\includegraphics[scale=0.248]{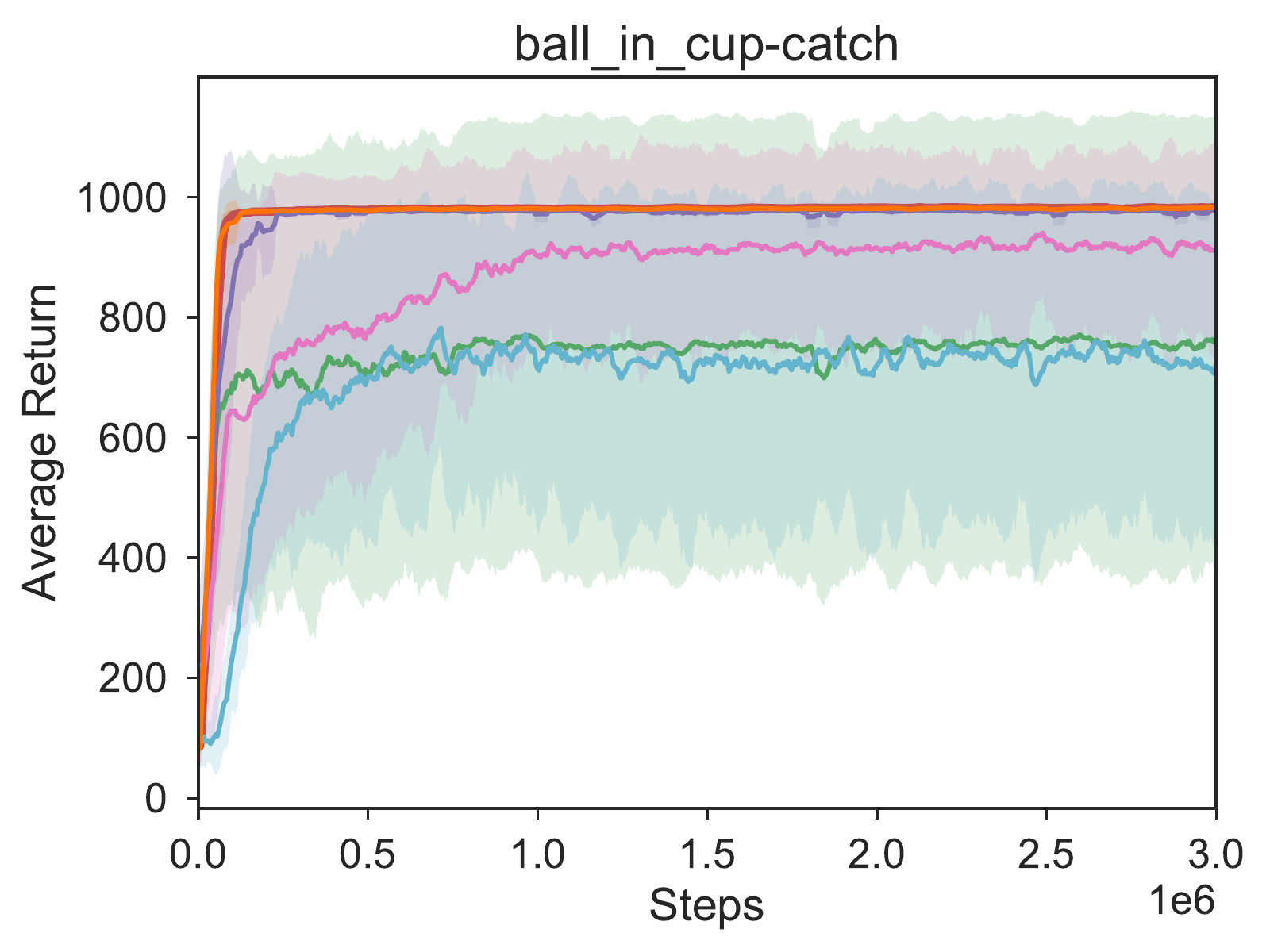}
		\hspace{0.1cm}
		\includegraphics[scale=0.248]{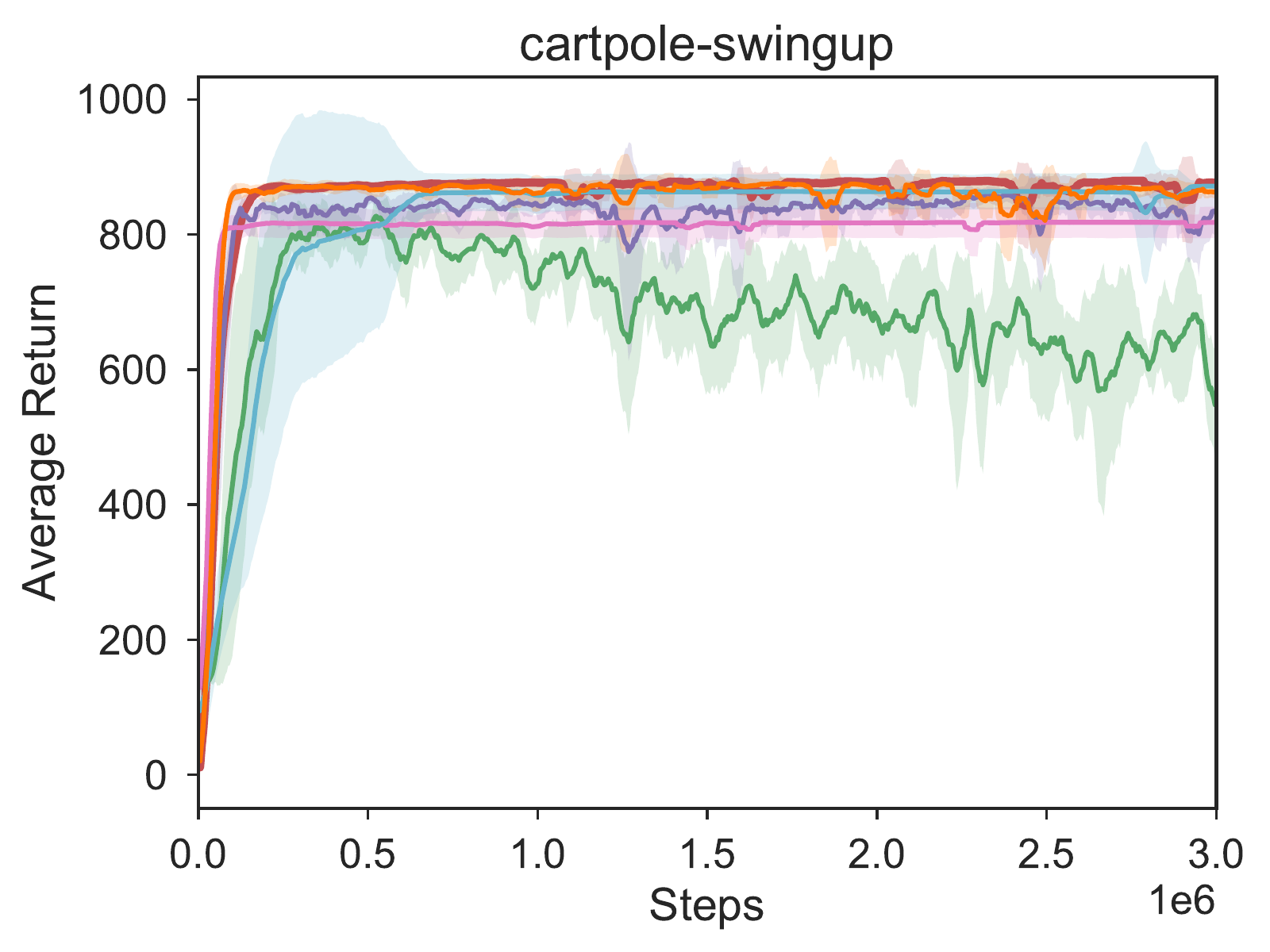}
		\hspace{0.1cm}
		\includegraphics[scale=0.248]{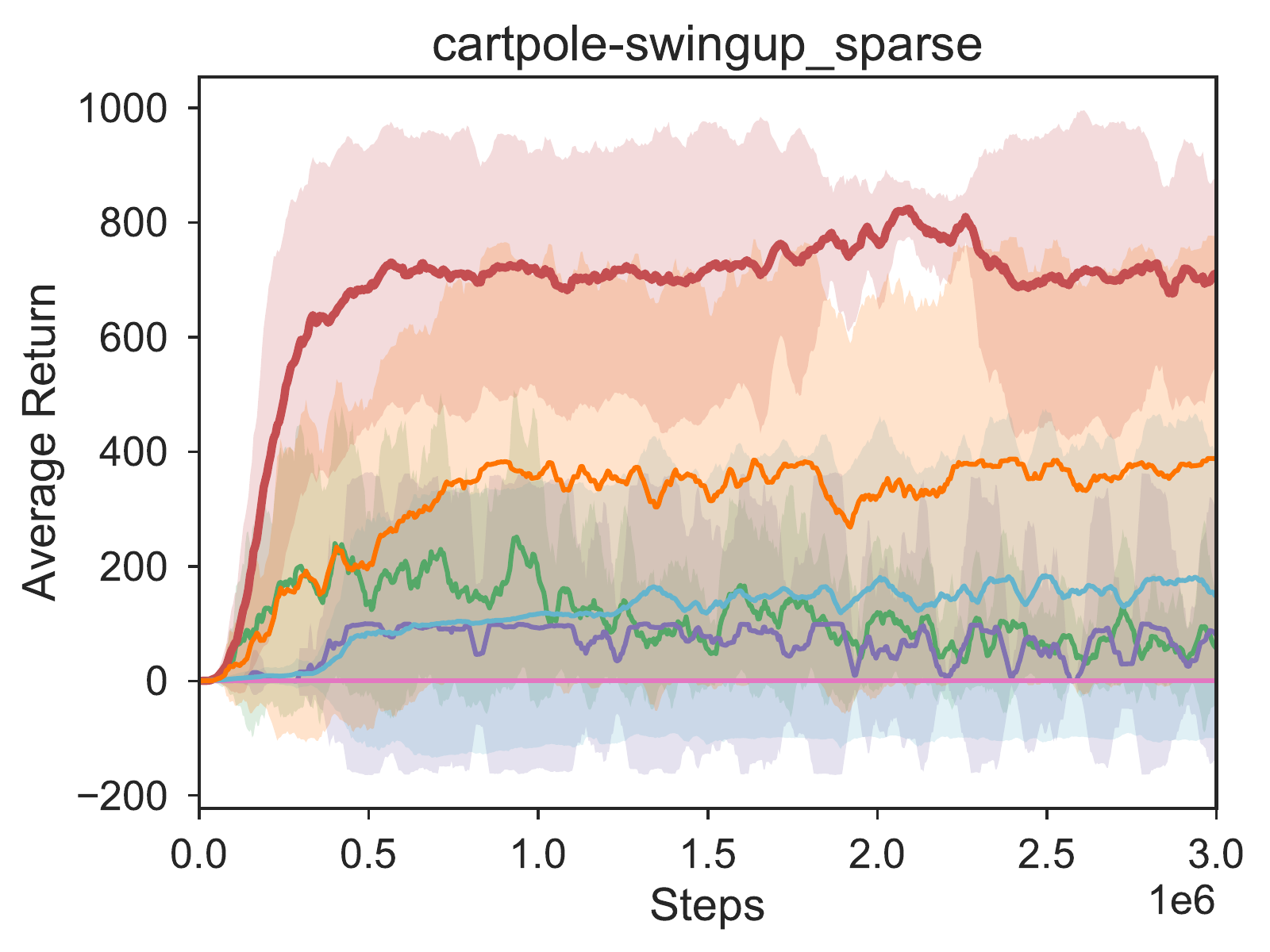}
		\hspace{0.1cm}
		\\
		\vspace{0.5cm}
		\includegraphics[scale=0.248]{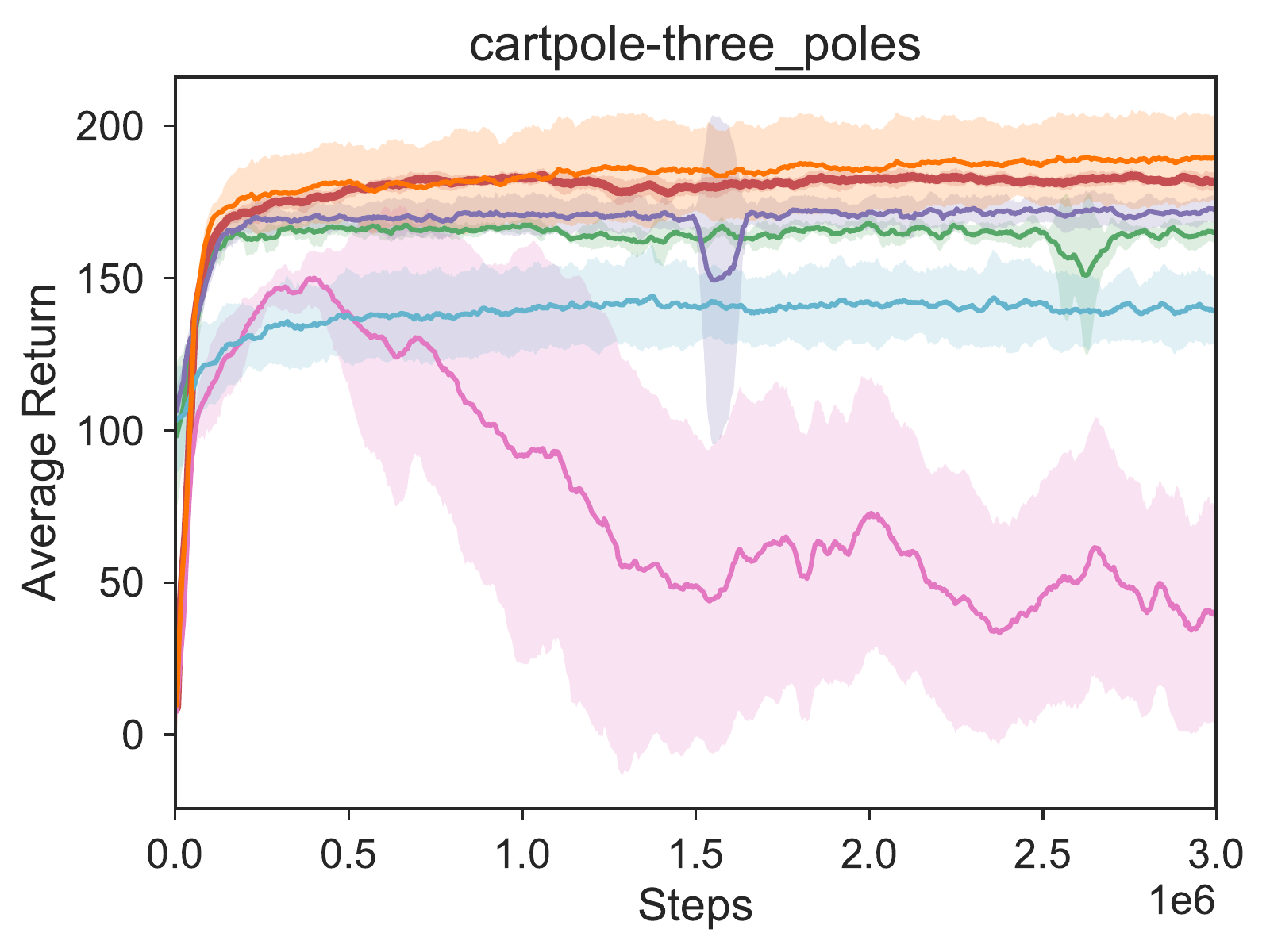}
		\hspace{0.1cm}
		\includegraphics[scale=0.248]{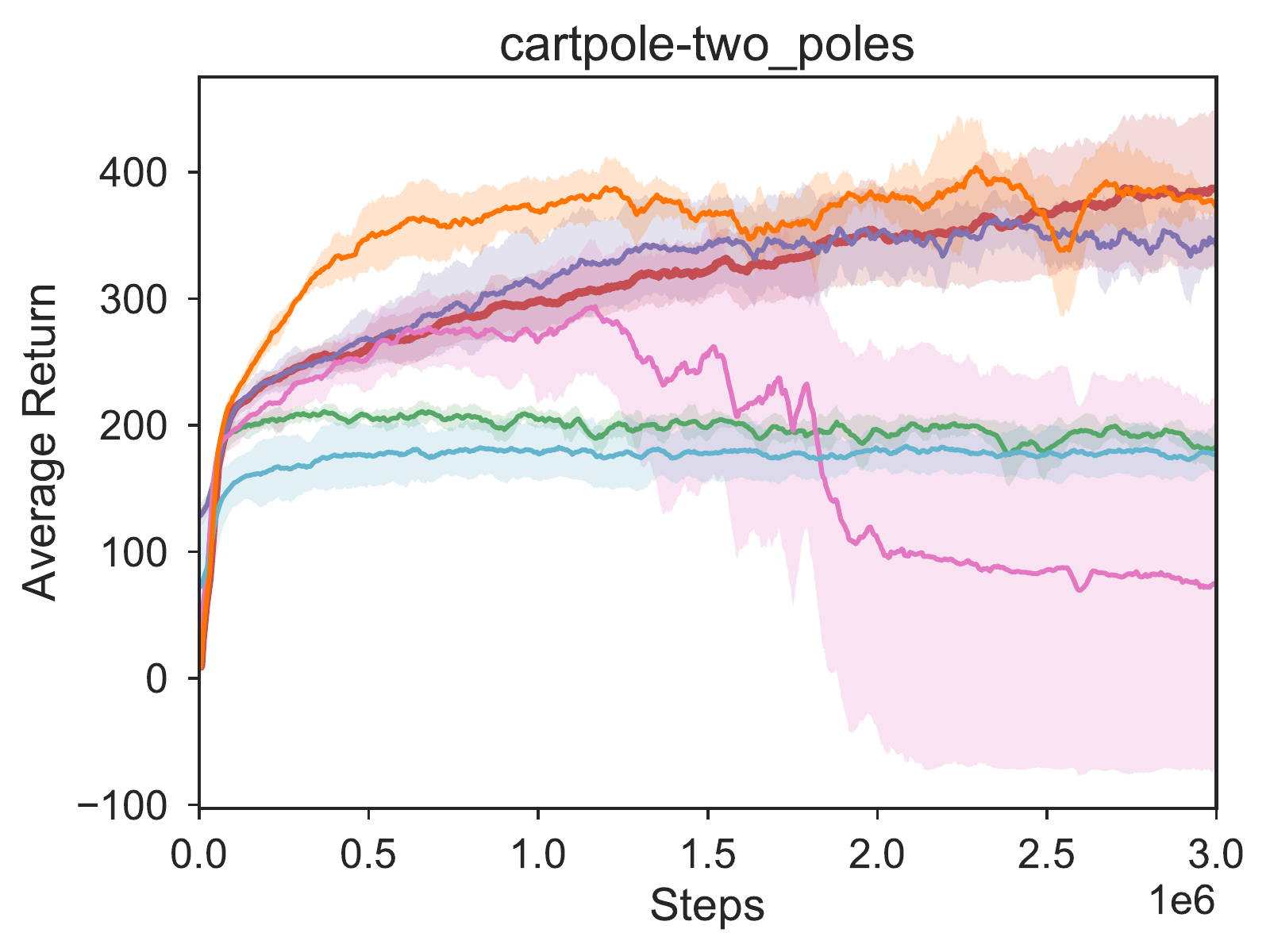}
		\hspace{0.1cm}
		\includegraphics[scale=0.248]{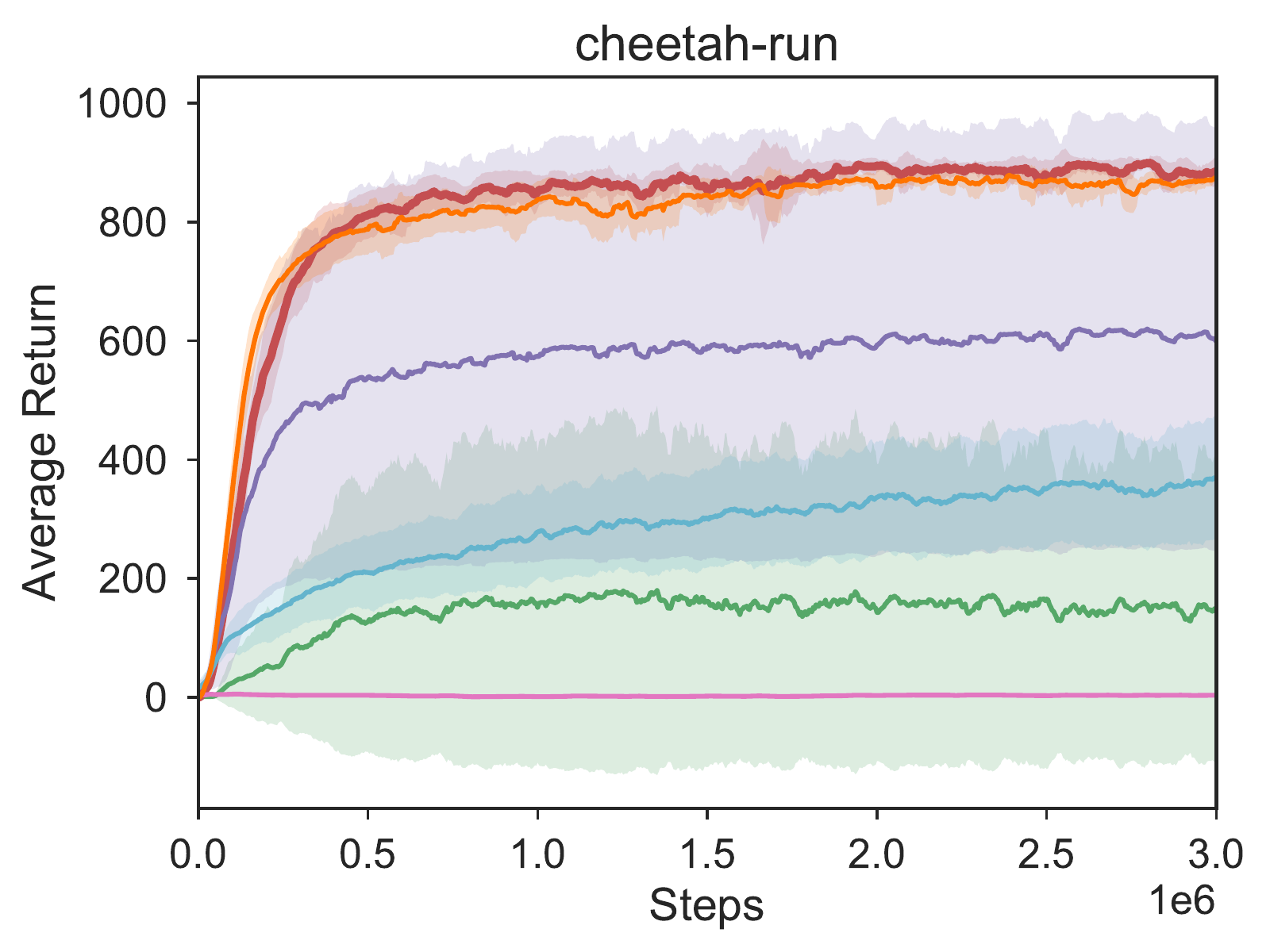}
		\hspace{0.1cm}
		\includegraphics[scale=0.248]{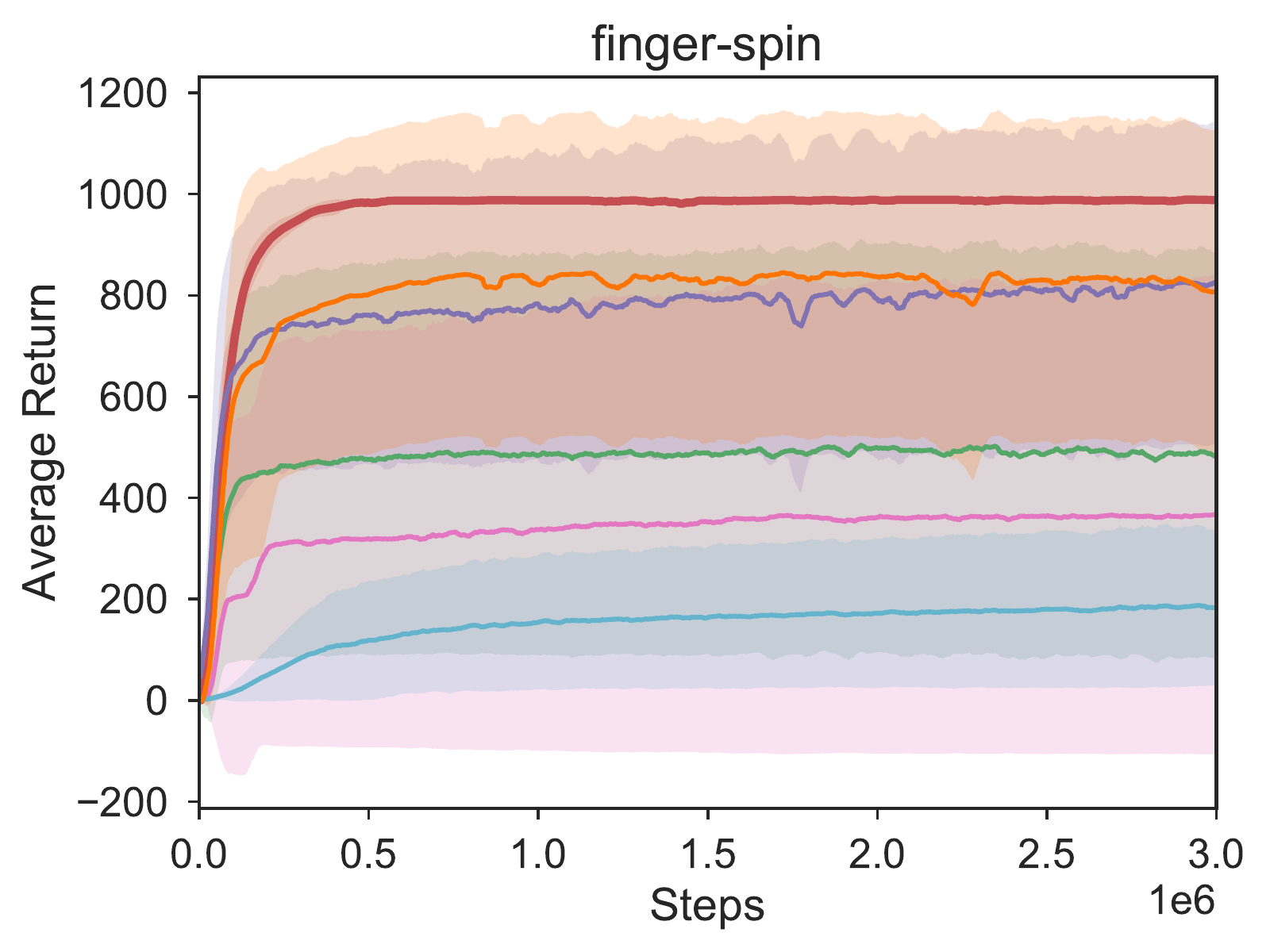}
		\hspace{0.1cm}
		\\
		\vspace{0.5cm}
		\includegraphics[scale=0.248]{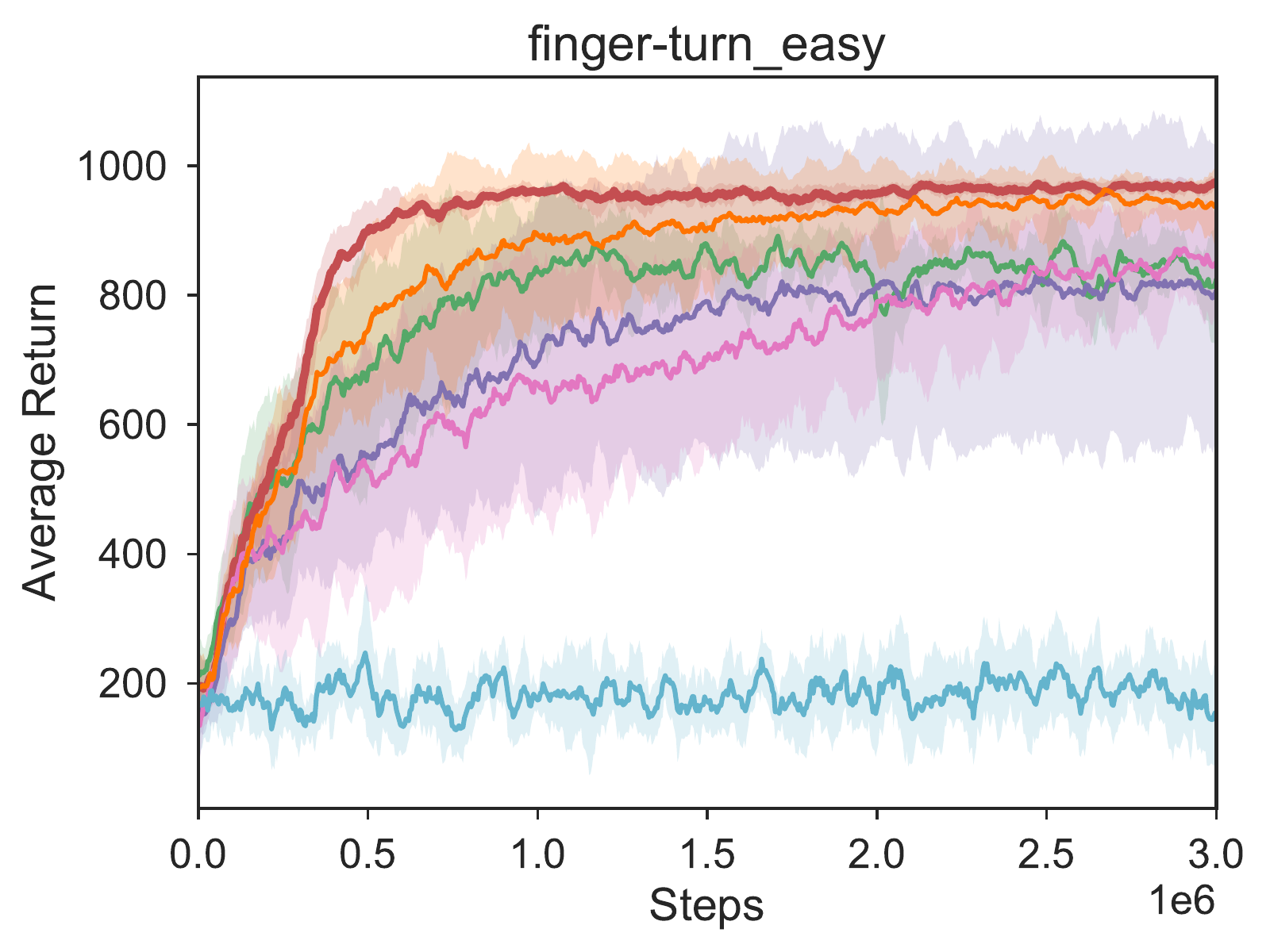}
		\hspace{0.1cm}
		\includegraphics[scale=0.248]{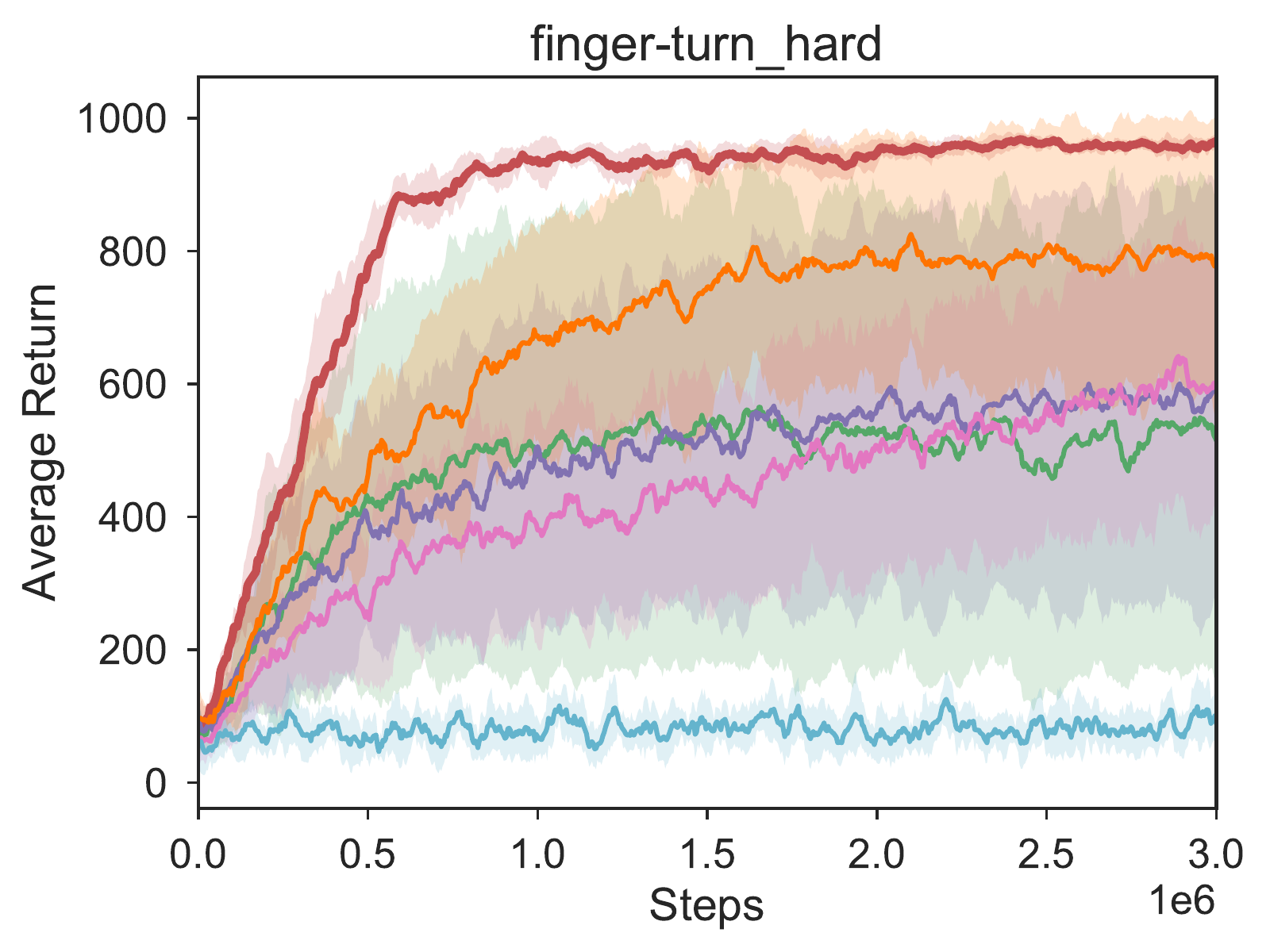}
		\hspace{0.1cm}
		\includegraphics[scale=0.248]{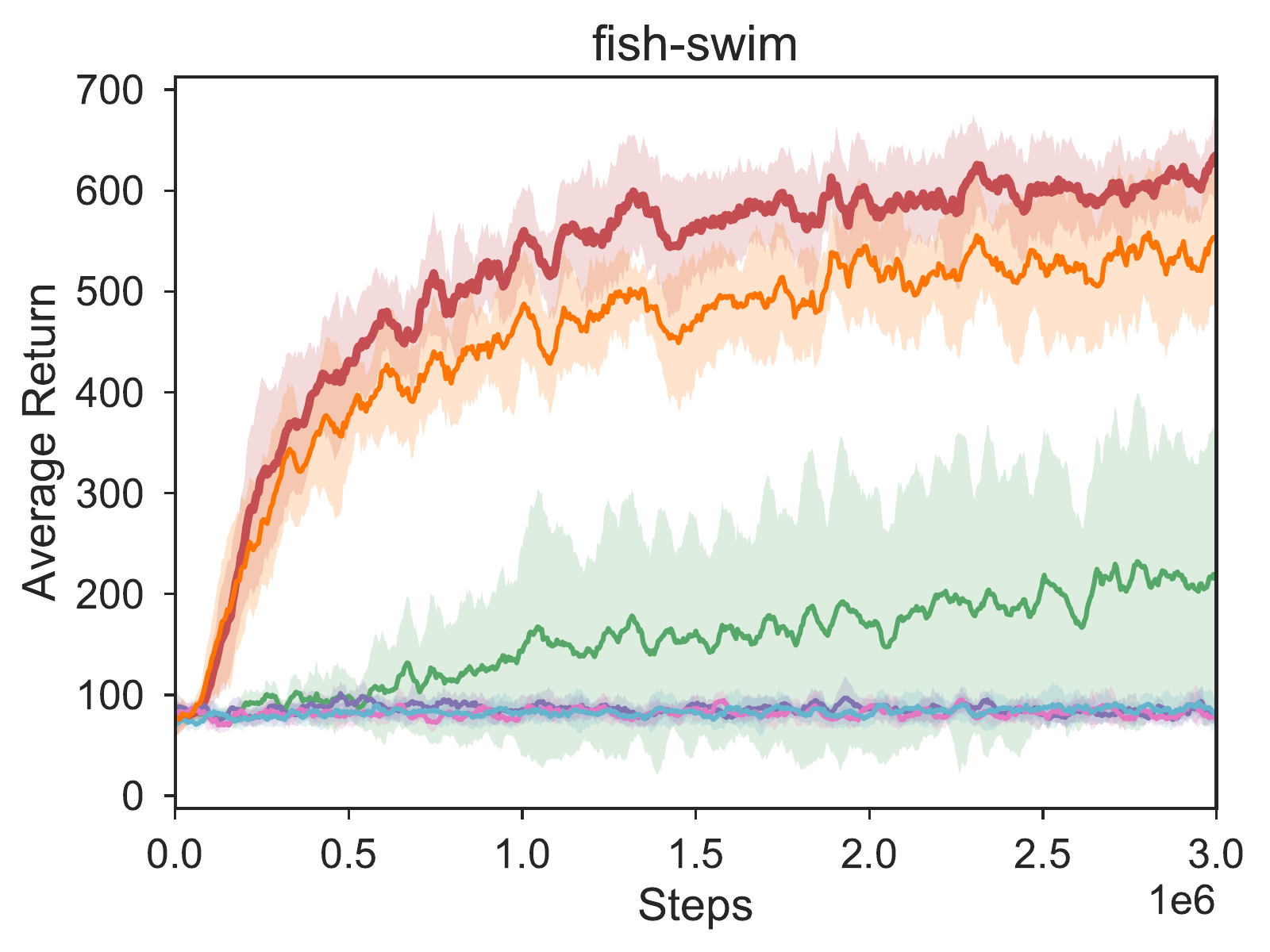}
		\hspace{0.1cm}
		\includegraphics[scale=0.248]{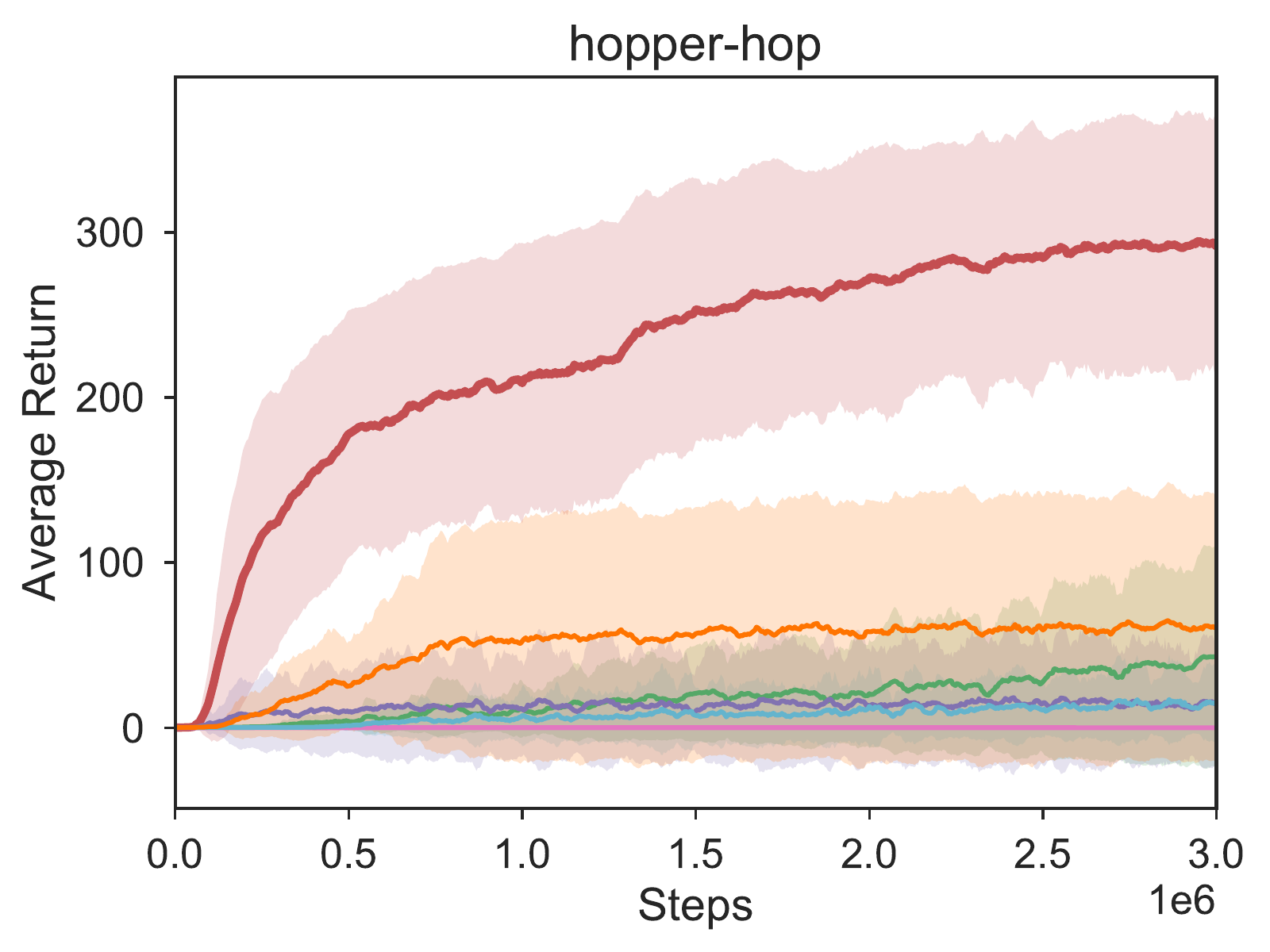}
		\hspace{0.1cm}
		\\
		\vspace{0.5cm}
		\includegraphics[scale=0.248]{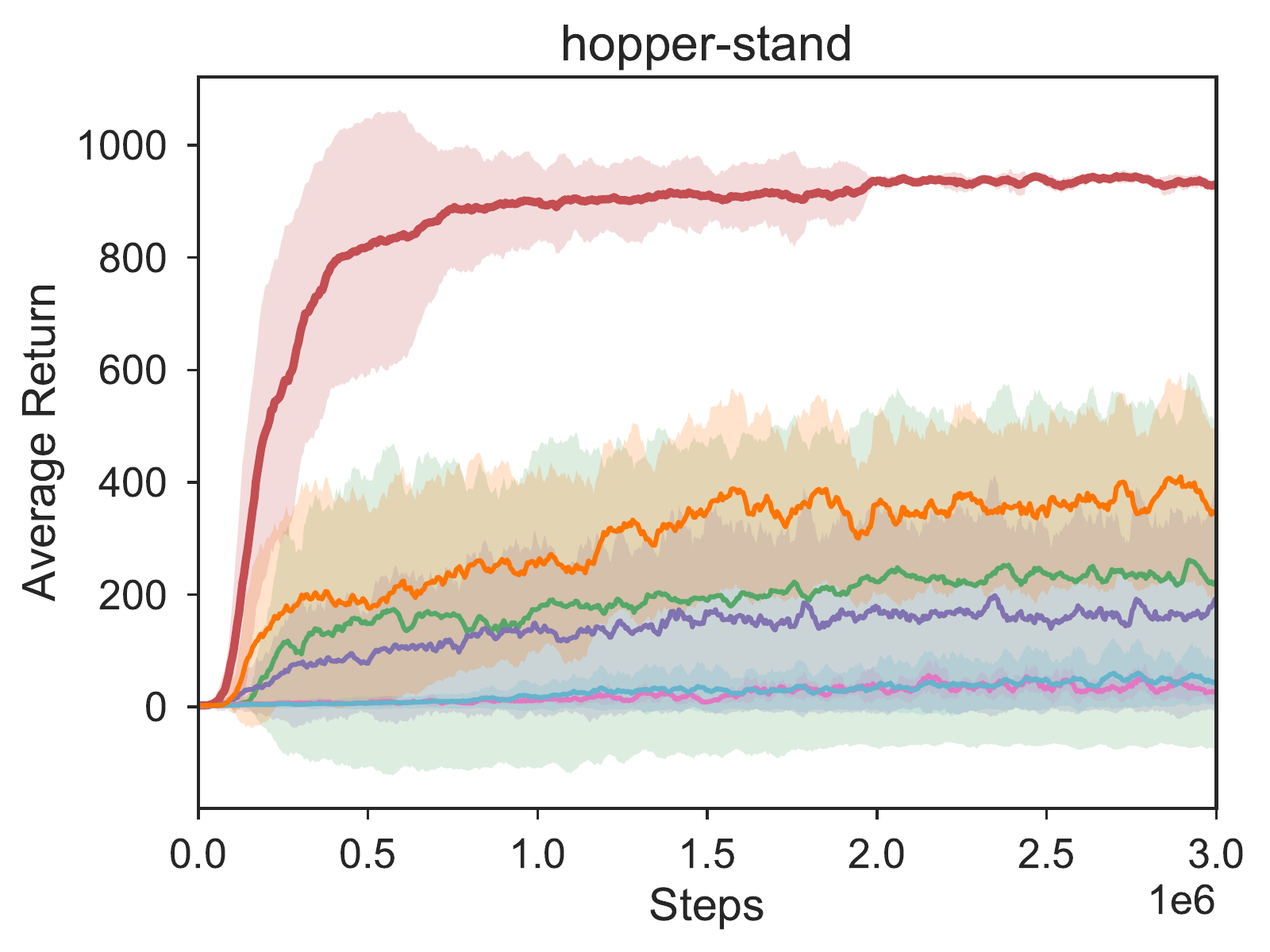}
		\hspace{0.1cm}
		\includegraphics[scale=0.248]{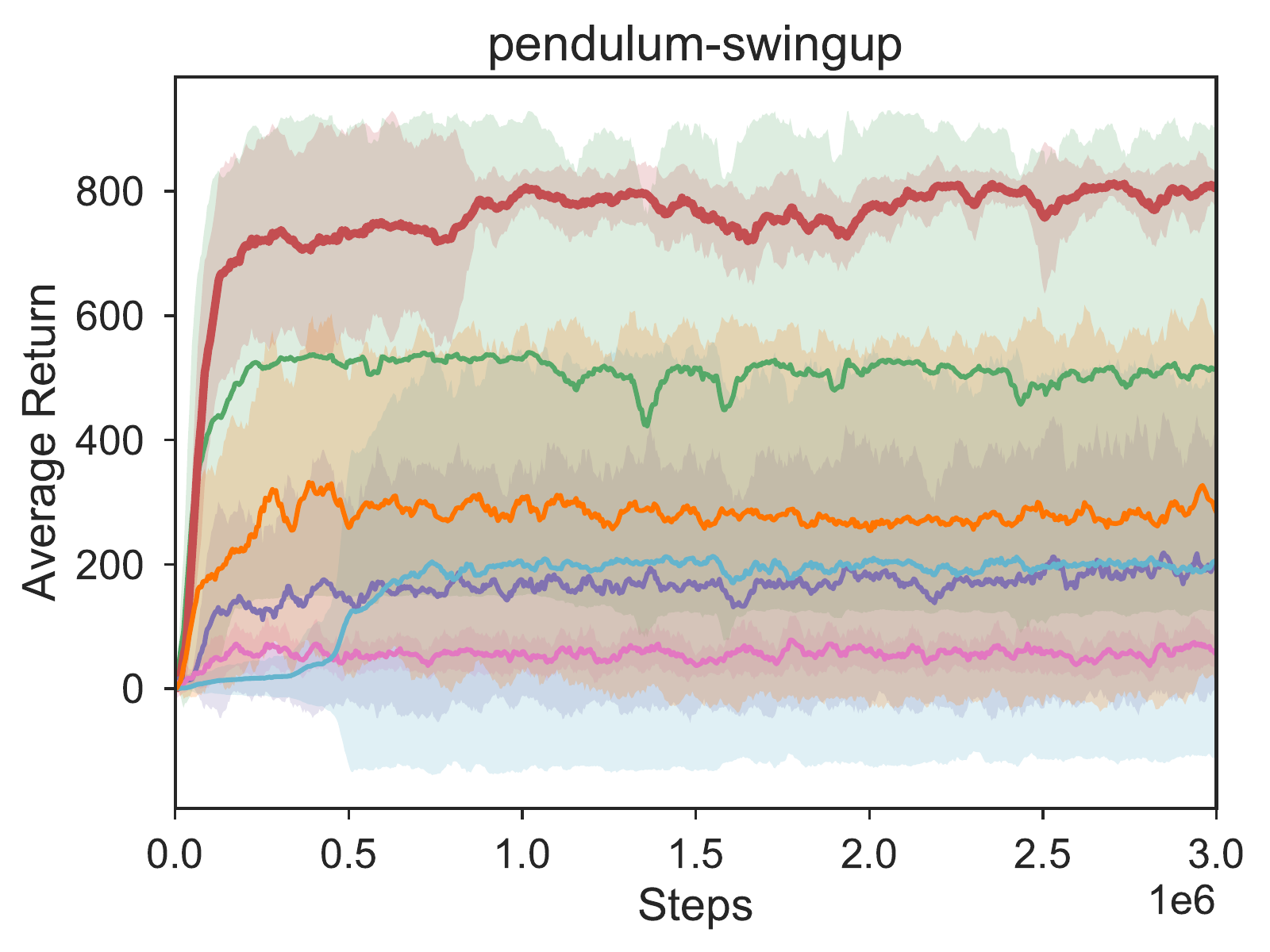}
		\hspace{0.1cm}
		\includegraphics[scale=0.248]{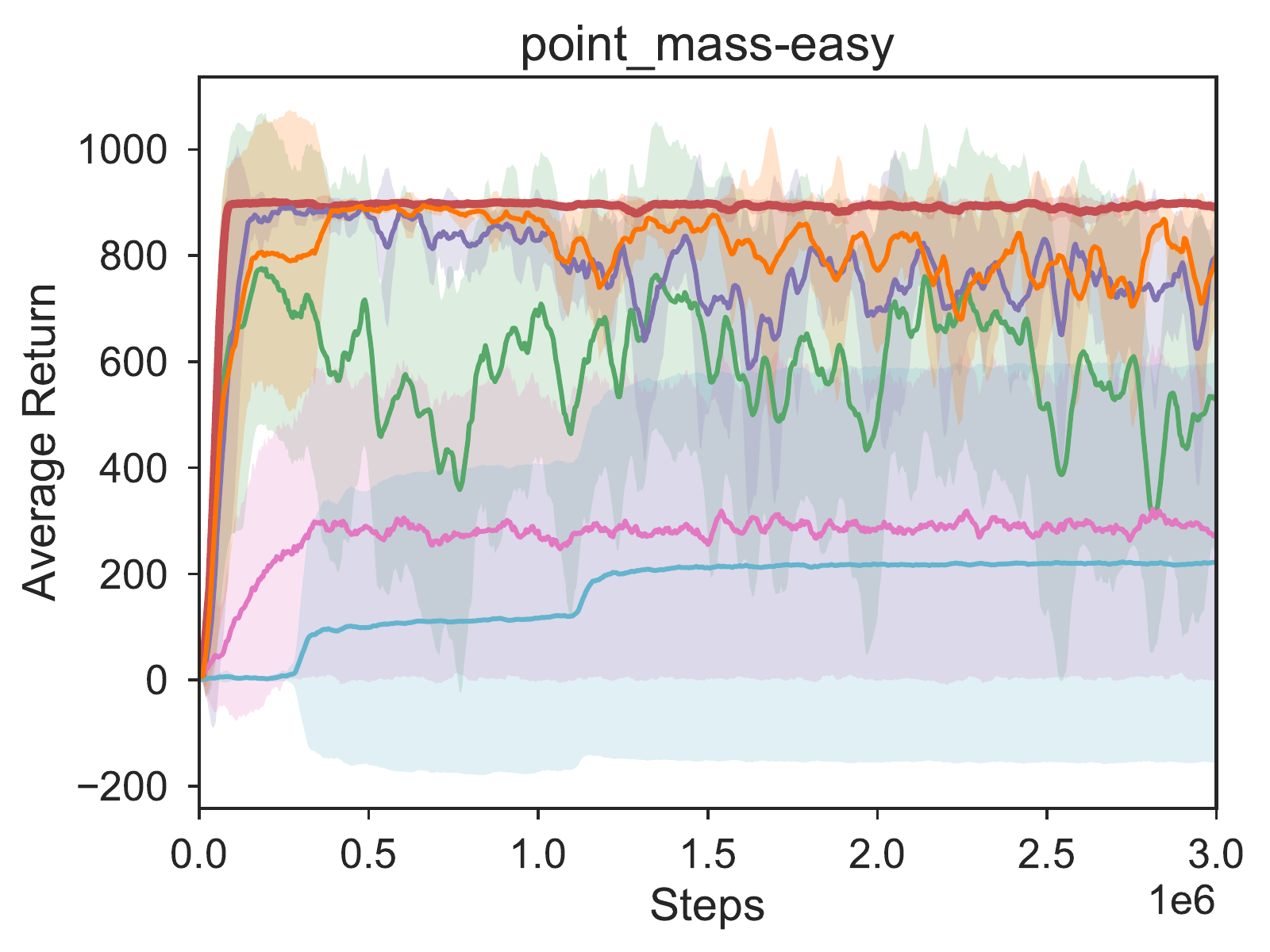}
		\hspace{0.1cm}
		\includegraphics[scale=0.248]{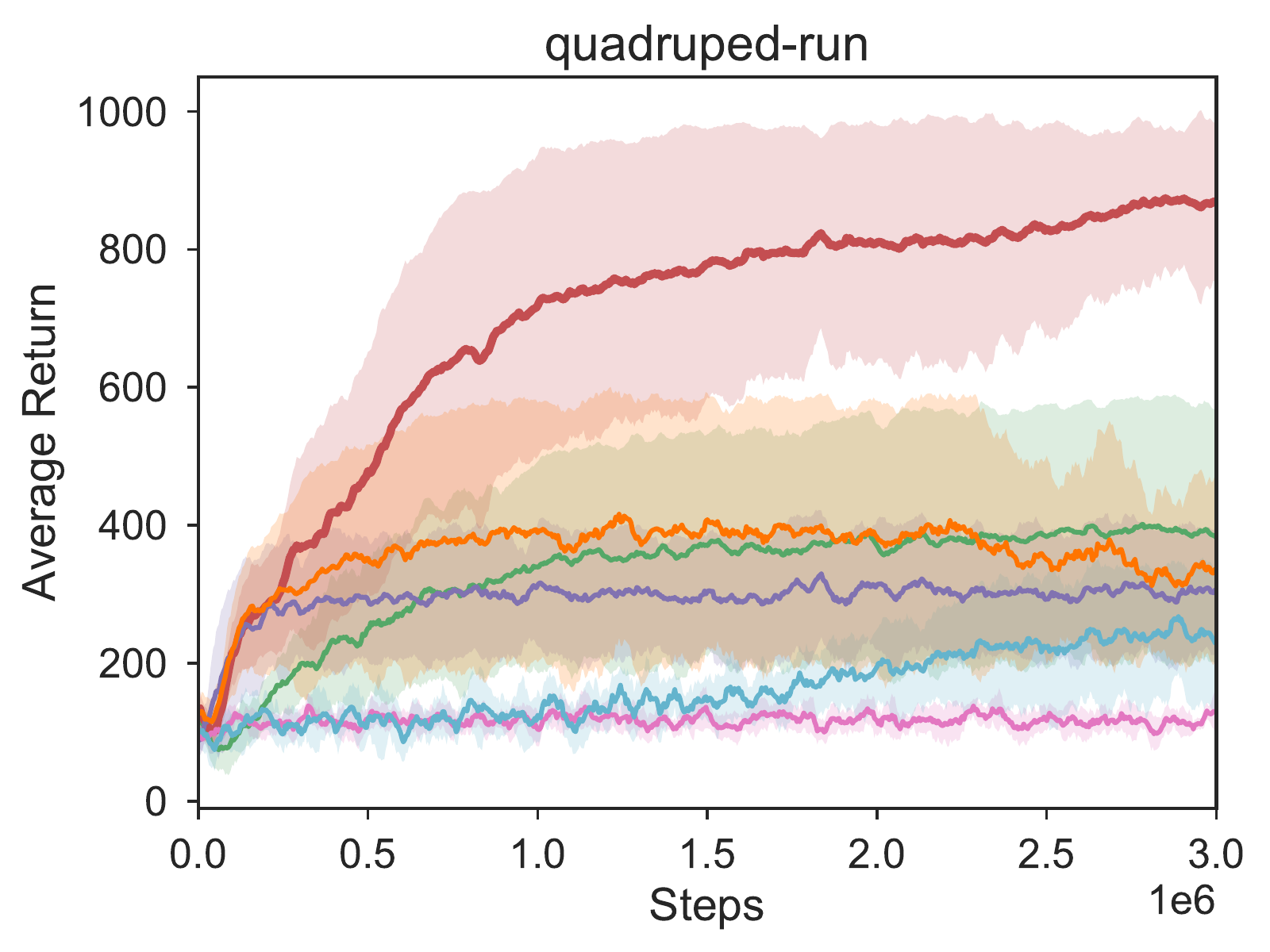}
		\hspace{0.1cm}
		\\
		\vspace{0.5cm}
		\includegraphics[scale=0.248]{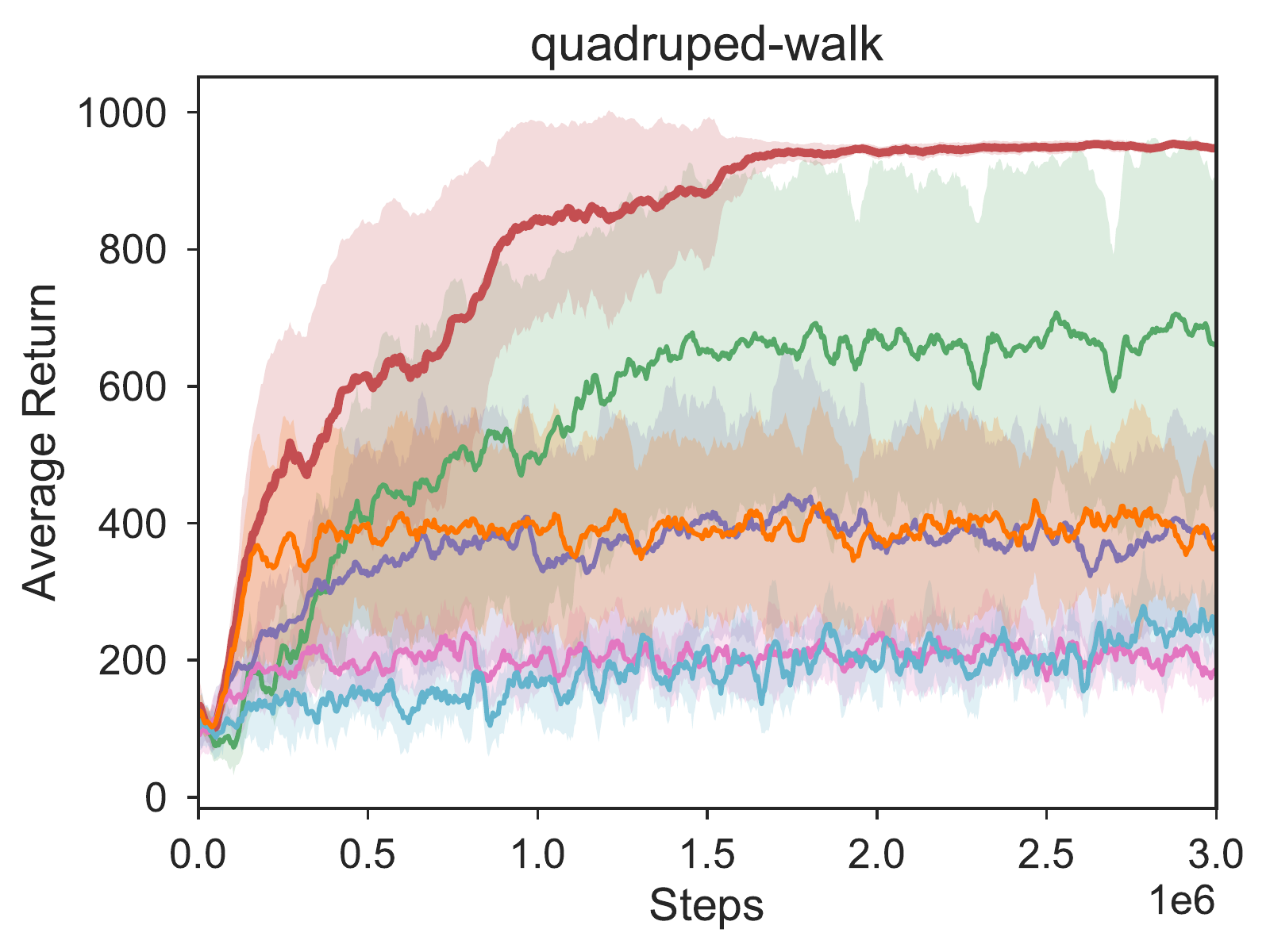}
		\hspace{0.1cm}
		\includegraphics[scale=0.248]{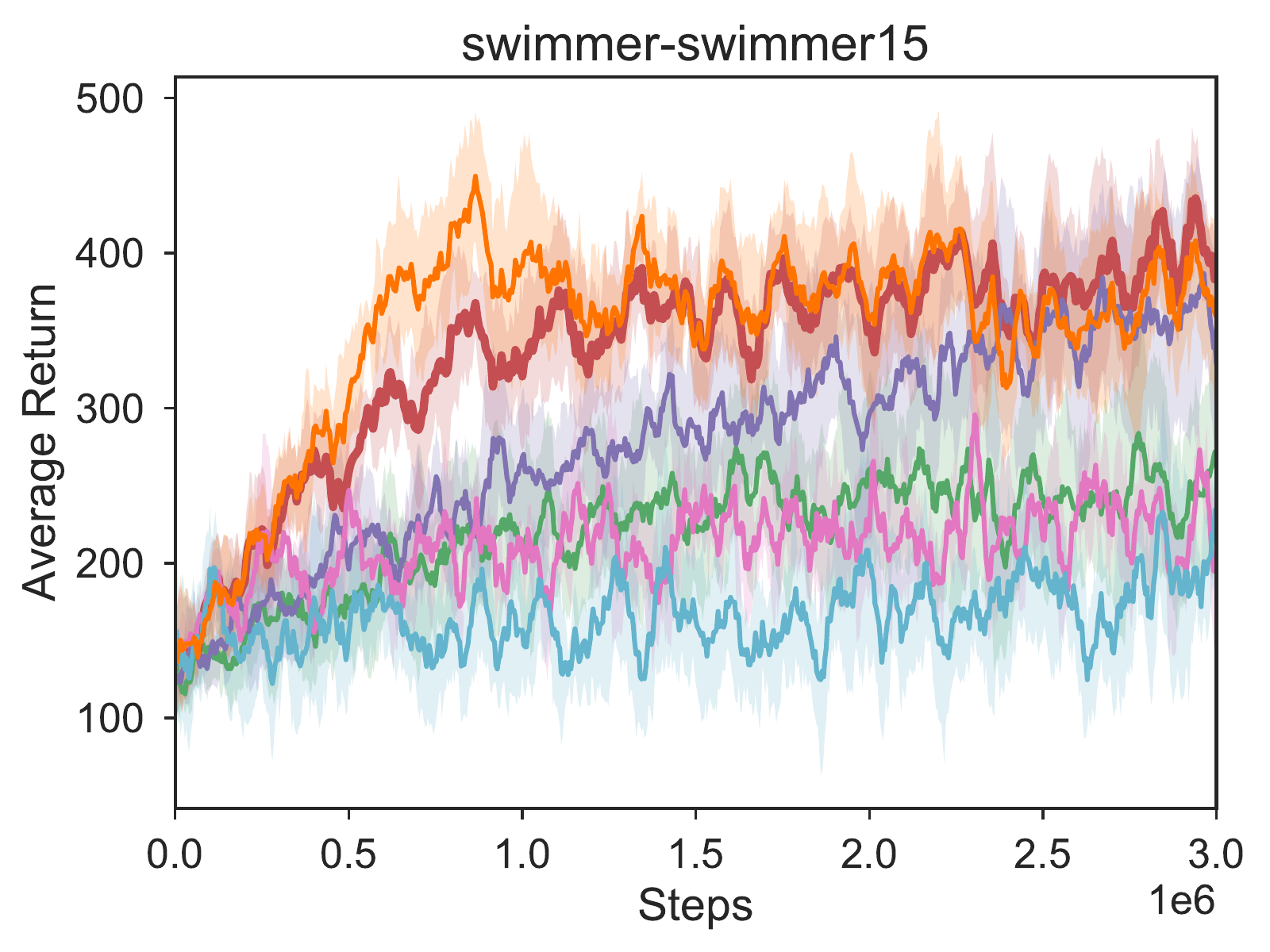}
		\hspace{0.1cm}
		\includegraphics[scale=0.248]{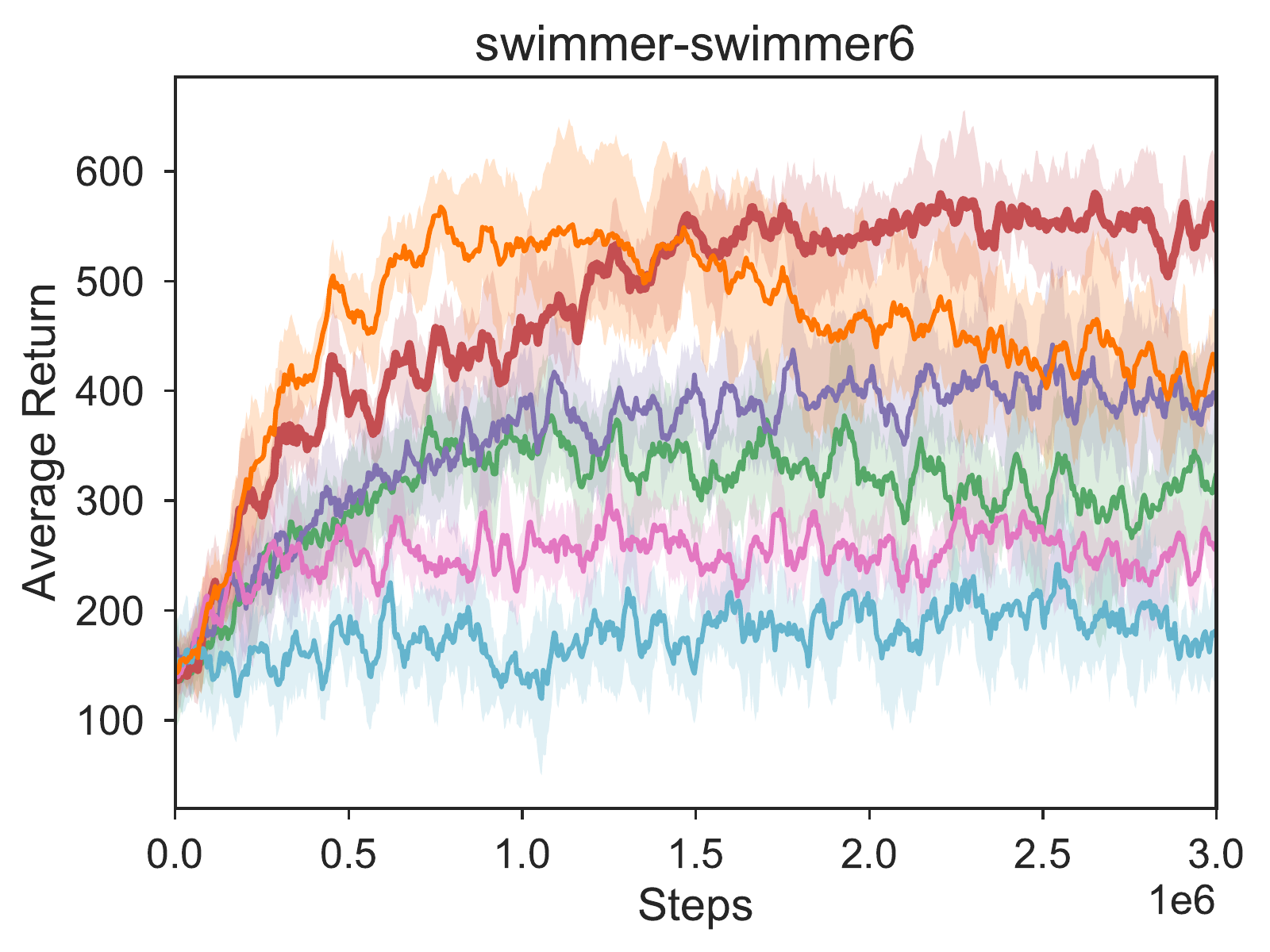}
		\hspace{0.1cm}
		\includegraphics[scale=0.248]{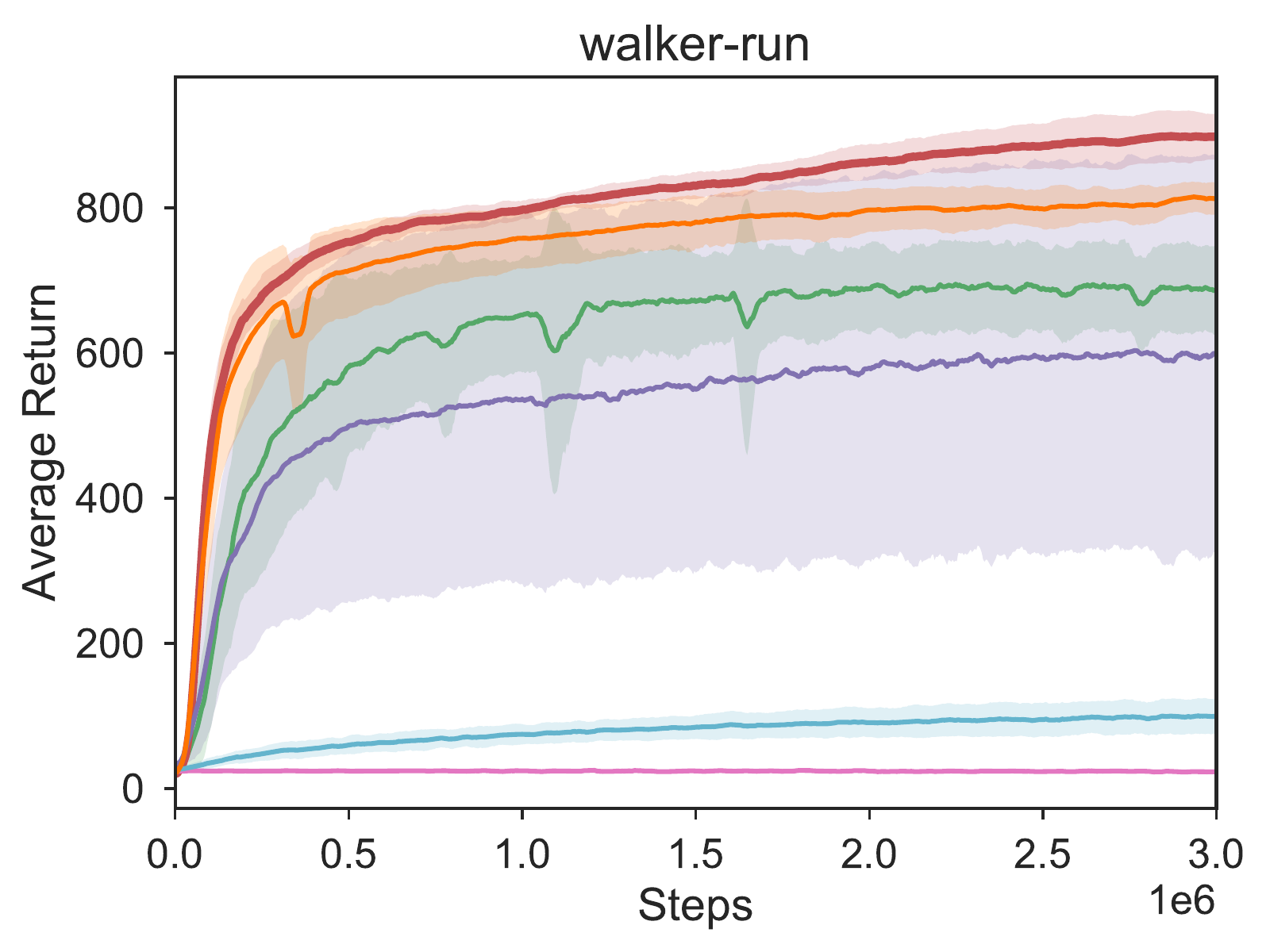}
		\\
		\hfill
		\caption{Comparison of SDQ-CAL to the top model-free actor-critic RL methods DDPG, PPO, TD3, SAC, and SD3 on $20$ DeepMind Control Suite environments, and $4$ MuJoCo environments. The curves show mean and standard deviation across $6$ seeds.}
		\label{fig:figure13}
	\end{figure*}
	
	\begin{figure}[htbp]
		\centering
		\small
		\includegraphics[width=0.8\linewidth]{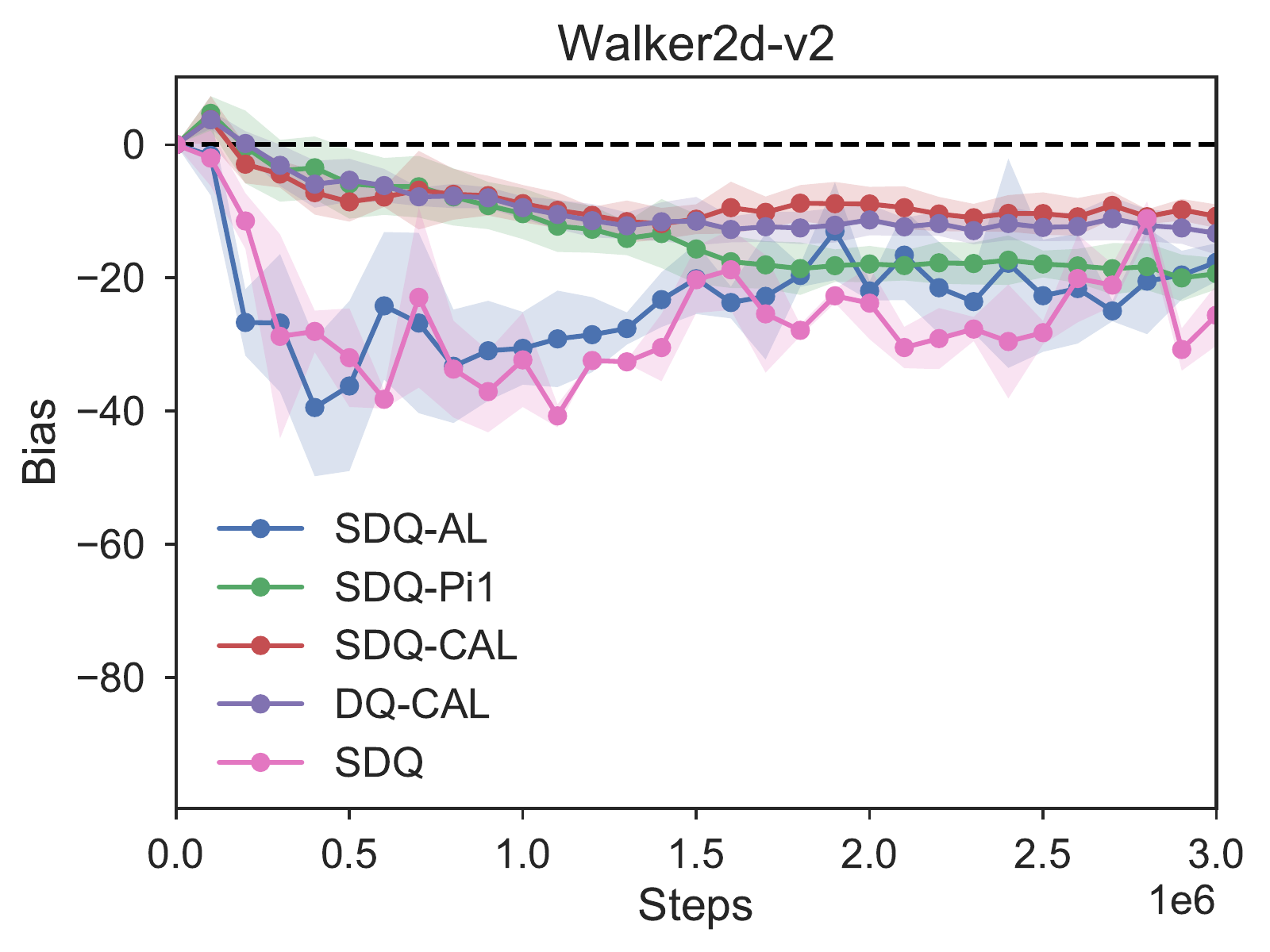}
		\caption{Measuring estimation bias in the value functions of SDQ-CAL (red), SDQ-AL, SDQ-Pi1, DQ-CAL, and SDQ on Walker2d-v2 environment over $3$ million steps.}
		\label{fig:figure12}
	\end{figure}
	
	\subsubsection{Analysis of value estimation bias} As discussed in Section~\ref{SDQQValueEstimation}, simultaneous Double Q-Learning is the key to obtaining less biased Q-value estimates. We further verify this opinion by computing the value estimates of SDQ-CAL and its variants in Walker2d-v2 environment. As shown in Fig.~\ref{fig:figure12}, SDQ-CAL and its variants both obtain relatively small estimation bias. This means that Double Q-learning is the key to mitigating the underestimation bias and the actor-critic variants of Double Q-learning can bring in lower estimation bias in continuous control tasks.
	
	\subsection{Additional Learning Curves for SDQ-CAL}	
	We provide a more detailed experimental comparison by comparing SDQ-CAL with DDPG, PPO, TD3, SAC, and SD3 on more continuous control tasks. The learning curves, plus learning curves for the additional $15$ games, are illustrated in Fig.~\ref{fig:figure13}. It is observed that SDQ-CAL performs consistently across all tasks and outperforms both on-policy and off-policy algorithms in the most challenging tasks.
	
	\section{Conclusion} \label{conclusion}
	In this work, we present a new off-policy actor-critic RL algorithm called Simultaneous Double Q-learning with Conservative Advantage Learning (SDQ-CAL). SDQ-CAL copes with overestimation bias and poor sample efficiency issues in actor-critic RL algorithms by updating a pair of critics simultaneously upon double estimators and modifying the reward functions with conservative Advantage Learning. 
	When extended to a deep RL setting, our algorithm makes the first attempt by incorporating Advantage learning for continuous control, achieving improved sample efficiency. 
	Extensive experimental results on standard continuous control benchmarks validate the effectiveness of SDQ-CAL, which exceeds the performance of numerous state-of-the-art model-free RL methods. Moreover, our general approach can be easily integrated into any other actor-critic RL algorithm. 
	Finally, we believe the sheer simplicity of our approach highlights the ease of reproducibility of RL algorithms made by the community.

	\bibliographystyle{IEEEtran}
	\bibliography{tnnls22}
	\newpage
	
\end{document}